%% file: main.tex
\documentclass[10pt,journal,compsoc]{IEEEtran}

\usepackage{newlfont}
\usepackage{color}
\definecolor{shadecolor}{rgb}{0.92,0.92,0.92}
\usepackage{colortbl}
\usepackage[table]{xcolor}
\usepackage[color,matrix,arrow,all]{xy}

\usepackage{marvosym}
\usepackage{tikz}
\usepackage{graphicx}
\usepackage{float}
\usepackage{subfigure}
\usepackage{multirow}
\usepackage{multicol}
\usepackage{marginnote}
\usepackage{CJKutf8}
\usepackage{pifont}
\usepackage{hyperref}
\usepackage{natbib} 
\usepackage[edges]{forest}
\usepackage{booktabs}   
\usepackage{ragged2e}
\usepackage{tabularx}   

\usepackage{multirow}   
\usepackage{enumitem}   
\usepackage{array}      
\usepackage[T1]{fontenc}
\usepackage{tabularx}
\usepackage{makecell}

\usepackage{graphicx,epsfig}
\usepackage{balance}
\usepackage{amsmath,amssymb,amsfonts,bm}
\usepackage{framed}
\usepackage{lipsum}
\usepackage{color}
\usepackage{hyperref}
\usepackage{balance}  
\usepackage{listings,upquote,soul,framed}
\usepackage{hyperref}
\usepackage{xspace}
\usepackage{epstopdf}
\usepackage{enumitem}
\usepackage{graphicx}
\usepackage[mathscr]{eucal}
\usepackage{graphicx}
\usepackage{algorithmic}
\usepackage{amssymb}
\usepackage{epstopdf}
\usepackage{epsfig,endnotes}
\usepackage{url}
\usepackage{dsfont}
\usepackage{array}
\usepackage{booktabs}
\usepackage{threeparttable}
\usepackage[lined,boxed,vlined,ruled,linesnumbered]{algorithm2e}
\usepackage[justification=centering]{caption}
\colorlet{shadecolor}{gray!20}
\usepackage{forest}
\usetikzlibrary{arrows.meta,shapes,positioning,shadows,trees}
\usepackage[english]{babel}

\usepackage[utf8]{inputenc}
\usepackage{amsmath}
\usepackage{amssymb}
\usepackage{tikz,lipsum,lmodern}
\usepackage[most]{tcolorbox}
\usepackage{graphicx}
\usepackage{afterpage}
\usepackage{lettrine}

\NewDocumentCommand{\zxh}{ mO{} }{\textcolor{blue}{\textsuperscript{\textit{Xuanhe}}\textsf{\textbf{\small[#1]}}}}

\definecolor{zhycolor}{RGB}{0,84,0}

\newcommand{\llm}{\textsc{LLM}\xspace}
\newcommand{\llms}{\textsc{LLMs}\xspace}

\newcommand{\hi}[1]{\vspace{.25em} \noindent {\bf #1}\xspace}

\newcommand{\llmasanalyst}{{LLM/Agent-as-Data-Analyst}\xspace}

\newcommand{\insertfig}{\includegraphics[width=.95\linewidth]{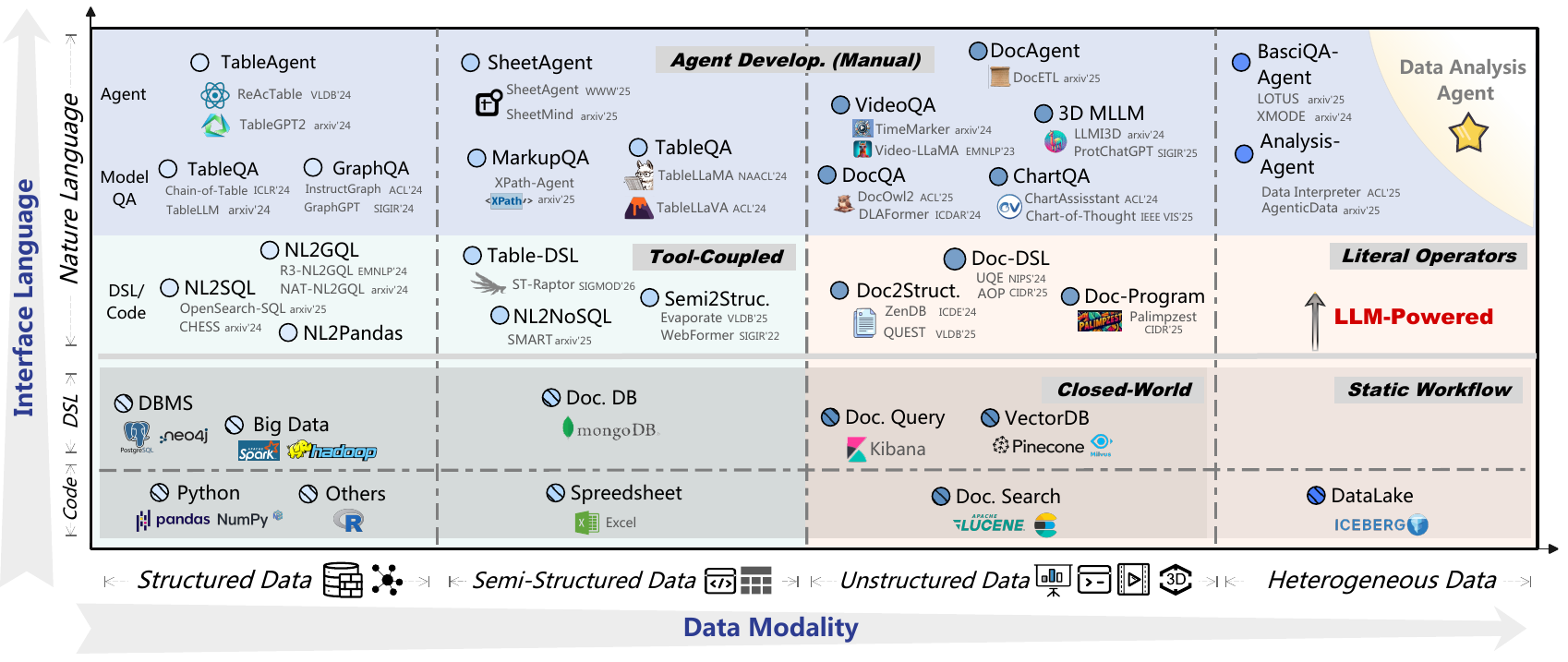}\vspace{-.25cm}\captionof*{figure}{Fig. 1: The  Evolution of {\bf \llmasanalyst} follows a four-dimension trajectory: (1) Knowledge Scope (closed-world $\rightarrow$ open-world); (2) Tool Integration (tool-coupled $\rightarrow$ tool-agnostic); (3) Analysis Functionality (literal $\rightarrow$ semantic); (4) Development Autonomy (static workflow $\rightarrow$ dynamic workflow and manual $\rightarrow$ fully autonomous).}\vspace{-.5cm}\label{fig:framework}}

\makeatletter
\apptocmd{\@maketitle}{\centering\insertfig}{}{}
\makeatother

\begin{document}

\pagestyle{plain}
\pagenumbering{roman}

\twocolumn


	
\pagestyle{plain}
\title{
\raisebox{-0.20\height}{\includegraphics[width=0.09\textwidth]{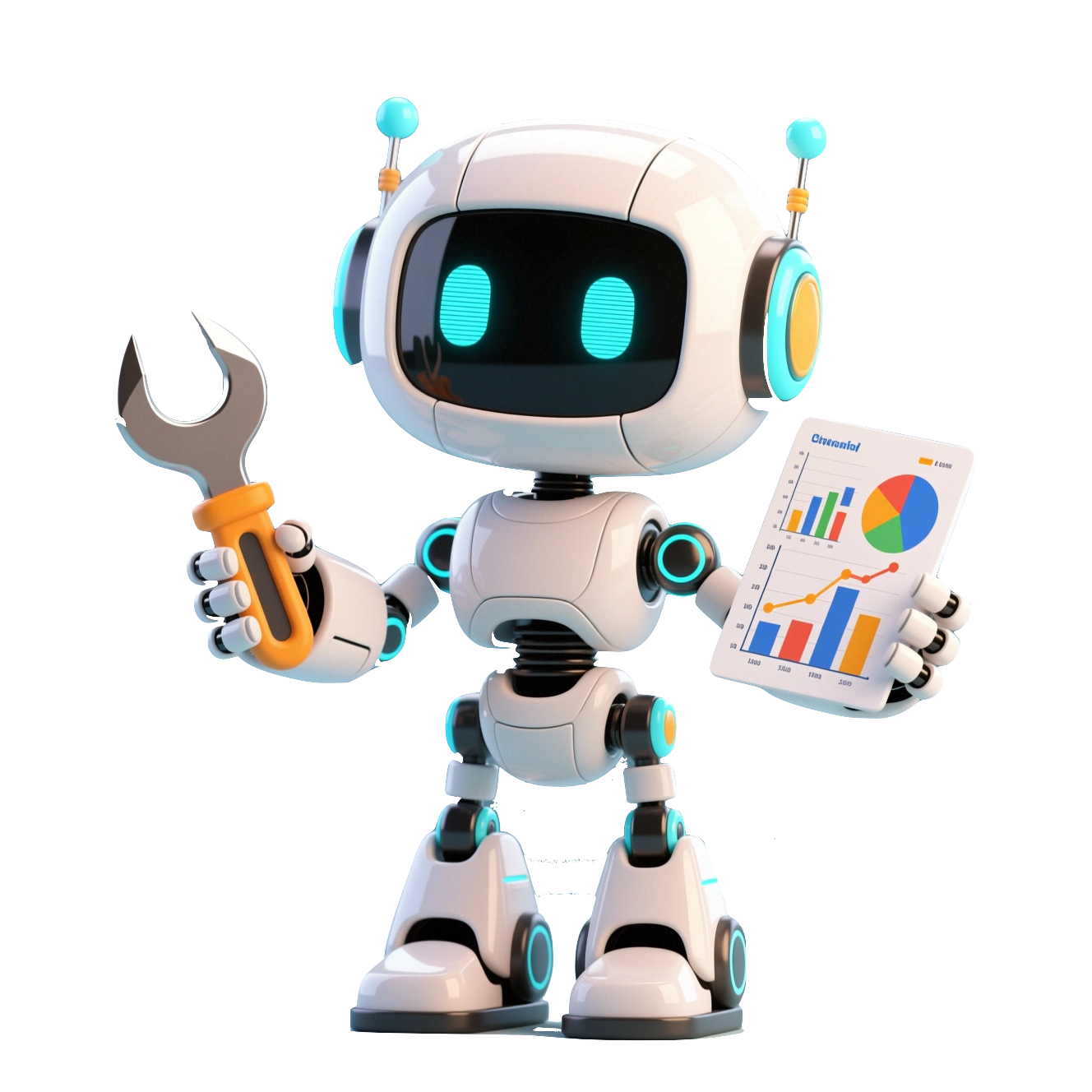}}
LLM/Agent-as-Data-Analyst: A Survey
}

\pagestyle{plain}
\pagenumbering{arabic}

\renewcommand\thesection{\arabic{section}}
\setcounter{section}{0}

\pagenumbering{arabic}
\setcounter{page}{1}
\setcounter{figure}{1}
\setcounter{table}{0}

\author{
    \IEEEauthorblockN{
    Zirui Tang\IEEEauthorrefmark{1}\IEEEauthorrefmark{5},
    Weizheng Wang\IEEEauthorrefmark{1}\IEEEauthorrefmark{5},
    Zihang Zhou\IEEEauthorrefmark{1},
    Yang Jiao\IEEEauthorrefmark{1},
    Bangrui Xu\IEEEauthorrefmark{1},
    Boyu Niu\IEEEauthorrefmark{1},
    Dayou Zhou\IEEEauthorrefmark{1},
    \\
    Xuanhe Zhou\IEEEauthorrefmark{1}\Letter,
    Guoliang Li\IEEEauthorrefmark{2},
    Yeye He\IEEEauthorrefmark{3},
    Wei Zhou\IEEEauthorrefmark{1},
    Yitong Song\IEEEauthorrefmark{1},
    Cheng Tan\IEEEauthorrefmark{4},
    Xue Yang\IEEEauthorrefmark{1},
    \\
    Chunwei Liu\IEEEauthorrefmark{9},
    Bin Wang\IEEEauthorrefmark{4},
    Conghui He\IEEEauthorrefmark{4},
    Xiaoyang Wang\IEEEauthorrefmark{6},
    Fan Wu\IEEEauthorrefmark{1}
    }
    \\
    \IEEEauthorblockA{\IEEEauthorrefmark{1}Shanghai Jiao Tong University}
    \IEEEauthorblockA{\IEEEauthorrefmark{2}Tsinghua University}
    \IEEEauthorblockA{\IEEEauthorrefmark{3}Microsoft Research}
    \\
    \IEEEauthorblockA{\IEEEauthorrefmark{9}MIT CSAIL}
    \IEEEauthorblockA{\IEEEauthorrefmark{4}Shanghai AI Laboratory}
    \IEEEauthorblockA{\IEEEauthorrefmark{6}Fudan University}

    \textcolor{blue}{\href{https://github.com/weAIDB/awesome-data-llm}{https://github.com/weAIDB/awesome-data-llm}}
    
\IEEEcompsocitemizethanks{
\protect \item
\IEEEauthorrefmark{5} Co-first authors with equal contributions.

\protect \item
\Letter\  Corresponding author.
}
}

\IEEEtitleabstractindextext{
{ 
\leftskip=0pt \rightskip=0pt plus 0cm
\input{sections/abstract}
}
\begin{IEEEkeywords}
LLM, Agent, Data Analysis, Structured Data, Semi-Structured Data, Unstructured Data, Heterogeneous Data
\end{IEEEkeywords}
}

\maketitle

\input{sections/introduction}

\input{sections/structured-data}

\input{sections/semistructured-data}
\input{sections/unstructured-data}
\input{sections/heterogeneous-data}
\input{sections/industry}
\input{sections/future-work}

\input{sections/conclusion.tex}

\balance
\scriptsize
\bibliographystyle{abbrv}
\bibliography{ref/DA}





\end{document}

%% file: sections/abstract.tex

\begin{abstract}

Large language models (LLMs) and agent techniques have brought a fundamental shift in the functionality and development paradigm of data analysis tasks (a.k.a. \llmasanalyst), demonstrating substantial impact across both academia and industry. In comparison with traditional rule or small-model based approaches, (agentic) LLMs enable complex data understanding, natural language interfaces, semantic analysis functions, and autonomous pipeline orchestration. 
From a modality perspective, we review LLM-based techniques for $(i)$ structured data (e.g., NL2SQL, NL2GQL, ModelQA), $(ii)$ semi-structured data (e.g., markup languages understanding, semi-structured table question answering), $(iii)$ unstructured data (e.g., chart understanding, text/image document understanding), and $(iv)$ heterogeneous data (e.g., data retrieval and modality alignment in data lakes). The technical evolution further distills four key design goals for \textit{intelligent data analysis agents}, namely semantic-aware design, autonomous pipelines, tool-augmented workflows, and support for open-world tasks.
Finally, we outline the remaining challenges and propose several insights and practical directions for advancing LLM/Agent-powered data analysis.

\end{abstract}


%% file: sections/introduction.tex


\begin{figure*}[!t]
    \centering
    \includegraphics[width=1\linewidth]{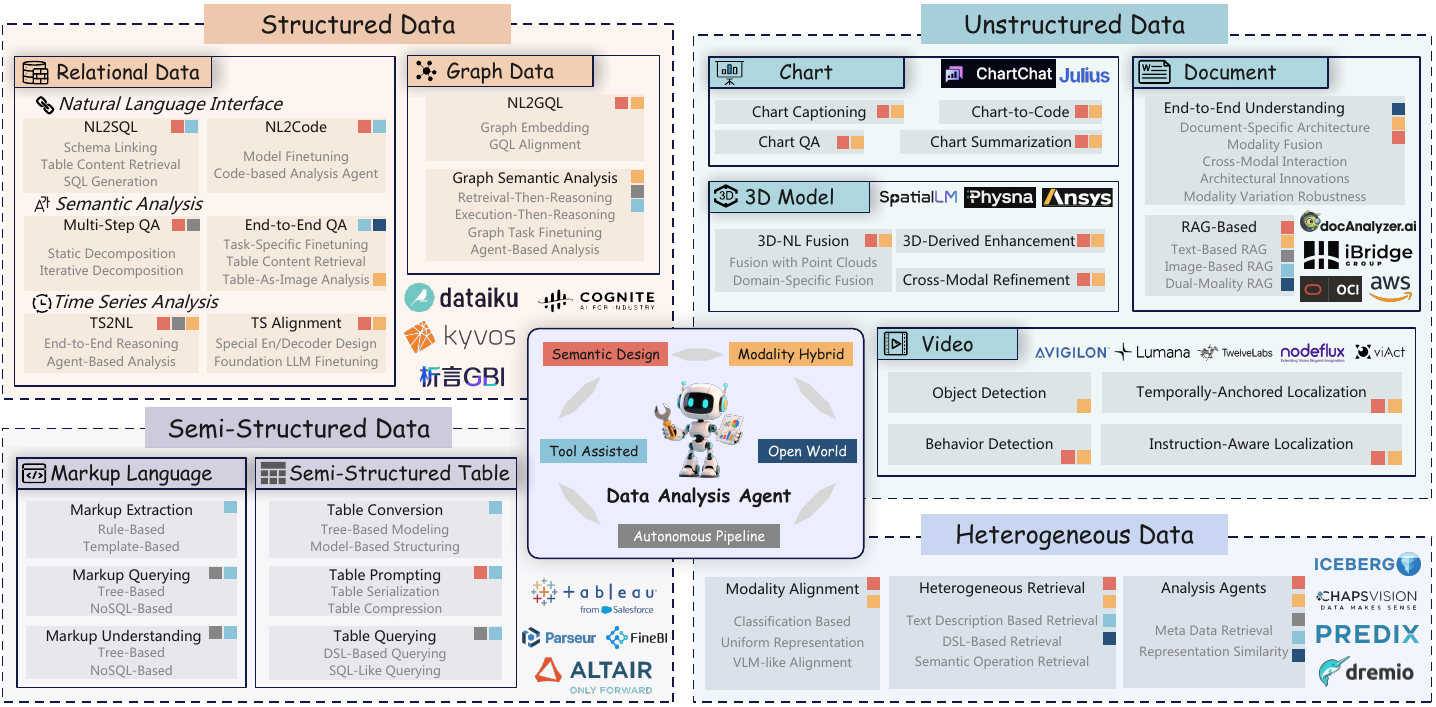}
    \caption{Technical Overview of \llmasanalyst. The five key design goals are illustrated in the center of the figure using distinct colors. The colored icons next to each technique indicate the specific design goal it supports.}
    \label{fig:content-overview}
\end{figure*}

\section{INTRODUCTION}
\label{sec:introducion}




\lettrine{D}{ata} analysis, broadly defined as the process of inspecting, transforming, and modeling data to discover useful information and support decision-making, constitutes a cornerstone of modern scientific research and business intelligence~\cite{hardy2004handbook,temporal2022,LLM4DM,chat2data2024}. 
It spans a wide spectrum of data modalities, from structured databases and semi-structured tables to unstructured documents and videos, underpinning critical applications across domains such as finance~\cite{cao2022ai}, healthcare~\cite{batko2022use}, and social sciences~\cite{nooralahzadeh2024explainable}.

\subsection{Limitations of Conventional Data Analysis}

Traditional data analysis pipelines, although effective in extracting information and statistical patterns, often demand substantial domain expertise, manual feature engineering, and the integration of multiple specialized tools~\cite{xiao2024cellagent}. These limitations become more severe as data grows in scale, complexity, and heterogeneity~\cite{liang2024natnl2gql}, ultimately constituting the inherent weakness of conventional data analysis.

\newcommand{\tightcircle}[1]{%
  \tikz[baseline=(c.base)]%
    \node[circle,draw,inner sep=0.5pt,line width=0.3pt]
         (c){{\small #1}};%
}

\noindent{\bf {\small \colorbox{gray!10}{L1: Closed-World Assumption.}}} Current data analysis systems are generally optimized for specific data types, with architectures and query engines tailored to particular formats. For example, relational databases are well-suited for structured data~\cite{LLMQO}, while systems such as MongoDB~\cite{mongodb} are designed for semi-structured formats like JSON. When analyzing videos with accompanying timestamped descriptions, human analysts must manually align frames with their corresponding documents to enable cross-modal analysis~\cite{mu2024snagscalableaccuratevideo}. This specialization hampers the analysis of heterogeneous data across modalities, thereby increasing analytical complexity, introducing errors, and limiting both the scope and efficiency of data-driven insights.

\noindent{\bf {\small \colorbox{gray!10}{L2: Rigid Tool Coupling.}}} One observation is that existing data analysis systems often come with a suite of tightly coupled built-in tools (e.g., visualization dashboards, statistical packages, or machine learning modules~\cite{torchptpth}) tailored to specific workflows, which stems from the inherent complexity of the underlying data types. For instance, to extract statistical information from a knowledge graph and subsequently model it with linear regression, a human analyst must first formulate GQL queries and invoke the built-in APIs of graph databases to obtain the data, and then employ machine learning libraries (e.g., scikit-learn, PyTorch) for further analysis.  In contrast, analyzing 3D models relies more heavily on domain-specific expertise and specialized systems (e.g., AutoCAD, PyMOL~\cite{pymol}). Such analyses are often constrained to the functionalities embedded in these systems, requiring analysts, particularly those without coding expertise, to integrate multiple systems in order to complete a single task. The integration of heterogeneous systems not only increases the complexity and overhead of the analysis process, but also hampers extensibility and complicates integration into broader analytical workflows. 

\noindent{\bf {\small \colorbox{gray!10}{L3: Basic Literal Analysis.}}} While conventional analysis methods effectively support operations such as filtering and aggregation, they operate primarily at the literal level and lack the ability to reason over the semantic content embedded within data. For example, in table analysis, when a cell contains lengthy textual descriptions, SQL queries can only retrieve or match keywords, without interpreting deeper meanings such as sentiment, intent, or causal relationships~\cite{zhang2023reactable}. Similarly, for unstructured data types like charts or documents, analysts are often required to manually interpret and summarize the content to meet analytical goals~\cite{han2023chartllama}. This reliance on surface-level processing limits the analytical depth of current systems and hinders their capacity to handle complex, knowledge-intensive tasks that require semantic reasoning, abstraction, or contextual understanding beyond mere syntactic patterns.

\noindent{\bf {\small \colorbox{gray!10}{L4: Manual-Based Development.}}} Manual efforts in data analysis pose significant bottlenecks in both workflow construction and system design.

\noindent{\it (1) Static Analysis Workflow Development.} A major source of labor intensity lies in the reliance on rigid, manually crafted analytical pipelines, which require domain experts to predefine modeling steps~\cite{zhou2025cracksql,zhou2025cracksqldemo,zhang2023reactable}. In large-scale enterprise databases, where data is distributed across separate tables, even a simple business question such as {\it ``What is the average delivery time for premium customers in the last quarter?''} demands a series of complex manual operations. Analysts must identify the relevant tables, select appropriate attributes, compose multi-table joins, impose domain-specific constraints, and align timestamp semantics. Similarly, in document-centric analytics, retrieving query-relevant content from lengthy, heterogeneous reports (comprising text, tables, charts, and images) often involves sequentially reading through pages and manually synthesizing information, which is both time-consuming and error-prone.

\noindent{\it (2) Manual System Design and Engineering.} Beyond workflow construction, existing data analysis systems themselves are heavily dependent on manual engineering, from designing features and crafting rule sets to integrating models, resulting in high development overhead, limited scalability, and poor adaptability to new analysis tasks or data modalities. Such reliance on static, human-defined logic restricts the ability of systems to generalize across use cases and hinders progress toward autonomous, self-evolving analytical tools.

\subsection{Opportunities by LLM/Agent-As-Data-Analyst}

Recent advances in large language models (LLMs) and LLM-based agents have introduced new opportunities to alleviate these challenges. As shown in Figure 1, by enabling auto-designed analysis pipelines, adaptive tool-assisted workflows, and natural language interaction, LLMs hold the potential to lower technical barriers, enhance interpretability, and accelerate the discovery of actionable insights from diverse forms of data.

\noindent{\bf {\small \colorbox{gray!10}{O1: Cross-Modal Open World Data Support.}}}
LLMs are capable of processing and reasoning across complex data, including relational data, semi-structured tables, and unstructured texts, owing to their ability to capture latent patterns and contextual dependencies~\cite{su2024tablegpt2,tang2024graphgpt}. For example, they cannot only comprehend the semantic content associated with nodes and edges but can also reason about the underlying structural properties (e.g., connectivity, community patterns, or hierarchical organizations). Such a deep and holistic understanding empowers analysts to derive richer insights that integrate quantitative measures with qualitative nuances~\cite{arora2023language}.


\noindent{\bf {\small \colorbox{gray!10}{O2: NL-based Tool Agnostic Interaction.}}}LLM-powered analysis agents employ natural language as the primary interface for interaction, allowing users to express analytical intents without relying on specialized query languages such as SQL or advanced programming skills~\cite{liang2024natnl2gql,birdsql}. By abstracting away tool-specific interfaces and technical barriers, this natural language interaction paradigm enhances usability, streamlines the analysis process, and significantly lowers the entry threshold for non-expert users. As a result, it broadens participation in data-driven decision-making and facilitates more inclusive access to complex analytical capabilities across diverse user groups~\cite{tang2025st,han2023chartllama}.

\noindent{\bf {\small \colorbox{gray!10}{O3: Semantic Operators.}}}
LLMs empower a new class of semantic-level operators, such as semantic aggregation, semantic filtering, and semantic joins, that go beyond syntactic matching to interpret and manipulate data based on latent meaning~\cite{wang2024interactive}. These operators are particularly beneficial when working with data that contains rich textual descriptions or complex relationships. For example, in a medical records database, a conventional query might aggregate all entries containing the term ``fever'', whereas a semantic aggregation operator can group or filter entries based on synonymous or contextually related expressions such as ``high temperature'' or ``febrile condition''. Similarly, semantic joins allow for linking records across tables even when the join keys are not exact matches but share implicit meaning (e.g., ``Type II Diabetes'' and ``adult-onset diabetes'')~\cite{jiang2025explainable,wang2024chainoftable}. By embedding these capabilities directly into the query execution layer, LLMs significantly enhance the expressiveness and flexibility of analytical workflows.

\noindent{\bf {\small \colorbox{gray!10}{O4: Automatic Workflow Orchestration $\&$ Evolution.}}}
First, LLM-powered analysis agents offer the potential for autonomous orchestration of analytical workflows by dynamically selecting appropriate data sources, operations, and tools based on user intent and data context. Unlike traditional static pipelines, such agents can reason about task requirements and adaptively construct end-to-end workflows, reducing the burden on analysts to manually specify each step. Beyond orchestration, recent research also explores the possibility of autonomous system evolution, such as automatically designing prompts, selecting tools, or even generating new agentic modules tailored to specific tasks. While these capabilities remain underexplored, they represent a promising future direction toward self-improving, minimally supervised analytical systems.

\subsection{Techniques of LLM-Powered Data Analysis}

As shown in Figure 1 and Figure~\ref{fig:content-overview}, given the diversity of data formats and application scenarios, we establish a taxonomy of LLM-powered data analysis which progresses along two dimensions: $(i)$ the range of {\bf data modalities} supported, i.e., structured, semi-structured, unstructured, and heterogeneous (the x-axis), and $(ii)$ the evolution of {\bf interaction paradigms}, i.e., code-based, domain-specific language (DSL)-based, and natural language (NL)-based  (the y-axis). 


\noindent\textbf{LLM/Agent for Structured Data Analysis.}
Structured data, including relational databases~\cite{relationaldata} and graphs~\cite{graphdata}, remains central in industry for its standardized schema and well-defined semantics. Traditional methods rely on SQL or related query languages, later extended with DSLs for domain-specific tasks~\cite{neo4j}.
With LLMs, users can interact with structured data via natural language through code generation, DSL mapping, or LLM-based question answering~\cite{sqlsurvey,shi2024sqlsurvey}, while agent frameworks further orchestrate multi-step analytical workflows~\cite{zhu2024sqlsurvey}.
The core of LLM-driven structured data analysis lies in employing LLMs for pipeline automation or end-to-end processing.

\noindent$\bullet$ \textit{\underline{Relational Data.}}
Relational data is typically stored and managed in specialized databases.
An LLM-based approach to relational data analysis leverages the model as a natural language interface that translates analytical intents into SQL or executable code, which is then used to query the database. To enhance alignment between natural and formal languages, techniques such as schema linking~\cite{zhang2024finsql}, information retrieval~\cite{DINSQL}, and task decomposition~\cite{hong2024datainterpreter} are integrated into the pipeline, while task-specific fine-tuning~\cite{yin2022code} improves end-to-end generation.
An alternative direction bypasses direct database manipulation, employing LLMs for semantic-level analysis. This requires deeper understanding of analytical intent and intra-table relationships. Accordingly, RAG~\cite{patnaik2024cabinet}, prompt engineering~\cite{tatllm}, and task decomposition~\cite{zhao2024tapera} are incorporated to enhance reasoning, while MLLMs~\cite{zheng2024multimodal} or LLMs~\cite{su2024tablegpt2} are trained or fine-tuned on textual or visual table representations.
Time-series data, a specialized subclass of relational data characterized by temporal dependencies, can similarly benefit from LLM-powered techniques, including natural language-to-code translation~\cite{DBLP:conf/iclr/0005WMCZSCLLPW24}, sequence retrieval and transformation, and direct temporal reasoning~\cite{alnegheimish2024can}.

\noindent$\bullet$ \textit{\underline{Graph Data.}}
Graph data represents entities and their inter-dependencies to model complex network semantics, posing challenges due to its vast search space and intricate path reasoning.
Specialized graph databases and query languages~\cite{graphdb,blazegraph} enable direct access, while LLMs can serve as natural language interfaces for generating graph queries. Techniques such as agent-based frameworks~\cite{liang2024natnl2gql}, fine-tuning, and prompt engineering~\cite{liang2024aligning} enhance the model’s understanding of graph structures and query syntax.
For semantic-level graph analysis, mainstream approaches leverage concepts from RAG~\cite{zhang2022subgraph}, agent-based reasoning~\cite{luo2025kbqa}, and fine-tuning~\cite{ye2024instructglm} to improve graph-aware comprehension and reasoning.

\noindent\textbf{LLM/Agent for Semi-structured Data Analysis.}
Semi-structured data lies between unstructured text and fully structured relational databases. It typically contains some organizational structure but does not conform to rigid schemas.

\noindent$\bullet$ \textit{\underline{Markup Language.}} 
XML, JSON, and HTML are common forms of markup language data. As markup languages consist of both tags and content, and inherently exhibit structural properties, their evolution bears strong similarity to that of semi-structured tables, which has also motivated the development of structure-aware PLMs~\cite{arora2023language,wang2022webformer}.

\noindent$\bullet$ \textit{\underline{Semi-Structured Tables}} represent a flexible form of tabular data whose evolution is closely tied to advances in structured table understanding and the semantic capabilities of LLMs. Traditional approaches, often based on PLMs~\cite{herzig2020tapas,RoBERTa}, struggle to capture the complex layouts, irregular headers, and hierarchical structures due to their limited ability to encode semantics. With the advent of LLMs, new paradigms have emerged for handling such tables, including semi-structured to structured transformations that convert semi-structured tables into relational-like forms~\cite{10.1145/3726302.3730071,li2023autotablessynthesizingmultisteptransformations}, and DSL-based modeling that leverages domain-specific languages to explicitly encode structure and operations~\cite{tang2025st}.

\noindent\textbf{LLM/Agent for Unstructured Data Analysis.}
Unstructured data encompass a wide range of modalities such as charts, videos, documents, and 3D models, which lack a fixed schema and therefore pose significant challenges for conventional analysis pipelines.

\noindent$\bullet$ \textit{\underline{Chart.}}
Traditional chart analysis systems typically rely on handcrafted feature extraction, template matching, or rule-based parsing techniques~\cite{kafle2018dvqaunderstandingdatavisualizations,mittal-etal-1998-describing,reiter-2007-architecture}, which often struggle with variations in design, layout, and data representation. With the advent of LLMs, we can leverage multimodal understanding capabilities to interpret visual and structural elements of charts~\cite{masry2023unichart}, generate contextual descriptions, perform semantic data extraction, and support natural language-based reasoning~\cite{masry2023unichart}, captioning~\cite{liu2024chartthinker,obeid-hoque-2020-chart,singh2025figcapshffiguretocaptiongenerativeframework} and QA~\cite{wu2024chartinsights,das2025chartsofthoughtenhancingllmvisualization,xu2025chartmoemixturediverselyaligned} tasks over different graphical data.

\noindent$\bullet$ \textit{\underline{Video.}}
Video encodes spatial content that evolves over time, requiring joint modeling of semantics and dynamics. 
Conventional methods rely on vision backbones with temporal pooling or attention, which struggle with annotation costs and long-sequence efficiency \cite{chen2024timemarker} \cite{wang2024grounded}. 
Recent advances reformulate videos as structured token sequences for LLM reasoning. Agents further decompose queries into temporal grounding, multimodal fusion, and summarization, enabling richer temporal understanding and efficient computation \cite{chen2024timemarker} \cite{deng2025seq2time}. 
Building upon this, LLMs have also been extended to {Video Emotional Analysis} , where multimodal fusion of visual, acoustic, and textual cues enables affective state inference.In addition, posture-based emotion analysis and 3D mesh reconstruction support nuanced social interaction modeling and relational emotional predictions~\cite{jadhav2023ai,muller2024predicting}. For {Object Detection}, integrating high-precision detectors  with multimodal reasoning allows object-centric summarization and reference grounding in videos to make correct object detections in videos~\cite{de2024video,yuan2025videorefer}. Furthermore, {Gesture and Behavior Detection} leverages LLM-driven pipelines to extract fine-grained motion features and gestures, supporting interaction analysis and embodied behavior reasoning, often via 3D reconstructions for higher fidelity \cite{whitehead2025utilizing,weng2025artificial}.

\noindent$\bullet$ \textit{\underline{Document.}}
Documents, such as PDFs, web pages, and scanned reports, serve as primary carriers of information in business and academia. Traditional document analysis relies on Optical Character Recognition (OCR) and rule-based template matching~\cite{dococr,ha2022information}, which are often ineffective for complex or varied layouts and fail to grasp the deep semantics of the content. LLMs, particularly Multimodal LLMs, have revolutionized document understanding by unifying the processing of text, layout, and visual information in three aspects: {$(i)$ Architectural Design} from the LayoutLM~\cite{xu2020layoutlm,xu2021layoutlmv2, huang2022layoutlmv3} series to DocLLM~\cite{wang2024docllm}, enable a synergistic understanding of document structure and content. Concurrently, {$(ii)$ Visual-Based Understanding} is employed for question answering and summarization across documents with diverse layouts and elements~\cite{lewis2020retrieval, lei2024visdommultimodal}, and {$(iii)$ Text-Based Understanding} effectively solves the document understanding on long and text-rich documents with techniques like RAG~\cite{lin2024zendb, sun2025quest}. These techniques collectively drive the shift from simple information extraction to deep document reasoning.

\noindent$\bullet$ \textit{\underline{3D Model.}} 3D models represent objects or scenes in Euclidean space, expressed as point clouds, meshes, or voxels, and are widely applied in scene understanding and scientific analysis. Conventional pipelines rely on geometric processing (e.g., mesh simplification, point cloud registration) \cite{hoppe1997view,garland1997surface,chen1992object} and modeling software such as Blender or Maya \cite{blain2019complete,tickoo2018autodesk}, which require manual annotation and lack semantic understanding. Recent LLM-based approaches  enable 3D–language alignment~\cite{hong20233d}, where geometry is transformed into structured embeddings or textual descriptions for reasoning. Agents orchestrate specialized 3D encoders and toolchains to support downstream tasks such as captioning, navigation, and scientific QA \cite{hong20233d,xiong20253ur}. Building on this foundation, {3D-Language Fusion} provides  unified frameworks that map point clouds and meshes into embeddings aligned with natural language, enabling tasks such as captioning and QA 3D-LLM~\cite{hong20233d}, 3UR-LLM~\cite{xiong20253ur} Domain-specific extensions integrate molecular and protein structures into multimodal reasoning, as in 3D-MoLM~\cite{li2024towards}, ProteinChat~\cite{guo2023proteinchat}, and ProtChatGPT~\cite{wang2024protchatgpt}. Moreover, {3D-Derived Task Enhancement} leverages multi-agent systems leverage textual mediation of geometry for scene description, navigation, and retrieval, improving interpretability and efficiency~\cite{hong20233d,xiong20253ur,li2024towards}. Finally, {Cross-Modal Refinement} employs feature enhancement and domain adaptation techniques(e.g., visual grounding, 2D–3D alignment, or joint pretraining) to bridge 2D and 3D modalities, strengthening generalization in multimodal LLMs~\cite{hong20233d,li2024towards,li2024towards,xiong20253ur}.


\noindent\textbf{LLM/Agent for Heterogeneous Data Analysis.}
Heterogeneous data refers to the integration of diverse data types (e.g., relational data, semi-structured tables, document images)~\cite{wang2017heterogeneous}.
Early research focused on heterogeneous data management~\cite{iceberg}, which supports data retrieval through SQL-like languages.
More recent advances with LLMs address three main directions: $(i)$ modality alignment across data types~\cite{Unicorn, chen2023symphony} (e.g., leveraging natural language descriptions to compute cross-modal similarity), $(ii)$ natural language interfaces for heterogeneous data retrieval~\cite{patel2024lotus,wang2024must} (e.g., translating user queries into sequences of predefined APIs), and $(iii)$ heterogeneous data analysis agents~\cite{nooralahzadeh2024explainable,wang2024interactive} (e.g., equipping LLMs with semantic operation tools tailored to different modalities).

\noindent\textbf{LLM/Agent for Data Analysis Evolution.}
As shown in Figure 1, the technological evolution of LLM-powered Data Analysis Agents can be summarized along four key dimensions, each corresponding to the design goals of a unified data analysis agent.

\noindent$\bullet$ \textit{\underline{From Closed World to Open World.}} Initially, agents are predominantly tailored to domain-specific tasks and types(e.g., financial or industry analysis). The trend is toward general-purpose agents capable of analyzing diverse and real-world data (e.g., documents, videos), lowering user entry barriers.

\noindent$\bullet$ \textit{\underline{From Rigid Tool Coupling to Flexible Tools.}} Traditional agents rely on tightly integrated, framework-specific tools. The evolution favors decoupled architectures, where LLMs can leverage arbitrary toolsets, enhancing flexibility and adaptability in diverse analytical contexts.

\noindent$\bullet$ \textit{\underline{From Basic Literal Analysis to Semantic Supported Analysis.}} 
Conventional analysis methods primarily operate at the literal level and lack the ability to reason over the semantic content embedded within data.
The evolution has shifted toward leveraging semantic reasoning, abstraction, or contextual understanding to derive analytical results.

\noindent$\bullet$ \textit{\underline{From Manual Development to Fully Autonomous Design.}} Early agents require manual decomposition of analytical workflows (e.g., task decomposition, code generation, operation execution). Modern agents increasingly support both autonomous analysis workflow design and analysis system design, with broader operational capabilities and reduced human intervention.



\subsection{Comparison \& Contributions}

Compared with existing surveys on LLMs for data analysis~\cite{sqlsurvey,shi2024sqlsurvey,zhu2024sqlsurvey,shang2024graphsurvey,ren2024graphsurvey,jin2022tablesurvey,wu2025tablesurvey,ding2024docsurvey,barboule2025docsurvey,farahani2023chartsurvey,al2024chartsurvey,tang2025videosurvey}, our work provides a more comprehensive and detailed overview of the key techniques applied across different data types, while also emphasizing the interrelationships among these data types. We uniquely examine the development trends from the perspectives of data modalities and interface languages, and further outline key dimensions for the design of a general LLM-based data analysis agent.

\begin{sloppypar}
\noindent$\bullet$ \underline{\textit{{\bf Full-Vision} Introduction to Typical Data Analysis Tasks.}} Different from existing surveys that typically focus on a single modality or task (e.g., NL2SQL~\cite{sqlsurvey,shi2024sqlsurvey,zhu2024sqlsurvey}, graph understanding~\cite{shang2024graphsurvey,ren2024graphsurvey}, table-QA~\cite{jin2022tablesurvey,wu2025tablesurvey}, document understanding~\cite{ding2024docsurvey,barboule2025docsurvey}, chart understanding~\cite{farahani2023chartsurvey,al2024chartsurvey}, video understanding~\cite{tang2025videosurvey}), we systematically organize the landscape of data analysis by categorizing techniques across structured, semi-structured, unstructured, and heterogeneous data. This roadmap also traces the technical evolution of LLM-powered data analysis and identifies five key dimensions that serve as design goals for a general-purpose data analysis agent.
\end{sloppypar}

\noindent$\bullet$ \textit{\underline{Detailed Review of Data Analysis Techniques.}}
Beyond high-level summaries, we provide an in-depth examination of representative methods, discussing their underlying principles, technical designs, and application scenarios. In contrast to existing surveys, we further emphasize the critical role of data curation methods tailored to specific downstream tasks and offer corresponding data analysis insights.

\noindent$\bullet$ \textit{\underline{Recent Advances in \llmasanalyst.}}
In addition to established techniques, we emphasize the latest developments that leverage LLM for data analysis (e.g., agentic design, multimodal alignment, interaction techniques). 
By incorporating these cutting-edge advances, our survey provides an up-to-date reference for researchers and practitioners seeking to understand the state-of-the-art.

\noindent$\bullet$ \textit{\underline{Open Challenges and Future Directions.}} 
We identify the key technical and practical challenges that remain unresolved, such as scalability, evaluation, and integration into real-world systems. Building on these insights, we also outline promising future research directions to guide the development of general-purpose LLM-based data analysis agents.

\subsection{Organization of Our Survey}

Section \ref{sec:sturctured-data} discusses LLMs for structured data analysis, covering relational data (Section~\ref{sec:relational}) and graph data (Section~\ref{sec:graph}). 
Section \ref{sec:semistructured-data} reviews LLMs for semi-structured data analysis, including markup languages (Section~\ref{sec:markup}) and semi-structured tables (Section~\ref{sec:semitable}). 
Section \ref{sec:unstructured-data} examines LLMs for unstructured data analysis, encompassing charts (Section~\ref{sec:chart}), videos (Section~\ref{sec:video}), documents (Section~\ref{sec:document}), and 3D models (Section~\ref{sec:3d}). 
Section \ref{sec:heterogeneous-data} addresses LLMs for heterogeneous data analysis.  
For each data type, we first present the data analysis techniques, followed by a subsection on data curation. 
Section \ref{sec:industry_practice} introduces some industry practices of LLM-powered data analysis.
Finally, we discuss the challenges and future directions associated with each data type in Section~\ref{sec:future-work}, and conclude the survey in Section~\ref{sec:conclusion}.

%% file: sections/structured-data.tex
\section{\llm for Structured Data Analysis} \label{sec:sturctured-data}

Structured data refers to data with well-defined schemas like relational data~\cite{relationaldata} and graph data~\cite{graphdata}. Both of them share pre-defined patterns that enable clear organization and efficient querying. Relational data is characterized by row and column tables, or associated keys. Graph data models entities and their relationships through nodes and edges. 


\subsection{Relational Data Analysis} 
\label{sec:relational}


\hi{LLM for Natural Language Interfaces.}
Relational data analysis typically involves well-defined operations such as aggregation (e.g., summation, averaging, ranking)~\cite{relationaldata}, statistical modeling (e.g., regression, clustering)~\cite{hoff2015multilinear,taskar2001probabilistic}, and data quality assurance (e.g., constraint validation, outlier detection)~\cite{maervoet2012outlier}, often supported by SQL or Python libraries like Pandas.
While traditional methods depend on rigid SQL syntax and predefined analytical rules, LLMs extend these capabilities by enabling natural language to SQL/code translation and advanced semantic reasoning through their strong language understanding and inference abilities.


\begin{figure}[!t]
    \centering
    \includegraphics[width=1\linewidth]{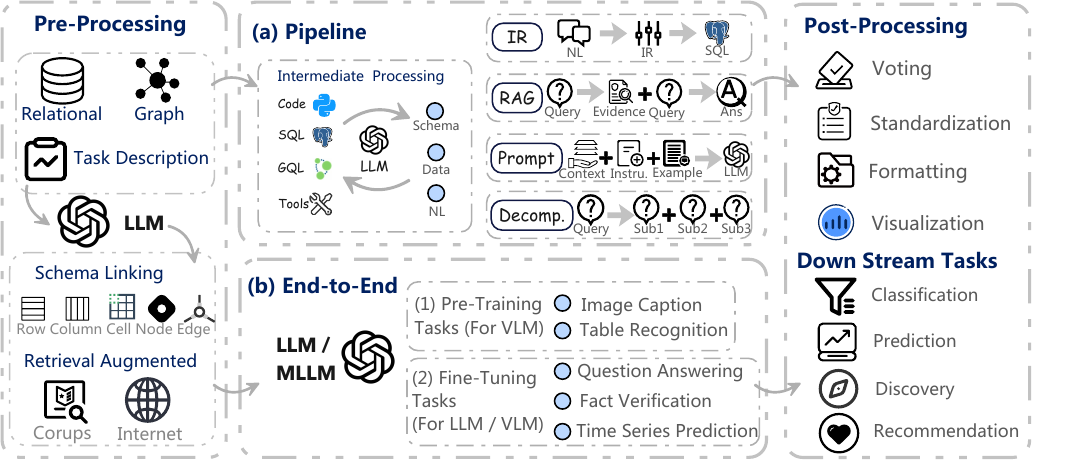}
    \caption{\textbf{LLM for Structured Data Analysis} - (a) Pipeline Method. (b) End-to-End Method.}
    \label{fig:frameworkLLM4Data}
\end{figure}

\noindent$\bullet$ \textit{\underline{NL2SQL.}} 
NL2SQL focuses on translating natural language queries into SQL commands by leveraging techniques such as $(i)$ schema linking, which aligns user intents with database schema to resolve ambiguities and filters out irrelevant schema to enhance the conciseness of input~\cite{zhang2024finsql,petsql}, $(ii)$ content retrieval, which dynamically extracts relevant information from the database to refine query generation~\cite{talaei2024chess,li2024codes}, and $(iii)$ SQL generation strategies such as multi-step generation that concentrates on different SQL parts at each step~\cite{fan2024combining, zhou2025cracksql, zhou2025cracksqldemo}, intermediate SQL representation or hybrid table representation for better \llm 's comprehension~\cite{DINSQL, DBLP:journals/pacmmod/XieXZG25, long2025bridging}, and different decoding strategies {(e.g., beam search, greedy search, or PICARD~\cite{SQL360})}. {Other surveys provide more comprehensive overviews for NL2SQL techniques~\cite{Liu2024ASO, luo2025natural, shi2024survey}.}

\noindent$\bullet$ \textit{\underline{NL2Code.}} Unlike NL2SQL, NL2Code approaches focus on enhancing relational data analysis by generating Python code (e.g., using Pandas or NumPy). These methods must cope with a vast number of library APIs that are highly variable and complex, often involving intricate chained operations.
{Recent studies have made progress in alleviating these challenges, though many remain open.}

{\noindent\underline{(1) Model Fine-Tuning:}  
PACHINCO~\cite{yin2022code} fine-tunes a 62B parameter PALM~\cite{chowdhery2022palm} model in two stages (i.e., separately using  a Python source code corpus with 64B tokens and a Jupyter notebook corpus with 9.6B tokens) so as to improve model performance on analysis-related tasks (e.g., calculate the amount of games added in each year for each month). 
DataCoder~\cite{huang2024datacoder} utilizes different types of contexts (e.g., code, text, and data) by employing dual encoders (e.g., data encoder and code + text encoder) and one general decoder to generate code in notebooks.}

\noindent\underline{(2) LLM-Based Analysis Agent:}  
Data Interpreter~\cite{hong2024datainterpreter} employs LLMs via APIs to generate task and action graphs , leveraging their semantic reasoning to decompose complex user queries (e.g., correlation analysis, data exploration, anomaly detection) into subproblems and iteratively refine and verify each to improve code generation for data science tasks.
Another approach~\cite{bai2025collaboration} fine-tunes a BART model~\cite{DBLP:conf/acl/LewisLGGMLSZ20} as an intelligent agent using datasets derived from task-instruction and role-definition prompts. The agent produces detailed prompts to guide LLMs through stepwise NL2Code execution, including role definition, requirement optimization, code writing, and code review.


\hi{LLM for Semantic Analysis.} Moreover, tasks that demand semantic understanding or natural language generation (e.g., table summarization) also benefit from LLM-based methods, including $(i)$ multi-step question answering with decomposition strategies and $(ii)$ end-to-end QA using optimized LLMs.


\noindent$\bullet$ \textit{\underline{Multi-Step QA.}} Multi-step question answering decomposes complex queries into sequential sub-questions for stepwise reasoning. Existing methods include $(i)$ static decomposition, following fixed processing steps (e.g., retrieve–select–reason), and $(ii)$ LLM-driven iterative decomposition, where the LLM dynamically determines each step based on contextual reasoning history.


\noindent\underline{(1) Static Decomposition:} 
Static decomposition frameworks, such as Retriever–Selector–Reasoner and its variants, divide tasks into modular components to support multi-step inference and improve interpretability. The Extractor–Reasoner–Executor paradigm~\cite{tatllm} extracts relevant context, generates logic rules or equations, and executes them via LLM prompting to get the final answer. Similarly, S3HQA~\cite{lei2023s3hqa} uses a retriever to filter heterogeneous resources, a selector to identify relevant knowledge, and a generation-based reasoner to produce the final answer.
For multi-table scenarios, SGAM~\cite{wang2025plugging} encodes schema links and join paths as a graph, converts queries into reasoning chains on the graph, and executes them to produce the final answer.

\noindent\underline{(2) Iterative Decomposition:} 
Static decomposition often struggles with multi-hop queries, whereas LLM-driven iterative decomposition dynamically overcomes it by refining sub-tasks via recursive reasoning.
TAPERA~\cite{zhao2024tapera} applies this approach by introducing query decomposition, where the Planner breaks the query into sub-queries, the Reasoner generates executable programs for each, and the Answer Generator derives results to fulfill the plan. Finally, the Planner updates or finalizes the plan as needed.
Similarly, ReAcTable~\cite{zhang2023reactable} and CHAIN-OF TABLE~\cite{wang2024chainoftable} iteratively generate operations and update tables, using LLMs and in-context learning to produce reasoning chains as proxies for intermediate thoughts.



\noindent$\bullet$ \textit{\underline{End-to-End QA.}} End-to-End Question Answering refers to methods where the LLM directly generates the final answer without intermediate steps or iterative refinement. These Approaches are classified by data handling into $(i)$ table-specific LLM fine-tuning, $(ii)$ table content retrieval, and $(iii)$ table-as-image analysis.

\noindent\underline{(1) Table-Specific LLM Fine-Tuning:} Fine-tuning LLMs on task-specific table datasets embeds analytical knowledge directly into model parameters. TableGPT~\cite{li2023tablegpt} fine-tunes GPT-3.5 on diverse real-world table tasks, while TableGPT2~\cite{su2024tablegpt2}, built on Qwen2.5~\cite{qwen2025qwen25} and pre-trained on 593.8K tables and fine-tuned on 2.36M QA pairs, introduces a table encoder for hybrid table representation, an adapter for query encoding, and an LLM decoder to generate agent workflows like tool execution pipelines for deriving the final answer.

\noindent\underline{(2) Table-As-Image Analysis:} To overcome the limitations of text-only LLM in understanding table structures, tables are converted to images for multimodal LLM analysis. Table-LLaVA~\cite{zheng2024multimodal} pretrains LLaVA-7B~\cite{liu2024llava} on 150K table recognition samples (e.g., output HTML of the input table) to align table structures and elements with textual modality and fine-tunes on 232K downstream tasks (e.g., fact verification, question answering) to enhance its instruction-following ability. TabPedia~\cite{zhao2024tabpedia} enables a single model to handle diverse table analysis tasks via a concept-synergy mechanism that abstracts tasks into concepts. Built on Vicuna-7B~\cite{zheng2023judging}, it adds meditative tokens to the LLM decoder input, adaptively activating visual token regions to interpret task-specific intents. 


\hi{LLM for Time Series Analysis.} Time series data, a specialized type of relational data stored in relational or time-series databases (e.g., InfluxDB~\cite{naqvi2017time}), records sequential observations over time, each capturing one or more variables. Its analysis targets temporal trends, periodic patterns, and inter-variable dependencies (e.g., causality, correlation, coupling) to enable prediction, classification, and detection~\cite{liu2025towards}. Unlike static relational analysis, time series emphasizes the dynamic evolution of variables along the timeline.


Traditional time series analysis relies on statistical and machine-learning methods, such as ARIMA and LSTM~\cite{8614252}. However, statistical approaches are limited by linear assumptions, manual feature engineering, and difficulty capturing complex nonlinear patterns~\cite{rahmani2022association}. Advances in machine and deep learning offer more flexible solutions~\cite{9005997}, alongside efforts to improve result interpretability~\cite{9713986}. Recently, LLMs have been applied to time series analysis, leveraging their strengths in sequence modeling, generalization, and long-term dependency capture.


\noindent$\bullet$ \textit{\underline{TS2NL.}}
Due to the sequential similarity between time series and natural language, TS2NL converts time series data into natural language to enhance intuitiveness and interpretability for LLMs.


\noindent\underline{(1) End-to-End Reasoning:} End-to-end reasoning requires the LLM to generate results directly from a prompt containing time series data, instructions, and contextual information (e.g., news, events, logs). For instance, SIGLLM~\cite{alnegheimish2024can} explores anomaly detection by either directly querying the LLM or using it for prediction and comparing results with actual data. TimeRAG~\cite{yang2025timerag} leverages historical time series for domain adaptation, retrieving reference sequences via dynamic time warping (DTW)~\cite{senin2008dynamic} to strengthen predictions. Other studies examine LLMs’ zero- or few-shot forecasting abilities in domains like finance~\cite{yu2023temporal} and climate~\cite{DBLP:journals/corr/abs-2411-13724}.


\noindent\underline{(2) Step-wise Agent Methods:} 
Step-wise agent methods employ multiple LLM agents, each completing tasks under specific prompts, to collaboratively perform analysis. TimeCAP~\cite{lee2025timecap} employs two agents—a summary agent that generates contextual summaries of time series data and a prediction agent that uses these summaries for forecasting———together with a multi-modal encoder integrating data, summaries, and sampled text.And the final prediction is derived from the linear combination of encoder and agent outputs. TimeXL~\cite{jiang2025explainable} extends this with a prototype-based encoder and three agents. The encoder produces preliminary predictions and explanations based on time series and textual inputs. One agent generates predictions based on textual explanations, another compares them with actual series to provide feedback, and the third refines textual quality and triggers encoder updates. The final prediction is obtained by weighting the outputs of the encoder and agents. For context-augmented forecasting, CABINET~\cite{wang2024news}, a reasoning agent selects relevant news to provide context for predictions, while an evaluation agent critiques both the news selection and prediction against ground truth, updating the reasoning agent’s logic.

\noindent$\bullet$ \textit{\underline{Time Series Alignment.}}
To address modality differences between time series data and natural language, some methods align the modalities to enable LLMs to capture temporal patterns and dependencies more effectively. Based on the model architecture, these approaches are classified into designed encoder/decoder frameworks and fine-tuned LLM bodies, with a comparison summarized in Table~\ref{tab:alignment}.


\noindent\underline{(1) Designed Encoder/Decoder:}
Specialized encoder or decoder modules adapt time series data to LLM input or output formats without modifying the LLM backbone.
TIME-LLM~\cite{DBLP:conf/iclr/0005WMCZSCLLPW24} integrates time series and textual context by encoding time series with a patch reprogram module using text prototypes from pre-trained embeddings, prepending prompts encoded by a frozen LLaMA-7B embedder, and generating forecasts through a trainable projection module. SEED~\cite{li2025seed} bridges structural, numerical and semantic gaps using a token-aware structural encoder, patch projection and alignment, and semantic reprogramming, concatenating multivariate time series tokens with the task prompt as LLM input.
In contrast, TimeCMA~\cite{liu2025timecma} avoids feeding all embeddings into the LLM due to its limitations in learning disentangled representations. It encodes time series through two branches, a time series branch and an LLM-empowered branch, and combines them via cross-modality alignment for multivariate forecasting.


\noindent\underline{(2) LLM Fine-Tuning:}
Some methods fine-tune LLMs with encoder or decoder modules on time-series tasks, enhancing their understanding and analysis of complex time series.
CALF~\cite{liu2025calf} targets embeddings, hidden features, and predictions, applying cross-modal token alignment to derive aligned text tokens queried by time tokens, feature regularization loss between the textual and temporal hidden feature and output consistency loss between textual and temporal prediction to reduce distribution discrepancies during training.
S2IP-LLM~\cite{pan2024sipllm}, on the other hand, decomposes time series by seasonal-trend to explicitly encode each component. Tokens of components and relevant prompts are concatenated as input, while the component predictions are combined as the overall prediction.
LLM4TS~\cite{chang2025llm4ts} uses a two-stage fine-tuning which aligns LLMs with time series data, then fine-tuning on forecasting tasks. A two-layer aggregation pools the sum of embeddings across time scales to improving temporal understanding ability.
Apart from prediction tasks, LLMFew~\cite{chen2025large} addresses multivariate time series classification, using a patch-wise temporal convolution encoder (PTCEnc) to align time series with LLM input, and fine-tunes LLMs via LoRA to enhance feature representation.

\begin{table}
\caption{Comparison of Time-Series Analysis Methods.}
\label{tab:alignment}
\renewcommand{\arraystretch}{0.7} 
\begin{tabularx}{\columnwidth}{>{\centering\arraybackslash}X>{\centering\arraybackslash}X>{\centering\arraybackslash}X>{\centering\arraybackslash}X}
\hline
\textbf{Method} & \textbf{Encoder} & \textbf{Backbone} & \textbf{Decoder} \\
\hline
\multirow{2}{*}{TIME-LLM~\cite{DBLP:conf/iclr/0005WMCZSCLLPW24}} & \multirow{2}{*}{Concatenation} & \multirow{2}{*}{--} & Linear Projection \\ 
                          &                          &                          & \\
\multirow{2}{*}{SEED~\cite{li2025seed}}     & \multirow{2}{*}{Concatenation} & \multirow{2}{*}{--} & Linear Projection \\
                          &                          &                          & \\
\multirow{2}{*}{TimeCMA~\cite{liu2025timecma}}  & \multirow{2}{*}{Dual-Modality} & \multirow{2}{*}{--} & Multivariate Transformer \\ \hline

\multirow{2}{*}{CALF~\cite{liu2025calf}} & \multirow{2}{*}{Dual-Modality} & \multirow{2}{*}{LoRA} & \multirow{2}{*}{Dual-Modality} \\ \\
                          &                          &                          & \\
S2IP-LLM~\cite{pan2024sipllm}  & Decomposition & PEFT & Combination\\
                          &                          &                          & \\
\multirow{2}{*}{LLM4TS~\cite{chang2025llm4ts}}    & \multirow{2}{*}{Multi-Scale} & Two-Stage PEFT & \multirow{2}{*}{RevIN}\\ \\
\multirow{2}{*}{LLMFew~\cite{chen2025large}}  & \multirow{2}{*}{PTCEnc} & \multirow{2}{*}{LoRA} & Classification Head\\ \hline
\end{tabularx}
\end{table}

\hi{Relational Data Curation.}
Training or fine-tuning \llms on domain-specific datasets is essential for adapting them to specialized knowledge (e.g., finance, medicine, climate) and structured data patterns (e.g., relational, time series, graph). Meanwhile, test datasets serve as benchmarks for evaluating t performance.

The traditional method of dataset construction is usually the combination of real data collection and manual annotation. For example, the widely used table QA dataset WikiTQ~\cite{pasupat2015compositional} uses tables from Wikipedia and question-label pairs annotated by humans. Similarly, the NL2SQL dataset Spider~\cite{yu2018spider} uses databases collected from existing resources and question-SQL pairs annotated by college students. Annotation quality, diversity, and readability are ensured through multiple mannuel reviews, with SQL correctness verified by scripts. MMTU~\cite{xing2025mmtu} provides a comprehensive collection of 52 such manually curated benchmarks on structured tables, organized across different tasks and benchmark datasets. 

However, in some domains, data are scattered in small batches or kept confidential for business and privacy reasons, making manual collection difficult. Moreover, manual annotation is not only labor- and time-intensive but also demands substantial domain expertise. In such cases, semi-automatic or fully automatic dataset generation becomes increasingly necessary.

REaLTabFormer~\cite{solatorio2023realtabformer} generates relational datasets with the transformer. It first generates the parent table with a GPT-2 model and then generates the relational dataset conditioned on that with a Seq2Seq model. 
Another method~\cite{hudovernik2024relational} utilizes the structural representation capability of GNNs and the generative capability of diffusion models. It represents a relational database by a heterogeneous graph with the method in ~\cite{xu2022synthetic}, where a parent-child table pair connected by a foreign key is modeled as an attributed bipartite graph. 

Regarding data augmentation, SyntaxSQLNet~\cite{DBLP:conf/emnlp/YuYYZWLR18} proposes a method to expand complex SQL datasets. It extracts some universal templates of question-SQL pairs from Spider, and then fill the templates with the schema and values from the target database.
Furthermore, CodeS~\cite{li2024codes} achieves domain adaptation with low annotation costs by Bi-directional augmentation. The Question-to-SQL augmentation gives \llms a few annotated question-SQL pairs, and \llms can generate new pairs by simulating them. The SQL-to-question augmentation fills in question-SQL templates with the target domain database, and the \llms then rewrite the question to more natural language.

~\cite{klopries2024itf} generates time series dataset by two interpretable methods. The first is the ITF-FM that operates the feature manipulation process based on the given dataset, The second is the ITF-GAN that is trained to generate indistinguishable data samples compared to the original data.
ChatTS~\cite{xie2024chatts} comes up with the Attribute-Based 
Time Series Generator to get synthetic text-time series pairs for model training. \llms are used to select proper features according to real-world settings for each attributes like trend, periodicity, local fluctuation and noise. The subsequent modules will generate the final time series based on the selection of the \llms.

\subsection{Graph Data Analysis} 
\label{sec:graph}

Unlike relational data, graph data captures entities (vertices) and their relationships (edges), enabling explicit modeling of complex network semantics (e.g., social networks, knowledge graphs). While highly expressive, this structure introduces unique challenges, including vast search spaces and intricate multi-hop reasoning~\cite{defgraphdata}.

Graph data analysis, compared with relational data, entails more complex tasks, such as summarizing multi-hop relationships and reasoning over text-attributed graphs, where nodes and edges carry textual information~\cite{liang2024natnl2gql,zhou2024r3}. Graphs can be stored not only in relational databases but also in knowledge graphs, accessed via SPARQL in RDF databases (e.g., Blazegraph~\cite{blazegraph}, GraphDB~\cite{graphdb}) or Cypher in Neo4j~\cite{neo4j}.


Traditional graph analysis methods, including statistical approaches and graph neural networks (GNNs), cover a wide range of tasks such as {node classification} (e.g., categorizing papers by research domain), {graph classification} (e.g., predicting node properties over molecular graphs), {link prediction} (i.e., inferring latent relationships), {community detection} (i.e., identifying densely connected subgraphs), {anomaly detection} (i.e., identifying deviations from expected patterns), {graph clustering}, and etc. However, these methods have their own limitations. Statistics-based methods fail to handle complex semantic information, while graph neural networks (GNNs) exhibit limited generalization capabilities, necessitating task-specific retraining on different tasks.


In contrast, the advent of \llms offers transformative potential by leveraging their advanced reasoning capacities and cross-domain generalization abilities, which can $(i)$ simplify the query writing costs (e.g., NL interfaces) and $(ii)$ achieve semantic-aware analysis unsupported in traditional ones.

\hi{Natural Language to Graph Analysis Query.} {Different from NL2SQL, the syntax of graph query language generation is more complex (i.e., MATCH, LOOKUP, GET and other operations unique to graph data manipulation) and there exist two operation objects (i.e., vertex and edge)~\cite{zhou2024r3}.} 
By integrating natural language interfaces with graph data, LLMs facilitate flexible and efficient query generation without the need for specialized model architectures.

To enhance LLMs' comprehension of the complex syntax of Graph Query Language (GQL), $R^3$-NL2GQL~\cite{zhou2024r3} proposes a hybrid approach leveraging relatively small \llm (e.g., LLaMA3-7B) as a selector and GQL rewriter, while employing a larger LLM (e.g., GPT-4) as a reasoner. The selector identifies the necessary CRUD functions, clauses, and schema, while the rewriter refines the query by aligning it with the relevant graph data retrieved by minimum edit distance and semantic similarity calculation. The LLM then synthesizes the aligned question, selected operations, and schema to generate the final GQL query.

To address the limitations of LLMs in planning and collaborating with other LLMs, NAT-NL2GQL~\cite{liang2024natnl2gql} introduces a three-agent framework. The Preprocessor agent constructs context information, including query rewriting, path linking, and the extraction of query-relevant schemas. The Generator agent, an LLM fine-tuned with NL-GQL data, generates GQL statements based on the rewritten queries and extracted schemas. The Refiner agent iteratively enhances the GQL or contextual information by leveraging error feedback from GQL execution results. 

To align LLMs with graph databases in specific fields such as finance and medicine, ~\cite{liang2024aligning} proposes a pipeline combining the fine-tuning and the prompt method. NL-GQL pairs are generated by LLMs for fine-tuning, and the consistency is ensured by the background graph databases and two mutual verification self-instruct methods. In addition, due to the importance of relevant schema in the queries generating  of LLMs, it will be extracted and integrated into prompts.

Note that, within the context of AI for Science (AI4Science), the integration of LLMs with graph data analysis has also shown significant potential and wide-ranging applications (e.g., treat polymers as graphs and predict their properties~\cite{li2024graph,pei2024bio}), which is not the primary focus of this survey.


\hi{LLM-Based Semantic Analysis.} Furthermore, some tasks require semantic-aware analysis, such as summarizing textual content embedded within graph nodes. Depending on the adopted LLM strategies, existing approaches can be categorized into four types: $(i)$ retrieval-then-reasoning methods, $(ii)$ execution-then-reasoning methods, $(iii)$ graph-task-based fine-tuning methods, and $(iv)$agent-based methods.

\noindent$\bullet$ \textit{\underline{Retrieval-Then-Reasoning.}}
Retrieval-then-reasoning first extracts a question-specific subgraph from the graph to identify the most relevant entities and then generates answers using \llms.
To address the challenge of a vast search space, \cite{zhang2022subgraph} introduces a two-stage approach. First, a trainable and decoupled subgraph retriever selects a relevant subgraph based on the query. Then, reasoning is performed over the retrieved subgraph to derive the final answer. UniKGQA~\cite{jiang2023unikgqa} integrates retrieval and reasoning within a unified model architecture.
{It consists of a semantic matching module and an information propagation module where the former employs a pre-trained RoBERTa~\cite{RoBERTa} to align questions with graph relations and the latter propagates matching signals along directed edges.}
Regarding the real-world textual graphs QA, G-Retriever~\cite{he2024g} retrieves relevant nodes and edges based on text encoding, and then uses PCST optimization algorithm to construct subgraphs. The subgraph structure is encoded through Graph Attention Network (GAT) while its content is extracted in text format. 
{These representations are then fed into LLMs to generate the final answer.}

\noindent$\bullet$ \textit{\underline{Execution-Then-Reasoning.}} Execution-then-reasoning refers to the process of parsing natural language queries into executable logical forms (e.g., SPARQL) that align with the graph data, followed by reasoning based on the output of the executed program.
Interactive-KBQA~\cite{xiong2024interactivekbqa} introduces an interactive \llm QA framework with a unified SPARQL-based toolset (e.g., entity search, graph pattern search, SPARQL execution, etc.) designed to address complex queries. 
MCTS-KBQA~\cite{xiong2025mcts} introduces the MCTS method into KBQA, which iteratively executes selection, expansion, evaluation, backpropagation and termination in the graph. A step-wise reward mechanism was designed, using LLMs to evaluate and score each step.
FlexKBQA~\cite{li2024flexkbqa} addresses the challenge of lacking high-quality annotated data in real-world scenarios. By prompting \llms as program translators, it samples program-answer pairs from the knowledge base and generates corresponding natural language questions. The synthetic question-program-answer dataset is used to train lightweight models through execution-guided self-training, which are subsequently employed to annotate real user queries. This approach addresses the distribution shifts between synthetic and actual data, leading to significant improvements in few-shot learning scenarios.

\noindent$\bullet$
\textit{\underline{Graph-Task-Based Fine-Tuning Methods.}} InstructGLM \cite{ye2024instructglm} enables generative graph learning by fine-tuning an \llm and leveraging natural language descriptions of graph structures (e.g., offer the first node and the 1-/2-/3-hop neighbors' information).
InstructGraph~\cite{wang2024instructgraph} introduces a stricter code-like graph representation format which constructs entities and triples in the form of list, whose backbone \llm (LLaMA2-7B) is fine-tuned on a graph-centric corpus comprising 1.6 million instances. To mitigate the issue of hallucination, it incorporates Direct Preference Optimization (DPO) algorithm~\cite{rafailov2024dpo} for preference alignment.
GraphGPT~\cite{tang2024graphgpt} enhances model performance in zero-shot scenarios by  incorporating a structural information encoding module based on Graph-SAGE~\cite{10.5555/3294771.3294869} and GCN~\cite{kipf2017gcn}. It fine-tunes the projector bridging the graph encoder and the LLM decoder to align the language capabilities of the foundation LLM (Vicuna-7B) with the graph learning tasks.
GLaM~\cite{dernbach2024glam} fine-tunes \llms to integrate domain-specific knowledge graphs directly into them, which enhances their reasoning capacity. It iteratively partitions and encodes the neighborhood subgraph around each node in a knowledge graph to obtain the context and QA data for fine-tuning. 

\noindent$\bullet$ \textit{\underline{Agent-Based Methods.}}
Agent-based methods involve leveraging LLM-based agents with predefined tools (e.g., human-written interfaces or graph processing library APIs) that iteratively interact with the graph data to retrieve, refine, and operate information.
StructGPT~\cite{jiang2023structgpt} introduces an iterative reading-then-reasoning framework, leveraging specialized interfaces to operate on graph data. It repeatedly applies an invoke-linearize-generate procedure to derive query results. 
KBQA-o1~\cite{luo2025kbqa} utilizes a ReAct agent and MCTS to explore the knowledge base. During the exploration process, the agent gradually generates the logical form of knowledge base operations. The MCTS with policy and reward models helps balance the exploration’s performance and search space.
Another approach is to generate an entire reasoning path based on the query and refine it only when necessary. Readi~\cite{cheng2024necessary} initially constructs a reasoning path and instantiates it on the graph. When execution errors occur, it collects error messages and invokes an LLM to revise the path. The final answer is inferred from the instantiated graphs.

\hi{Graph Data Generation.}
To cope with the lack of large-scale structured graph datasets, a framework~\cite{10943225} to generate Large-Scale Graph Dataset is proposed. It generates structures and features separately by fitting the original graphs, and aligns them to obtain the generated graph.
~\cite{TKGGen} presents the pipeline that generates temporal knowledge graph (TKG) datasets based on documents. It utilizes LLMs to generate timestamps for existing triplets, thereby obtaining the quadruples needed for constructing TKG.

\begin{tcolorbox}[colback=white,colframe=black!40,
  boxrule=0.5pt,arc=4pt,left=6pt,right=6pt,top=4pt,bottom=4pt]
\textbf{Takeaway (Structured Data):}

\textbf{O1:} LLMs capture structural and semantic relations across tables, graphs, and time series data, enabling unified understanding.

\textbf{O2:} LLMs translate user intents into executable queries (e.g., SQL, graph queries) for intuitive data interaction.

\textbf{O3:} LLMs integrate semantic understanding to execute compositional operations with semantic awareness.

\textbf{O4:} LLM agents refine the analysis pipeline and evolve with dynamic data and tasks to boost analytical abilities.
\end{tcolorbox}

%% file: sections/semistructured-data.tex
\section{\llm for Semi-Structured Data Analysis} \label{sec:semistructured-data}

Semi-structured data are neither strictly schema-defined like relational data nor completely raw like text or images~\cite{abiteboul1997querying}. They retain partial organizational properties (e.g., tags, headers) and often have hierarchical or nested structures (e.g., \textit{County}–\textit{Province}–\textit{City} in JSON). This allows representation in diverse formats such as web tables, spreadsheets, HTML, JSON, and XML. They can be broadly categorized into markup languages and semi-structured tables, both challenging downstream tasks due to their structural complexity.



\subsection{Markup Language}
\label{sec:markup}

Markup languages (e.g., XML, JSON, HTML) are widely used for structuring and exchanging data, yet research on integrating LLMs with them remains limited. Traditional approaches for markup understanding fall into two categories. Token Linearization converts markup documents into plain text sequences for PLMs~\cite{herzig2020tapas}, but this often oversimplifies and fails to capture structural complexity. Tree/Graph-Based PLMs explicitly encode structures (e.g., DOM trees), yet they suffer from limited generalization, short context windows, high pre-training costs, and poor scalability.

\begin{figure}[!t]
    \centering
    \includegraphics[width=1\linewidth]{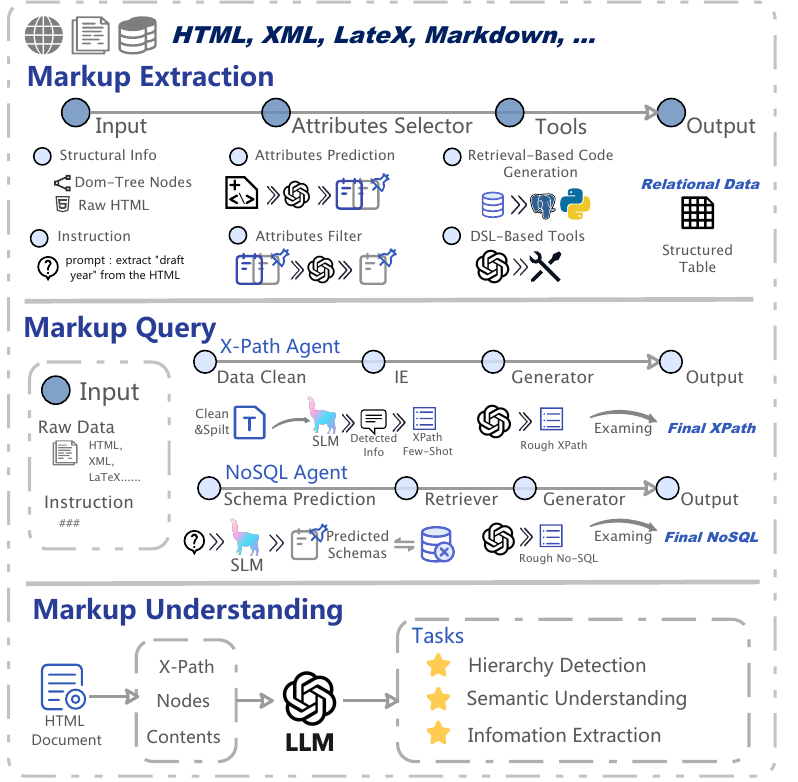}
    \caption{\textbf{LLM for Markup Data Analysis.}}
    \label{fig:markupoverview}
\end{figure}

LLM-based methods have recently emerged as a promising alternative, offering stronger semantic reasoning and greater adaptability for complex markup data, which is summarized in Figure~\ref{fig:markupoverview}.


\hi{Markup Extraction.}
Markup Extraction aims to transform semi-structured data (e.g., HTML, XML) into structured formats such as relational tables. Traditional methods rely on heuristic rules (e.g., extracting content within $<h1>$ or $<title>$ tags~\cite{wang2002detecting}) or data templates (e.g., extracting product names from $<$h1 class=``product-name''$>$~\cite{crescenzi2001automatic}), but they often suffer from template drift and high annotation costs. Leveraging LLMs enables direct extraction and interpretation of hierarchical relationships, attributes, and nested structures, reducing dependence on tags or templates.


Evaporate~\cite{arora2023language} is a prototype system that uses LLMs to generate structured views from semi-structured documents. It employs Evaporate-Code+, which first synthesizes multiple candidate extraction functions via LLMs. Functions are generated in two modes, where one provides few-shot examples encouraging general-purpose Python solutions, and the other specifies only task requirements to produce regex-based solutions. A weak supervision–inspired aggregation mechanism then evaluates and filters candidates based on output quality.

Similarly, WebFormer~\cite{wang2022webformer} targets HTML specifically by leveraging its layout structure. It encodes DOM nodes as graph-attended tokens and models text with relative-position self-attention. Field tokens guide heterogeneous cross-attention with HTML and text tokens, enabling accurate and scalable structured field extraction.

\hi{Markup Information Querying.}
Markup information querying extracts user-intended data from markup languages. While conceptually similar to SQL querying over relational databases, markup files are often irregular and loosely structured, posing unique challenges. Existing research mainly explores tree-based and NoSQL-based approaches.



\noindent$\bullet$ \textit{\underline{Tree-Based Approaches.}} 
XPath Agent~\cite{li2024xpath} uses a two-stage pipeline for XPath queries. In the information extraction stage, a lightweight LLM analyzes a simplified webpage, with irrelevant nodes and attributes removed via HTML cleaning algorithm to produce a compressed HTML. In the XPath Programming stage, a stronger LLM constructs XPath queries bottom-up, traversing from target nodes to parent elements and flexibly using attributes like class names and IDs.


\noindent$\bullet$ \textit{\underline{NoSQL-Based Approaches.}} 
SMART~\cite{lu2025bridging} is a Text-to-NoSQL agent that integrates small language models (SLMs), retrieval-augmented generation (RAG), and LLMs. An SLM first predicts schemas and an initial NoSQL query from the natural language input. RAG then retrieves relevant data using the preliminary query and schema, which, combined with few-shot exemplars, guides the LLM to generate a refined NoSQL query applicable to large-scale semi-structured data.



\hi{Markup Information Understanding.}
Unlike markup querying, markup information understanding focuses on using LLMs for more complex operations on markup data (e.g., revision and error correction). To support this, both end-to-end and pipeline-based models have been developed for comprehensive markup file understanding.



DOM-LM~\cite{deng2022dom} introduces a structure-aware Transformer that jointly encodes HTML text and DOM tree structure. To handle large DOM trees, it partitions them into subtrees and uses structure-aware positional embeddings (e.g., node depth, parent, and sibling indices) to capture hierarchy. The model is trained with a self-supervised masking strategy over both tokens and DOM nodes.
MarkupLM~\cite{li2021markuplm} extends token representations from four perspectives, including text embeddings, 1D positional embeddings for horizontal order, segment embeddings for paragraph distinctions, and XPath embeddings. XPath embeddings are created by decomposing an XPath into hierarchical layers of tags and subscripts, combining each layer’s tag and subscript embeddings, concatenating across layers, and passing through a feed-forward network to produce the final representation. For pre-training, MarkupLM uses three markup-specific objectives. Masked Markup Language Modeling (MMLM) for joint textual-structural understanding, Node Relationship Prediction (NRP) to learn DOM hierarchies, and Title–Page Matching (TPM) for global document coherence.

WebLM~\cite{xu2024hierarchical} presents a hierarchical multimodal pre-training framework for HTML webpage understanding, integrating structural, textual, and visual information within a unified Transformer. Visual features are aligned with HTML nodes, and two training objectives are used. Tree Structure Prediction (TSP) captures parent–child and sibling relations, and Visual Misalignment Detection (VMD) improves robustness to layout variations.



\hi{Insights of Markup Language Analysis.}
Most studies on markup language understanding still rely on PLMs for two reasons. $(i)$ Controllability: PLMs allow explicit control over model size, training data, and task settings, whereas LLM APIs are often closed-source and costly. $(ii)$ Structural Adaptation: PLMs can use specialized tokenization and positional encodings to capture HTML/XML hierarchies~\cite{deng2022dom,xu2024hierarchical}, while general-purpose LLMs depend on prompt engineering, leading to redundant tokenization, long prompts, and difficulty jointly modeling semantics and structure (e.g., converting an HTML table to text requires auxiliary schemas).


However, given the strong semantic capabilities of LLMs, leveraging them for markup language understanding remains promising, provided that effective methods are developed to compensate for their limitations while maximizing their semantic reasoning power. To this end, we identify two key research insights, which can be categorized into information compression and tool integration.



\noindent$\bullet$ \textit{\underline{Information Compression.}} 
Information compression addresses the challenge of encoding markup languages by providing LLMs with more concise representations. Traditional methods, such as structural or schema-aware prompting, often produce redundant rules and long prompts. Segment-wise prompting partitions documents into manageable segments and selectively filters relevant content. For renderable languages like HTML or LaTeX, vision–language models can assist by extracting visual anchors and generating tree-structured encodings, allowing LLMs to receive task-relevant information while reducing prompt length and preserving structural context.



\noindent$\bullet$ \textit{\underline{Tool Integration.}} 
Tool Integration for LLMs aims to enable LLMs to interact with and modify external environments. Existing methods such as DSL-based soft tools, retrieval-based code generation, and API-calling frameworks, are often rigid, with APIs providing fixed mappings from natural language to predefined functions.More dynamic approaches are needed, including planning modules, learning-based tool selection, and self-reflective reasoning loops.


\subsection{Semi-Structured Tables}
\label{sec:semitable}



This section focuses on semi-structured table processing, encompassing techniques for understanding, representing, and extracting information from such tables. Unlike structured relational data, semi-structured tables feature complex structures, including merged cells, hierarchical headers, and nested tables.



\begin{figure}
    \centering
    \includegraphics[width=1\linewidth]{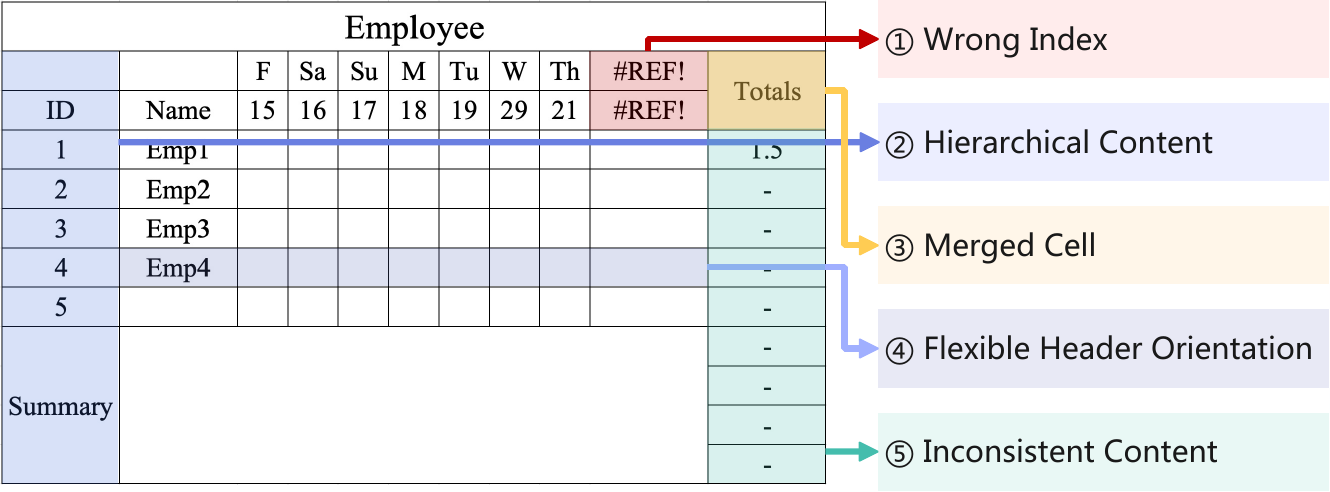}
    \caption{Example Characters of Semi-Structured Tables.}
    \label{fig:SemiTable}
\end{figure}

Semi-structured tables share a set of properties, which can be summarized into five core characteristics, illustrated in Figure~\ref{fig:SemiTable}:
$(i)$ Wrong Index: row or column indices may be missing, ambiguous, or inconsistent~\cite{suhara2022annotating}.
$(ii)$ Hierarchical Content: rows or columns may not follow a uniform schema (e.g., a ``summary'' cell under an ``ID'' column)~\cite{yang2022tableformer}.
$(iii)$ Merged Cell: a single cell may span multiple rows or columns~\cite{guo2022trust}.
$(iv)$ Flexible Header Orientation: keys and values can appear in varying positions (e.g., left-of or above their associated values).
$(v)$ Inconsistent Content: cell values vary widely, including words, numbers, or full sentences.

With the aid of LLMs, it becomes possible to directly reason over semi-structured tables, interpret their intricate layouts, and bridge the gap between natural language queries and heterogeneous table structures, thereby enabling more accurate and context-aware query answering.

\begin{figure}[!t]
    \centering
    \includegraphics[width=1\linewidth]{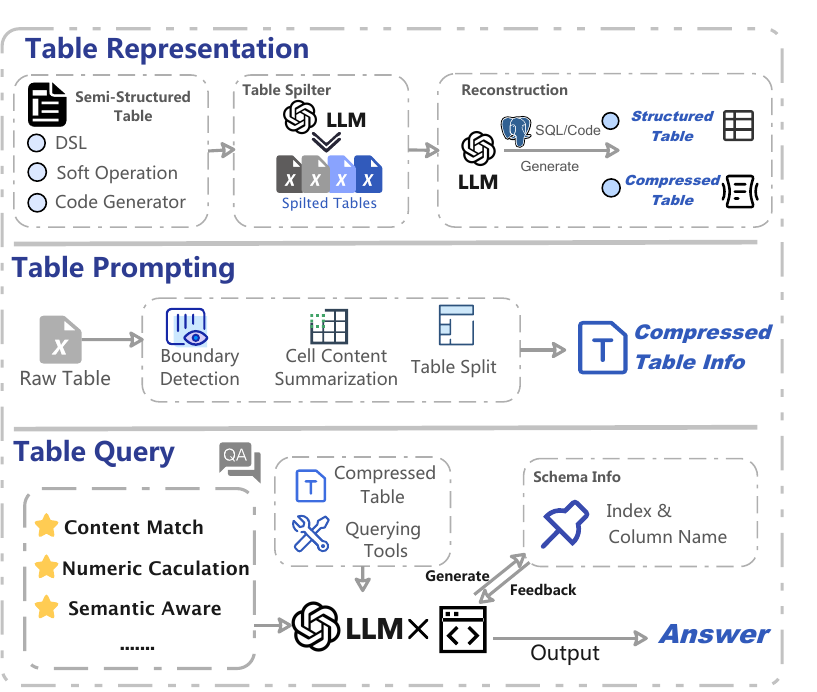}
    \caption{{LLM for Semi-Structured Table Analysis.}}
    \label{fig:semitableoverview}
\end{figure}

Traditional table processing approaches like TAPAS~\cite{herzig2020tapas} mainly rely on lightweight models (e.g., BERT\cite{devlin2019bert}, RoBERTa\cite{liu2019roberta}) to analyze the table. These methods rely on specialized embedding mechanisms to capture the two-dimensional structure of tables, including rows, columns, and numerical order. To support tasks such as cell selection or aggregation, they usually require large-scale weakly supervised pre-training~\cite{yang2022tableformer}. However, due to the small number of parameters and weak NLP capabilities of lightweight models, this method cannot handle some tasks based on semi-structured tables with confusing layouts.
Leveraging the strong semantic understanding capabilities of LLMs, LLM-based approaches generally preprocess the semi-structured table and then perform table understanding or reasoning. We conducted some research on this, and the work is shown in Figure~\ref{fig:semitableoverview}.


\hi{Table Representation.}
By converting semi-structured tables into structured representations, LLMs can better capture the underlying logical relationships within tables, thereby improving their effectiveness in downstream tasks. However, the diverse and complex layouts of semi-structured tables (e.g., hierarchical headers, merged cells, and nested subtables) pose significant challenges to this process. To address these issues, tree-based modeling approaches and model-driven methods have been proposed.



\noindent$\bullet$ \textit{\underline{Tree-based Table Modeling.}}
TUTA~\cite{wang2021tuta} introduces a tree-based encoding for semi-structured tables, where the root splits into row and column nodes, each expanding hierarchically into trees. Table entries are nodes with parents corresponding to their row and column (e.g., column nodes like ``ID'' or ``Phone Number'' and row nodes representing individual names).
ST-Raptor~\cite{tang2025st} proposes a HO-Tree to encode tables, combining a Meta-Tree for header structure and semantics with a Body-Tree for content. It uses a VLM to detect meta-information, heuristic rules to segment tables, and a depth-first strategy to construct the HO-Tree.

\noindent$\bullet$ \textit{\underline{Model-based Structuring.}} 
TabFormer~\cite{10.1145/3726302.3730071} converts semi-structured tables into relational data using a chain-of-thought (CoT) approach. LLMs first apply step-by-step transformations with ``soft operators'' (e.g., splitting tables with multiple header groups). After preprocessing, SQL is used to reconstruct the table, including predicting indices, generating \texttt{CREATE} statements for the schema, and inserting the extracted contents into the relational table.

Auto-Tables~\cite{li2023autotablessynthesizingmultisteptransformations} integrates OCR for structure recognition and applies domain-specific language (DSL) operators. These operators abstract common spreadsheet operations (e.g., melt, transpose), thereby enhancing generalization and adaptability in table conversion tasks.
TradeSweep~\cite{lee2024can}, while closely related to Auto-Tables, differs in its approach. Instead of predefined DSL operators, it accepts user-specified transformation requirements and directly generates table conversion code by leveraging a code repository, offering greater flexibility in addressing diverse user needs.


\hi{Table Prompting.}
Table prompting converts table data into text input for LLMs. Traditional approaches, like table linearization, flatten all cell contents into a single string, often producing overly long prompts and losing structural information. To address this, table content and structure should be compressed, enabling LLMs to handle large tables while keeping structural loss controllable.

SHEETCOMPRESSOR~\cite{dong2024spreadsheetllm} efficiently serializes large spreadsheets for LLM prompting. Structural Anchors identify key heterogeneous rows and columns, pruning redundant ones to create a skeletonized sheet. Inverted-Index Translation encodes non-empty cells in a JSON dictionary, merging duplicates to reduce storage. Data Format Aggregation clusters adjacent numerical cells of the same type or format, capturing distribution patterns without redundancy.
HySem~\cite{pp2024hysem} introduces a lightweight LLM pipeline for local deployment. Its Context Optimizer compresses HTML table tokens so each cell uses only one or two words (e.g., compress ``Theme 3: Factors affecting stuff response'' into ``Theme 3''), establishing a direct mapping between markup and content while minimizing token usage.

\hi{Table Querying.}
LLMs require effective methods to extract table contents, known as table querying. While conceptually similar to SQL, this section focuses on approaches that directly extract information from semi-structured tables rather than traditional SQL querying.
CoS~\cite{dong2024spreadsheetllm} is a downstream table query system built on SHEETCOMPRESSOR. Given a spreadsheet and a natural language query, it first performs Table Identification and Boundary Detection to locate the relevant subtable. The subtable and query are then passed to the LLM for focused, context-aware response generation.
In ST-Raptor~\cite{tang2025st}, table querying relies on the HO-Tree. Complex questions are decomposed into sub-questions mapped to sequences of basic operations (e.g., retrieval, condition). The alignment operation grounds abstract query terms to actual headers or values, while Top-down and Bottom-up retrieval navigate the HO-Tree. For large tables, Column-Type Annotation (e.g., discrete or continuous) limits the search space. A dual validation mechanism—forward constraints for intermediate consistency and backward reasoning to reconstruct the query—reduces errors and hallucinations.

\hi{Semi-Structured Data for LLM.}
To facilitate research on table understanding and reasoning under challenging conditions, a number of datasets have been developed. 
Below, we introduce several representative datasets that are widely used for semi-structured table understanding.

TEMPTABQA~\cite{gupta2023temptabqa} contains 11,454 question–answer pairs centered on temporal queries while SPREADSHEETBENCH~\cite{ma2024spreadsheet} provides a challenging benchmark for spreadsheet manipulation, consisting of 912 questions derived from real-world scenarios.
MiMoTable~\cite{li2024mimotable} targets reasoning across multiple sheets and files, with 1,719 queries distributed over 428 spreadsheets. Evaluation results on these benchmarks reveal a significant performance gap, ranging from 20\% to 50\% between state-of-the-art models and human performance, underscoring the need for further advances in this area.
INFOTABS~\cite{gupta2020infotabs} is designed for the natural language inference task on semi-structured tables. It contains 23,738 premise–hypothesis pairs, where premises are Wikipedia infoboxes and hypotheses are short sentences.
WikiTableQuestions~\cite{pasupat2015compositional} focuses on question answering over semi-structured HTML tables, comprising 22,033 question–answer pairs together with their associated tables.
FeTaQA~\cite{nan2022fetaqa} is a free-form table question answering dataset containing 10k Wikipedia-based pairs of tables, questions, answers, and supporting table cells. Unlike extractive QA datasets, FeTaQA provides human-written free-form answers including entities and their high-level relations.
For comparison, Table~\ref{tab:table-datasets} summarizes the characteristics of different datasets. The table highlights structural complexity including total cell counts and the number of merged cells, as well as question–answering types, including: $(i)$ Content Match, where the task is to locate the correct cells, $(ii)$ Numeric Computation, which requires performing arithmetic or aggregation operations, and $(iii)$ Semantic Awareness, which involves semantically related or reasoning-intensive tasks. 
For instance, INFOTABS contains a large number of natural language inference questions, whereas SPREADSHEETBENCH features long, complex natural language queries (often exceeding 50 words) that describe multi-step tasks.


\begin{table}[]
\caption{Statistics of Semi-Structured Table Datasets.}
\label{tab:table-datasets}
\resizebox{\columnwidth}{!}{%
\begin{tabular}{l c c c c c}
\hline
~ & \multicolumn{2}{l}{Table Features} & \multicolumn{3}{l}{Q\&A Features} \\ \hline
~ & \#-Cell & \#-Merged-Cell & \begin{tabular}[c]{@{}l@{}}Content \\ Match\end{tabular} & \begin{tabular}[c]{@{}l@{}}Numeric \\ Computation\end{tabular} & \begin{tabular}[c]{@{}l@{}}Semantic\\Aware\end{tabular} \\ \hline
TEMPTABQA~\cite{gupta2023temptabqa}          & 11–100  & 1–10   & \checkmark & $\times$ & $\times$ \\
SPREADSHEETBENCH~\cite{ma2024spreadsheet}   & 101–1K  & 11–100 & \checkmark & \checkmark & \checkmark \\
MiMoTable~\cite{li2024mimotable}          & 11–100  & 1–10   & \checkmark & $\times$ & $\times$ \\
INFOTABS~\cite{gupta2020infotabs}           & 11–100  & 1–10   & $\times$  & $\times$ & \checkmark \\
WikiTableQuestions~\cite{pasupat2015compositional} & 101–1K  & 11–100 & \checkmark & \checkmark & \checkmark \\
FeTaQA~\cite{nan2022fetaqa}             & 101–1K  & 1–10   & \checkmark & $\times$ & $\times$ \\ \hline
\end{tabular}%
}
\end{table}

\begin{sloppypar}
\begin{tcolorbox}[colback=white,colframe=black!40,
  boxrule=0.5pt,arc=4pt,left=6pt,right=6pt,top=4pt,bottom=4pt]
\textbf{Takeaway (Semi-Structured Data):} 

\textbf{O1}: LLMs enable reasoning over irregular and nested markup and table structures. 

\textbf{O2}: LLMs facilitate intuitive interaction and query generation through natural language. 

\textbf{O3}: LLMs empower semantic-level operations such as structural interpretation, summarization, and context-aware reasoning.

\textbf{O4}: Autonomous evolution ability of such analysis needs further research.
\end{tcolorbox}
\end{sloppypar}

%% file: sections/unstructured-data.tex
\section{\llm for Unstructured Data Analysis} \label{sec:unstructured-data}

Unstructured data refers to data that lacks explicit structure, as it does not adhere to a predefined schema~\cite{bigdataanalysis}. Additionally, it exhibits high variability in format, length, and modality, which further complicates its processing and analysis~\cite{baviskar2021efficient,wang2025aop}.

\begin{figure}[t!]
   \centering
   \includegraphics[width=1\linewidth]{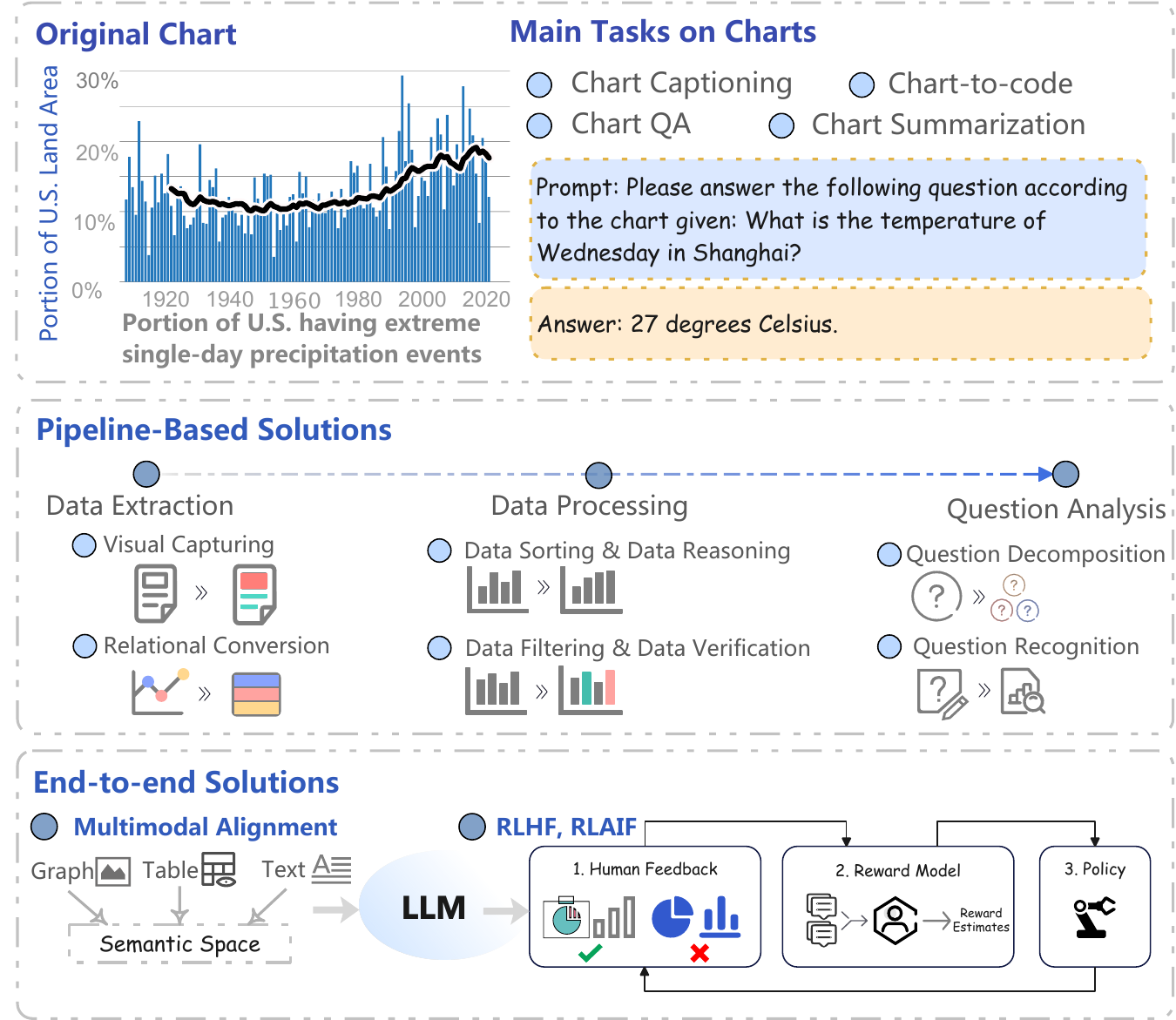}
   \caption{LLM for Chart Data Analysis.}
   \label{fig:chart-ovreview}
\end{figure}

\subsection{Chart}
\label{sec:chart}
A chart is a graphical method for representing data. Since humans generally process visual information more effectively than raw tabular data, charts visualize data in symbolic forms such as bars, pies, and lines. However, as charts are inherently designed for human perception, developing rule-based methods for comprehensive understanding remains difficult. Early approaches introduced an additional layer with a fixed vocabulary to simulate rules. For example, DVQA~\cite{kafle2018dvqaunderstandingdatavisualizations} employed LSTM networks to encode questions and CNNs to encode chart images, but the reliance on a fixed vocabulary limited generalization. Subsequent works such as SANDY~\cite{kafle2018dvqaunderstandingdatavisualizations} alleviate this issue by leveraging OCR ground truth to link chart elements with dictionary entries, though the accuracy of OCR recognition remained a critical bottleneck.

From a technical perspective, chart understanding poses unique challenges as it requires integrating visual perception with structured data extraction, numerical reasoning, and logical inference. As shown in Figure~\ref{fig:chart-ovreview}, tasks on charts show diverse features, requiring a multi-perspective understanding on the structure and content of charts. As a result, multiple tools are useful for chart understanding, each focusing on a specific side of the task. Research in this domain has primarily followed two paradigms: $(i)$ pipeline-based methods, which decompose the task into modality conversion and semantic reasoning, and $(ii)$ end-to-end models, which directly map chart images and instructions to responses within a unified framework. In the following, we provide an overview of chart-related tasks and examine these two LLM-based technical paradigms.

\hi{Chart Captioning.}
Chart Captioning (or Summarization, Chart-to-Text) refers to a task on generating a descriptive caption for a given chart. The caption should contain key information or serve as a summary of the graph.
Early studies primarily target common graphical features such as axes and legends. These features enable different charts can be summarized by successively mentioning all of the features, giving rise to planning-based pipelines for caption generation. For instance, Mittal et al~\cite{mittal-etal-1998-describing} propose feature-specific strategies for different types of graphs, while Reiter~\cite{reiter-2007-architecture} outline a four-stage data-to-text framework comprising $(i)$ Signal Analysis to look for possible patterns and trends, $(ii)$ Data Interpretation to identify more complex messages, $(iii)$ Document Planning to decide which of the messages should be mentioned in the generated text, and $(iv)$ Microplanning and Realisation to create an actual text. Such pipeline-based approaches are stable and extensible in fixed scenarios but tend to be heavy and lack general applicability. Building on this line of work, ChartThinker~\cite{liu2024chartthinker} employs a large-scale dataset and a chain-of-thought strategy to enhance the logical coherence and accuracy of chart summarization. Leveraging LLMs, it first identifies the chart type and then iteratively generates reasoning steps to capture a holistic understanding of the chart.


At the same time, end-to-end solutions have also emerged. \cite{obeid-hoque-2020-chart} introduces a transformer-based model that generates summaries containing data by taking both the underlying table and chart metadata as input. FigCaps-HF~\cite{singh2025figcapshffiguretocaptiongenerativeframework} fine-tunes a figure-captioning model using RLHF. Other approaches extend beyond captioning to jointly address captioning and question answering by encoding chart elements with a grounded text decoder. For instance, UniChart~\cite{masry2023unichart} is a pretrained model that incorporates both low-level tasks (e.g., extracting visual elements) and high-level tasks (e.g., chart interpretation) into its training objectives, enabling more comprehensive chart understanding.

\hi{Chart Question Answering.}
Chart question answering (Chart QA) requires a model to answer questions related to the content of a chart. Compared to the task of captioning a chart, QA also has to grab the key information of a chart. However, the task requires a more flexible understanding of the charts, since a question may focus on subtle and imperceptible features of the charts. As mentioned in ChartQA\cite{masry2022chartqabenchmarkquestionanswering}, the questions might involve both visual and logical reasoning on charts.

Researches relevant to chart QA explores specific aspects of Chart QA task. \cite{wu2024chartinsights} addresses the underexplored area of low-level ChartQA tasks (e.g., identifying correlations) by introducing the ChartInsights dataset and a tailored ``Chain-of-Charts'' prompt strategy. 
For the explanability of answering, Charts-of-Thought \cite{das2025chartsofthoughtenhancingllmvisualization} designed prompts for altogether four steps in data processing: extraction, sorting, verification and question analysis. For multimodal alignment, ChartMoE \cite{xu2025chartmoemixturediverselyaligned} used Mixture of Expert architecture to bridge the modality gap, as well as building ChartMoE-Align, a dataset with nearly 1 million chart-table-JSON-code quadruples to conduct alignment tasks.

Besides, some researches noticed the few-shot ability of LLM and designed relevant end-to-end solutions. ChartGemma~\cite{masry2024chartgemmavisualinstructiontuningchart}, for example, is trained on instruction-tuning data generated directly from chart images, thus exploiting both high-level and low-level information of charts. mPLUG-Owl\cite{ye2024mplugowlmodularizationempowerslarge} is a general-purposed model that also fits the problem. Researchers design a two-stage method to align visual information with text: In the first stage, visual knowledge and abstractor modules are trained; in the second stage, LLM and abstractor modules are fine-tuned using LoRA while freezing visual knowledge model.

\hi{Chart-to-Code.}
Chart-to-Code refers to generating executable code (e.g., matplotlib) that can be rendered into chart images, thereby transforming charts into editable forms. Current solutions are primarily end-to-end and rely on VLMs. ChartMimic~\cite{yang2025chartmimicevaluatinglmmscrossmodal} and ChartMoE~\cite{xu2025chartmoemixturediverselyaligned} provide large-scale benchmarks with extensive chart–code pairs to support this task.

Text2Chart31~\cite{zadeh2025text2chart31instructiontuningchart} introduces a reinforcement learning–based instruction tuning approach for chart generation without requiring human feedback. The models are trained under two phases: supervised fine-tuning followed by reinforcement learning with tailored rewards to enhance performance. Chen et al.~\cite{chen2025breakingsftplateaumultimodal} employ multimodal structured reinforcement learning to surpass the performance plateau of supervised fine-tuning in chart-to-code tasks. The study leverages a multi-granularity structured reward
system that uses multimodal textual and visual feedback to guide the RL training. The rewards are rule-based at textual level and model-based at the visual level.


\hi{LLM-Based Data Synthesis.} A primary challenge in Chart QA is the scarcity of high-quality, multimodal instruction tuning datasets. To address this, several works have focused on {data synthesis}. Using GPT-4, ChartLlama \cite{han2023chartllama} introduces a sophisticated pipeline with three phases successively: Chart data generation, chart figure generation, and instruction data(code, QA pairs, narrative paragraphs, etc.) generation. The pipeline enables the dataset to cover a wide array of chart types and tasks including QA, summarization and even chart editing. Similarly, ChartBench \cite{xu2024chartbenchbenchmarkcomplexvisual} adopts both desensitized and GPT-generated data. For QA diversity, the work provides GPT with 200 different question templates, with human checks to ensure their correctness. \cite{huang2025evochart} presents EvoChart. It is a self-training method with three phases: Compositional chart generation, chart evaluation and refinement, and QA pair generation and training. The three phases will run cyclically until the completion of data synthesis.

\begin{tcolorbox}[colback=white,colframe=black!40,
boxrule=0.5pt,arc=4pt,left=6pt,right=6pt,top=4pt,bottom=4pt]
\textbf{Takeaway (Chart Analysis):} 

\textbf{(O1):} Process charts with different patterns and organize data results~\cite{mittal-etal-1998-describing, reiter-2007-architecture, liu2024chartthinker, masry2024chartgemmavisualinstructiontuningchart}.

\textbf{(O2):} Pay attention to details that users care about while allowing NL interactions~\cite{das2025chartsofthoughtenhancingllmvisualization, ye2024mplugowlmodularizationempowerslarge}.

\textbf{(O3):} Capture semantic information hidden in positional relationships among the elements~\cite{masry2023unichart, wu2024chartinsights, xu2025chartmoemixturediverselyaligned}. 

\textbf{(O4):} Notice the distribution of information in charts, and extend its ability to solve a lot of downstream tasks\cite{han2023chartllama, huang2025evochart}.
\end{tcolorbox}


\subsection{Video}
\label{sec:video}
Video inherently represents evolving spatial content over time, requiring models to jointly capture spatial semantics and temporal dynamics~\cite{wang2024grounded}. Traditional methods use vision backbones for frame-level features followed by pooling, token merging, or attention to model temporal relations. Although effective, these architectures are manually designed, annotation-heavy, and computationally inefficient for long sequences, often losing fine-grained temporal details during compression.


Recent advances integrate LLMs into multi-modal video understanding by representing videos as structured token sequences instead of independent frames. This design improves visual–text alignment and addresses key limitations of prior methods: $(i)$ computational inefficiency via adaptive frame selection and token reduction, $(ii)$ visual fidelity loss through selective filtering of redundant frames, and $(iii)$ weak temporal grounding using modules that encode event order and duration~\cite{shen2024tempme,wang2024grounded}. Rather than replacing vision encoders, lightweight adapters and temporal reasoning components extend their capabilities while maintaining efficiency.

\begin{figure}
   \centering
   \includegraphics[width=1\linewidth]{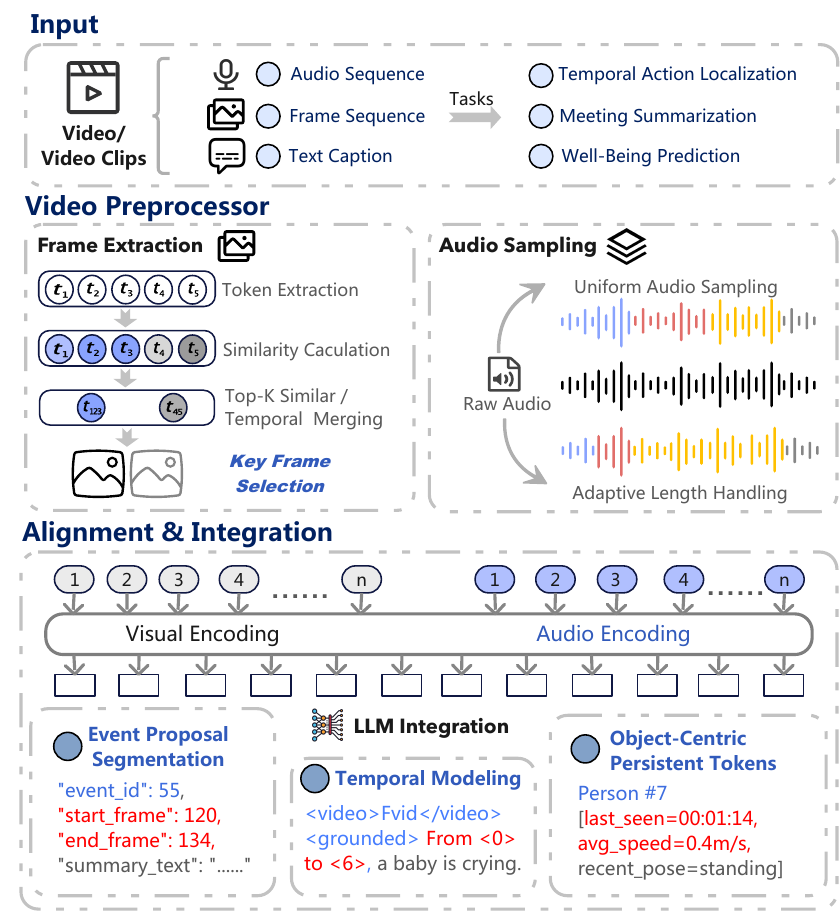}
   \caption{LLM for Video Data Analysis.}
   \label{fig:video-overview}
\end{figure}

Figure~\ref{fig:video-overview} illustrates the general pipeline for LLM-aided video analysis, where existing work primarily focuses on optimizing video preprocessing (e.g., frame extraction, temporal sampling, and feature extraction) and semantic understanding (e.g., temporal modeling, multimodal fusion, and LLM integration). 

\hi{Temporally-Anchored.}
Existing video–language models struggle with precise temporal localization, as it requires detecting sparse, transient signals within long and redundant sequences. This entails $(i)$ distinguish relevant information (e.g., object appearance, key gestures, or sound onsets)~\cite{shen2024tempme}, $(ii)$ identifying event boundaries (e.g., action start/end or scene transitions), and $(iii)$ maintaining temporal coherence by preventing misalignment of object states or actions across time~\cite{chen2024timemarker}. Moreover, handling videos of varying lengths remains challenging, where fixed input sizes waste computation on short clips, and uniform sampling in long videos often overlooks critical events.


To tackle these challenges, Temporally-Anchored methods explicitly align model reasoning with specific time points or intervals. They integrate three key techniques: $(i)$ dynamic sampling, which adaptively selects frames by visual or semantic importance (e.g., \cite{chen2024timemarker} adjusts sampling density based on motion intensity, allocating more frames near action boundaries); $(ii)$ temporal positional encoding, which injects explicit time-aware representations into token sequences (e.g., \cite{wang2024grounded} augments frame embeddings with temporal position codes to model event order and duration); and $(iii)$ flexible tokenization, including training-free similarity-based merging~\cite{shen2024tempme} and learnable saliency-driven merging~\cite{lee2024video}, both designed to retain informative tokens while removing redundancy.


Specifically, TimeMarker~\cite{chen2024timemarker}, a versatile Video-LLM based on LLaMA3-8B, enables high-quality video-grounded dialogue with temporal localization. It introduces separator tokens to represent physical time, temporal separator tokens to enhance temporal awareness, and an AnyLength mechanism for dynamic frame sampling and adaptive token merging which adjusts the frames-per-second (FPS) and visual tokens to match the LLM’s context window and GPU budget.
Similarly, Grounded-VideoLLM~\cite{wang2024grounded} augments the architecture with a temporal stream and discrete time tokens to enable explicit timestamp reasoning, trained through progressive video–language localization tasks. 
Seq2Time~\cite{deng2025seq2time} advances this direction through sequential knowledge transfer, learning temporal grounding in a self-supervised manner by converting ordered image or clip sequences into time-annotated signals. Specifically, 10-second clips from Kinetics-700 are concatenated to simulate event progression in longer videos, each paired with a caption and positional index. The model learns to predict or align their sequential order, internalizing temporal structure without explicit timestamps.

 
\hi{Instruction-Aware Relative Temporal Localization.}
In contrast to temporally-anchored approaches that output absolute timestamps or time-interval tokens (e.g., ``at 00:02:15''), the instruction-aware relative temporal localization focuses on answering relative temporal queries (e.g., ``what happened after the second bell?'').
Early studies typically combined video foundation models (e.g., Conv3D/I3D-style, two-stream, or transformer-based encoders) with text-only LLMs, but these systems relied on heavy pretraining and complex multi-module architectures.


RED-VILLM~\cite{huang2024image} shows that image-based LLMs can be efficiently extended to video understanding by introducing lightweight temporal adapters that align instruction semantics with temporal video features, avoiding full video-model retraining. 
Similarly, IVA-LLM~\cite{li2024llms} integrates a spatial feature interactor and a temporal frame selector within LLM blocks. The selector retrieves query-relevant frames, while the interactor focuses on salient regions or objects within those frames. This design enhances efficiency and visual fidelity under token aggregation while preserving strong temporal grounding aligned with natural language instructions.


\hi{Video Emotional Analysis.}
Video emotion analysis aims to infer affective states, social dynamics, and well-being from multimodal cues. As understanding and promoting well-being is crucial for long-term health and performance~\cite{muller2024predicting}, traditional survey-based assessments are limited by their time-consuming nature and inability to capture the subtle, context-dependent, and synchronized dynamics of visual, auditory, and textual signals.


To provide real-time feedback for behavior awareness, PERMA~\cite{muller2024predicting} predicts individual well-being during teamwork interactions. Panoramic videos are recorded, and for each person, nonverbal cues, including facial expressions, head pose, and 3D gaze, and environmental factors like scene brightness are extracted and summarized into 125 statistical features. These features are aligned with self-reported PERMA survey responses, and machine learning models predict each PERMA pillar as either classification or regression tasks, capturing discrete well-being states.
Another system~\cite{jadhav2023ai} assesses interview candidates by fusing multimodal cues. Faces are detected using HaarCascade classifiers~\cite{adepu2020interviewee}, audio is represented via Mel-frequency cepstral coefficients, and visual and acoustic sequences are modeled with Inception-V3 and LSTM encoders. Textual data undergo standard NLP preprocessing before integration (e.g., tokenization, stop-word removal, sentiment features), enabling joint predictions of personality traits and performance from frame- and clip-level inputs.

\hi{Object Detection.}
Object detection identifies and classifies instances in images by assigning bounding boxes and category labels. In video analysis, it extends to maintain temporal consistency and track objects across frames.


Recent methods isolate region-level tokens and maintain object-specific trajectories over time. 
To capture fine-grained spatial and temporal details, VideoRefer~\cite{yuan2025videorefer} uses a spatial–temporal object encoder with adaptive temporal token merging to generate precise region representations for single- and multi-frame reasoning.

To enhance video analysis precision and efficiency, \cite{de2024video} uses YOLO-V8 for object detection and event-region segmentation, while Gemini-2.5-flash converts these visual cues into structured textual reports. An incident–context fusion pipeline aggregates temporal object trajectories and scene information before input to Gemini, producing summaries that preserve fine-grained entity details and overall context. By combining high-precision, anchor-free detection with multimodal reasoning, the framework reduces manual review while reliably capturing critical events.
\cite{sengonul2025abnormal} proposes a three-stage pipeline for surveillance anomaly detection: $(i)$ preprocessing with background subtraction and noise reduction, $(ii)$ feature learning via an LSTM autoencoder of normal behavior trajectories, and $(iii)$ anomaly detection using distribution-based regularity scoring with local minima analysis.

  
\hi{Gesture and Behavior Detection.}
Gesture and behavior detection aims to identify fine-grained body movements (e.g., hand gestures, nods) as well as higher-level behaviors (e.g., approach/avoidance, turn-taking), typically leveraging pose estimation, temporal classifiers, and interaction models.

Recently, MLLMs have been explored as tools for feature extraction in gesture and behavior detection. To explore how multimodal large language models (MLLMs) can be utilized as part of the multimodal learning analytics (MMLA) process,  ~\cite{whitehead2025utilizing} employed MLLMs to automatically derive postural behavior indicators in a collaborative physics learning task, demonstrating that generative pipelines can reduce reliance on handcrafted annotations while preserving analytical depth. In this framework, MLLMs processed video data to extract features related to postural behaviors, leveraging their capacity to interpret visual and spatial information. 
As a method to generate detailed quantitative measurements of 3D human behavior previously unattainable through manual efforts or 2D methods , HARMONI~\cite{weng2025artificial} integrates three-dimensional mesh estimation with spatial interaction modeling to analyze caregiver–child interactions in naturalistic environments. It first applies video tracking algorithms to isolate individual body trajectories, then employs a deep neural network to estimate 3D human mesh models for each frame, which are subsequently refined through optimization and temporal filtering.

\hi{Video Data for LLM.}
High-quality video–language datasets are critical for large language models (LLMs) to acquire temporal reasoning, multimodal alignment, and grounding capabilities beyond static images or text. Video inherently encodes temporal dynamics, causal relationships, and fine-grained nonverbal signals—all of which are essential for tasks such as action understanding, spatio-temporal reasoning, and multimodal planning which can boost the capability of VLMs. To enable LLMs to capture these capabilities, researchers have developed several complementary approaches to collect or generate video–language data. These approaches can be categorized into four parallel paradigms, each offering unique advantages in terms of annotation fidelity, scalability, and diversity. 

\noindent$\bullet$ \textit{\underline{Human Annotation.}} Human-annotated video datasets offer the highest label fidelity and domain specificity, making them indispensable for tasks where nuanced human judgment is required. 

Gesture and interaction datasets, such as \cite{whitehead2025utilizing} and \cite{weng2025artificial} incorporate manual verificationto ensure the accuracy and relevance of the labeled data.  and curated gestures and interaction detection datasets.
For emotion and behavior analysis, \cite{muller2024predicting} and \cite{jadhav2023ai} combine human annotation  with multimodal feature extraction—covering facial, audio, and textual cues—to produce rich, fine-grained annotations. 
Object detection–oriented datasets like \cite{guo2024xs} utilize  staged annotation processes with multiple rounds of verification to guarantee consistency . 

\noindent$\bullet$ \textit{\underline{Human-in-the-Loop / Assisted Annotation.}} 
Human-in-the-loop pipelines strike a balance between annotation quality and scalability by combining expert oversight with lightweight automation.This approach leverages lightweight automated modules to support annotators. 
Examples include the building process in TimeMarker’s temporal-separator tokens  or the construction of VidThinker's  clip captions and key-frame selection modules. Additionally, annotation workflows often employ lightweight modules to assist human experts. For instance, temporal frame selection can be guided by models like VideoITG’s VidThinker\cite{wang2025videoitg}, an automated pipeline for clip-level captioning, instruction-aware retrieval, and fine-grained frame localization, as demonstrated in \cite{wang2025videoitg} with the VideoITG-40K dataset. Such tools ensure fine-grained alignment across video frames by combining automated support with expert oversight.
    
\noindent$\bullet$ \underline{\textit{Automated Curation} \& \textit{Filtering}.}
Fully automated curation pipelines prioritize scalability and efficiency, generating large datasets while employing built-in quality control mechanisms.
These are end-to-end automated generation pipelines with built-in quality control, often implemented as multi-agent engines with dedicated Reviewer agents, or via detection to LLM flows (such as YOLO-V8 to Gemini). For example, \cite{yuan2025videorefer} employs a multi-agent engine to automatically curate fine-grained spatio-temporal object-level video instruction data for the VideoRefer-700K dataset, where a dedicated ``Reviewer'' agent filters erroneous or mismatched samples. As an example of detection to LLM flows, \cite{de2024video} implements a pipeline that begins with automated object detection using YOLO-V8, followed by the Gemini model to generate incident- and context-aware video summaries, thereby reducing the need for exhaustive manual review.
    
\noindent$\bullet$ \textit{\underline{Synthetic Generation.}} 
Synthetic data generation provides virtually unlimited scalability by creating aligned video–text pairs using generative models.Fully synthetic data generation produces aligned video–text pairs via text-to-video and domain-specific generators, offering high scalability. 
Models such as NUWA~\cite{wu2022nuwa} employ a unified 3D transformer encoder–decoder architecture capable of handling text, images, and videos in a single framework for multimodal conditional generation. NUWA-Infinity extends this to arbitrarily long or high-resolution sequences using a hierarchical autoregressive ``patch-over-patch'' strategy with context caching and position control. Text2Video-Zero~\cite{khachatryan2023text2video} demonstrates zero-shot text-to-video generation using Stable Diffusion without additional training, enriching latent codes with motion dynamics and reprogramming frame-level self-attention via cross-frame attention anchored to the first frame for temporally consistent, high-quality videos. Align Your Latents~\cite{blattmann2023align} converts an image-only latent diffusion model into a video generator by adding a temporal dimension to the latent space and fine-tuning only temporal components, enabling high-resolution, temporally coherent generation. Domain-specific generators such as SadTalker~\cite{zhang2023sadtalker} and DreamTalk \cite{ma2023dreamtalk} synthesize talking-head videos from audio, while choreography-guided systems~\cite{wang2024disco} produce human dance sequences from pose trajectories. High-fidelity text-to-video frameworks, including \cite{ho2022imagen} and Make-A-Video~\cite{singer2022make}, accelerate large-scale creation of scene-specific clips by extending powerful text-to-image backbones with spatial–temporal modules, substantially reducing the need for manual annotation. Collectively, these synthetic methods provide a scalable and automated alternative to traditional pipelines, delivering richly diverse, automatically aligned training data that accelerates the development of multimodal LLMs. 

\begin{sloppypar}
\begin{tcolorbox}[colback=white,colframe=black!40,
  boxrule=0.5pt,arc=4pt,left=6pt,right=6pt,top=4pt,bottom=4pt]
\textbf{Takeaway (Video Analysis):} 

\textbf{(O1):} Processing spatial semantics and temporal dynamics, multi-modal cues for emotional analysis, and 3D human behavior~\cite{wang2024grounded,weng2025artificial}. 

\textbf{(O2):} {Driven through natural language using methods including  adapters for instruction-conditioned reasoning~\cite{chen2024timemarker,li2024llms}.}

\textbf{(O3):} {Perform instruction-aware relative temporal localization to recognize conceptual relationships and contextual details~\cite{huang2024image,de2024video}.}

\textbf{(O4):} {Advance through adaptive methods like dynamic frame sampling and token merging~\cite{chen2024timemarker,yuan2025videorefer}.}
\end{tcolorbox}
\end{sloppypar}

\subsection{Documents}
\label{sec:document}

Documents present information with complex, non-sequential layouts, incorporating visual elements such as tables, lists, figures, and footnotes alongside standard text (e.g., PDFs, web pages). 
which involves tasks like accurate {factual question answering} by locating specific information amidst layout noise, generating {concise summaries}, and performing {key-value extraction} from structured content. For {long-form documents} (e.g., scientific papers, technical reports), the challenges escalate to {integrative comprehension} and {cross-document synthesis}, along with a robust capability for {handling distractors}. Finally, handling {multilingual corpora} requires not only {high-fidelity translation} but also complex downstream tasks like {cross-lingual QA or summarization}.

The primary challenge lies in parsing this intricate combination of structural, spatial, and semantic information into a coherent, machine-readable representation (e.g., vectors)~\cite{xu2021layoutlmv2, huang2022layoutlmv3}.


To tackle these challenges, the research community has developed several distinct technical paradigms. We categorize them into two primary approaches: $(i)$ visual-based analysis, ranging from unimodal to multimodal designs, with particular emphasis on fusion strategies and cross-modal interactions, $(ii)$ text-based analysis for enhancing factual accuracy and context.
Finally, we introduce the synthetic data generation techniques to address data scarcity and improve model training for document understanding.

\hi{Architecture Design for Document Understanding.} 
The evolution of document understanding architectures reflects a fundamental shift from processing individual modality in isolation to multimodal integration. Early approaches treated text, layout, and visual elements as separate channels, which often led to suboptimal performance due to the loss of critical cross-modal relationships inherent in document structure. Unlike plain text processing, document understanding inherently requires the simultaneous consideration of textual content, spatial layout, and visual appearance. This realization has driven the field toward multimodal architectures that can capture these complex interdependencies.

\noindent$\bullet$ \textit{\underline{Multimodal Fusion Strategies.}} 
The effectiveness of multimodal document understanding critically depends on how different modalities are integrated. {Early fusion} concatenates features at the input level; for example, {LayoutLM}~\cite{xu2020layoutlm} creates a single input representation for each token by combining textual embeddings with 1D sequential and 2D spatial position embeddings derived from bounding boxes. {Intermediate fusion} learns modality-specific representations separately before integrating them at deeper network layers, as exemplified by the dual-stream designs of {LayoutLMv2}~\cite{xu2021layoutlmv2} and {DocFormer}~\cite{appalaraju2021docformer}. By contrast, {late fusion} combines predictions from separately trained models; the shared-backbone, multi-head design of {DLAFormer}~\cite{wang2024dlaformer} can be viewed as a variant of this strategy.

\noindent$\bullet$ \textit{\underline{Cross-Modal Interaction Designs.}}
Beyond fusion timing, cross-modal interaction mechanisms determine how effectively models capture fine-grained relationships. Attention-based designs such as {VLCDoC}~\cite{bakkali2023vlcdoc} adopt a two-step ``align-then-integrate'' mechanism: Inter-Modality Cross-Attention (InterMCA) first establishes semantic links between modalities (e.g., using text queries to attend to visual keys and values to obtain visually enhanced text representations) followed by Intra-Modality Self-Attention (IntraMSA) to fuse the enriched representations with each modality’s own context. Prompt-guided methods like {VisFocus}~\cite{abramovich2024visfocus} inject natural-language queries directly into the visual encoder, where cross-attention generates a weight map that highlights task-relevant regions. Coordinate-aware designs such as {CREPE}~\cite{okamoto2024crepe} further extend interaction by triggering a parallel coordinate-decoding head during autoregressive text generation, allowing the model to directly regress the bounding box of generated text when a special $</ocr>$ token is emitted.

\vspace{1em}
\noindent$\bullet$ \textit{\underline{Specialized Architectural Innovations.}}
Recent work explores specialized designs to improve universality, efficiency, and reasoning capability. End-to-end unified frameworks like {DLAFormer}~\cite{wang2024dlaformer} jointly handle layout detection, role classification, and reading-order prediction as a single relation-prediction problem, taking a document image as input and outputting detected objects with class and relation information. To address the need for task-agnostic solutions, {OmniParser}~\cite{wan2024omniparser} proposes a universal framework that unifies outputs across tasks into three sequences (i.e., center points, polygonal shapes, and text contents) and decodes key–value pairs through a two-stage process that first predicts the center point of a value and then conditions on it to generate its boundary and text. Efficiency-oriented approaches such as {Rationale Distillation}~\cite{hsieh2024rationale} train a smaller model (e.g., Pix2Struct) to predict both final answers and intermediate reasoning steps of a larger model, optimizing a weighted combination of rationale and answer losses. Finally, multi-document reasoning systems like {VisDoM}~\cite{lei2024visdommultimodal} adopt a dual-pipeline RAG architecture, retrieving evidence separately from visual and textual sources and fusing them only when their consistency exceeds a predefined threshold.

\noindent$\bullet$ \textit{\underline{Robustness to Modality Variations.}} Real-world document understanding systems must remain effective when one or more modalities are missing or corrupted. To address this challenge, {MissModal}~\cite{ma2023missmodal} introduces a missing-modality adaptation strategy that aligns representations across different modality combinations. By minimizing the distributional distance between features derived from complete data and those obtained from incomplete inputs, the model enforces consistency within a shared semantic space. Complementarily, {MMP}~\cite{huang2024mmp} explores masked-modality learning by training a projection function to generate ``pseudo'' representations for absent modalities (e.g., vision) using information from the remaining ones (e.g., text and audio). This dynamic projection enables the model to reconstruct missing signals during inference, thereby preserving performance even under severe modality dropouts.

\hi{Visual-Based Document Understanding.} 
The diverse types of elements in documents challenge the accurate document understanding. To solve the challenge, an intuitive way is to apply VLMs to enable visual understanding over such documents. 

DocOwl1.5~\cite{hu2024mplugdocowl15} is a model for visual document understanding trained on document, chart, and table data. It designs a vision-to-text module which maintains the layout information and reduces the length of visual features through merging horizontal adjacent patches through convolution.
Based on DocOwl1.5, DocOwl2~\cite{hu2024mplugdocowl2} incorporates a layout-aware compress module to compress each high-resolution document image into a fixed length of 324 tokens, which further improves the efficiency. 
For multi-page documents, M3DOCRAG~\citep{M3DOCRAG} uses ColPali~\cite{faysse2025colpali} to embed document pages and the query, and compute cross-modal similarity to retrieve top-k relevant pages for multi-page visual inference.
SV-RAG~\citep{SV-RAG} fine-tunes a VLM to first search for relevant page regions and then generate an answer from the retrieved visual evidence.
VisDoM~\citep{lei2024visdommultimodal} implements parallel visual and textual retrieval branches and synthesizes evidence from both to formulate the final answer, improving reasoning by training models to focus on relevant retrieved context while ignoring distractors \citep{zhang2024raft, yan2024corrective}.

\hi{Text-Based Document Understanding.}
Apart from the visual-based methods, another approach is to convert the document into pure text format and perform text-based inference.

ZenDB~\cite{lin2024zendb} assumes that many documents within a collection share similar templates that encode a common semantic structure. Accordingly, it first performs layout analysis to segment document elements, then applies visual clustering and LLM-based semantic analysis to construct a document tree. Visual elements such as charts and tables are converted into textual descriptions, enabling text-based RAG to retrieve relevant information for question answering.
EVAPORATE~\cite{arora2023language} transforms text-based documents into relational data by prompting LLMs to generate table schemas relevant to the query, synthesize functions to retrieve the necessary data, and aggregate the results into a unified relational table, which is then used to answer the original question.
Similarly, QUEST~\cite{sun2025quest} builds retrieval indices for both documents and their segments by summarizing and embedding their contents, thereby enabling document querying through SQL-like operations.
UQE~\cite{dai2024uqequery} accepts queries in a Universal Query Language (UQL), a dialect of SQL that provides full natural language flexibility in specifying conditions and operators. The UQL is applied on the unstructured text data for data retrieval and question answering. 
AOP~\cite{wang2025aop} defines semantic operators for building execution workflows (e.g., semantic retrieval, filtering, aggregation). Given an online query, AOP extracts relevant operators and uses these operators to automatically and interactively compose optimized pipelines with the assistance of LLMs to ensure high accuracy question answering.
DocETL~\cite{shankar2025docetl} is a document agent which provides a declarative YAML-based interface to define pipelines with LLM-specific operators. It first logically rewrites the analysis pipeline and evaluates the plan guided by a LLM agent. 
An optimization algorithm is applied to find promising plans, considering the latencies of agent-based plan generation and evaluation.

\hi{Document Synthetic Generation for LLM.}
Beyond processing existing documents, a burgeoning area of research focuses on using generative models to synthesize them. For instance, \cite{pisaneschi2023automatic} adopts transformer-based models to generate plausible scientific paper layouts. A particularly clever approach, \texttt{PosterLlama} \cite{seol2024posterllama}, reformats the layout generation task as HTML code generation. Instead of outputting coordinates directly, it instructs an LLM to generate HTML and CSS code describing the document layout, leveraging the model's inherent knowledge of web layout languages. To overcome one-shot generation limitations, \texttt{LayoutCoT} \citep{shi2025layoutcot} uses Chain-of-Thought (CoT) prompting for iterative refinement, while \texttt{VASCAR} \citep{zhang2024vascar} introduces a visual self-correction loop. Meanwhile, diffusion models like \texttt{LayoutDM} \citep{inoue2023layoutdm} and \texttt{LDGM} \citep{hui2023ldgm} offer strong controllability. The frontier is moving towards aligning outputs with human aesthetic judgments, as exemplified by \texttt{AesthetiQ} \citep{patnaik2025aesthetiq}, supported by new datasets like \texttt{SciPostLayout} \citep{tanaka2024scipostlayout} and metrics like \texttt{LTSim} \citep{otani2024ltsim}.

\begin{tcolorbox}[colback=white,colframe=black!40,
  boxrule=0.5pt,arc=4pt,left=6pt,right=6pt,top=4pt,bottom=4pt]
\textbf{Takeaway (Document Analysis):} 

\textbf{(O1):} Multimodal LLM architectures~\citep{xu2020layoutlm,wang2024docllm} holistically integrate textual, spatial, and visual information from complex layouts.

\textbf{(O2):} Prompt-guided interaction where natural language queries direct visual analysis~\citep{abramovich2024visfocus}.

\textbf{(O3):} Employing advanced RAG systems to retrieve and synthesize information from both text and images, providing context-aware answers and summaries~\citep{lewis2020retrieval, lei2024visdommultimodal}.

\textbf{(O4):} Iterative refinement and self-correction progressively improve the quality of document analysis~\citep{shi2025layoutcot, zhang2024vascar}.
\end{tcolorbox}

\begin{figure*}[!t]
   \centering
   \includegraphics[width=0.95\linewidth]{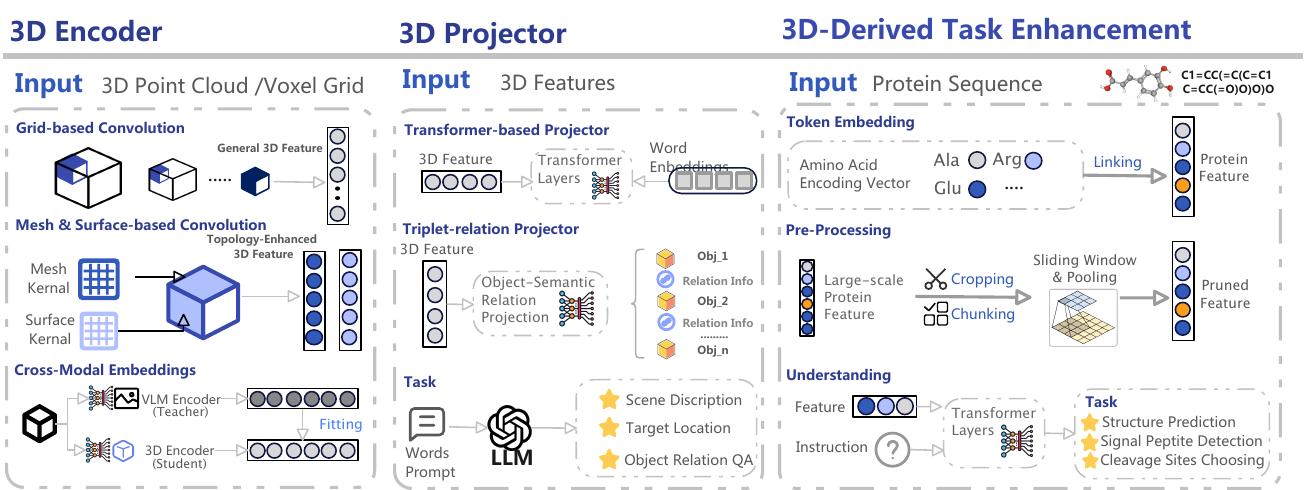}
   \caption{LLM for 3D Model Analysis.}
   \label{fig:3d-overview}
\end{figure*}

\subsection{3D Models}
\label{sec:3d}
3D models are digital representations of objects, scenes, or structures embedded in three-dimensional Euclidean space~\cite{wikipedia_3dmodel}. They are defined by explicit geometric data (e.g., point clouds, meshes, voxels, implicit surfaces) and may be further enriched with semantic or physical attributes (e.g., textures, materials, part-level annotations). Serving as a fundamental medium for spatial perception, reasoning, and interaction, 3D models provide the basis for computational systems to support tasks such as scene understanding, navigation, planning, and content creation.

Traditional 3D model analysis involves using 3D modeling software (e.g. Blender, 3ds Max, Maya) and need human input for manipulation, annotation, and interpretation.
As LLMs evolve, their integration with 3D spatial data enables unprecedented capabilities in understanding and interacting with physical environments. This section provides a comprehensive overview of methodologies that allow LLMs to process, interpret, and generate 3D data from point clouds to Neural Radiance Fields (NeRFs), and examines their applications in tasks such as scene understanding, captioning, question-answering, and dialogue. We further discuss the role of LLM-powered agents in spatial reasoning, planning, and navigation, highlighting their potential to bridge natural language with the inherent complexities of 3D spaces.
As shown in Figure~\ref{fig:3d-overview}, the fusion of LLMs with 3D data primarily manifests in three key areas: 3D Language Fusion, which aligns 3D geometry with language descriptions; 3D-Derived Task Enhancement, which leverages 3D information to improve performance on downstream tasks; and Cross-Modal Capability Refinement, which uses existed experience on models (like 2-D image models) to refine 3D model's understanding and generation abilities across different modalities.

\hi{3D-Language Fusion.} Fusing three-dimensional data representations with the semantic space of natural language aims to establish a unified modality that enables models to seamlessly interpret, reason about, and generate grounded descriptions of physical 3D structures. This process faces several challenges, including the inherent sparsity and irregular nature of 3D data (like point clouds), the significant computational cost of processing large 3D models, and the difficulty of establishing a precise semantic alignment between geometric features and linguistic concepts.  As shown in Figure~\ref{fig:3d-overview}, the process includes using a 3D-Encoder to capture spatial features and relationships at first,and then uses a projector to align 3D features with text space so that the 3D model features can be injected into LLM well.

\noindent$\bullet$ \textit{\underline{Fusion with Point Clouds.}}
3D-LLM~\cite{hong20233d} pioneered this by converting point clouds or scenes into multi-view renderings, extracting per-view visual tokens with a frozen vision–language encoder, and aggregating them into 3D-aware tokens that are fed to the LLM, which attaches 3D positional embeddings and outputs location tokens aligned with language. 
Scaling to direct point-cloud input, 3UR-LLM~\cite{xiong20253ur} proposes an end-to-end 3D multimodal large language model that ingests raw point clouds, fuses them with text instructions, and projects them into a compact set of vision tokens via a 3D Compressor, which utilizes a multi-layer Transformer architecture with learnable queries and cross-attention mechanisms to interlink visual and textual features, and a 3D Query Fusion mechanism,which selects high-confidence queries from the perception component based on objectness probability and integrates them with the original queries from the compressor. 

\noindent$\bullet$ \textit{\underline{Domain-Specialized Fusions.}}
For scientific domains, 3D-MoLM~\cite{li2024towards} equips an LLM with a dedicated 3D molecular encoder (typically a graph-based neural network that encodes atomic coordinates and conformations)  and a learnable molecule-text projector (usually implemented as a lightweight MLP or transformer adapter ) to encode atomic coordinates and conformations into language-context embeddings. The model is trained on a 3D molecule-text dataset to learn cross-modal retrieval, captioning, and open-ended QA. 

ProteinChat\cite{guo2023proteinchat} is designed to enable ChatGPT-like conversational functionality centered on protein 3D structures. Its architecture follows a four-part blueprint: (i) a protein encoder that derives sequence and 3D embeddings; (ii) a Protein–Language Pretraining Transformer (PLP-former), a Transformer-based module trained with contrastive and self-supervised objectives to align protein and language representations; (iii) a projection adapter, typically a lightweight neural network trained jointly with the PLP-former while keeping the LLM frozen, which converts protein embeddings into prompts suitable for language generation; and (iv) a foundation LLM (Vicuna-13B) that generates language-grounded responses.

Building on the framework of \cite{guo2023proteinchat}, ProtChatGPT \cite{wang2024protchatgpt} extends ProteinChat by introducing a three-stage pipeline:3D feature extraction, protein–language pretraining, and instruction tuning—to further enhance grounded question answering and explanatory capabilities for protein structures. This enhancement comes from its integration of both sequence- and structure-based protein encoders, contrastive pretraining that better aligns protein representations with textual descriptions, and instruction tuning that adapts the LLM for domain-specific reasoning, leading to more accurate and interpretable protein-grounded responses.


\begin{table*}[t!]
\centering
\caption{Comparison of 3D Generation Methods across different paradigms.}
\label{tab:3d-compare}
\resizebox{\textwidth}{!}{ 
\begin{tabular}{ l l l l l l l }
\hline
\textbf{Method} & \textbf{Paradigm} & \textbf{Uses 2D Prior} & \makecell{\textbf{Multi-view} \\ \textbf{Consistency}} & \textbf{Lantency} & \makecell{\textbf{Editable} \\ \textbf{Output}} & \textbf{Key Techniques} \\
\hline
Fantasia3D~\cite{chen2023fantasia3d} & Optimization-based & \checkmark & Medium & Slow (minutes) & \checkmark & \makecell[l]{SDS + DMTet mesh + BRDF} \\
\hline
SweetDreamer~\cite{li2023sweetdreamer} & Optimization-based & \checkmark (tuned) & High & Slow (minutes) & \checkmark& \makecell[l]{SDS + Aligned Geometric Priors} \\
\hline
RichDreamer~\cite{qiu2024richdreamer} & Optimization-based & \checkmark (multi-prior) & High & Slow (minutes) & \checkmark & \makecell[l]{SDS + Normal/Depth/Albedo Priors} \\
\hline
Zero-1-to-3~\cite{liu2023zero} & Feed-forward + Reconstruction & \checkmark & Medium & Fast (seconds) & Limited & \makecell[l]{View-conditioned diffusion + NeuS}\\
\hline
Hunyuan3D-1.0~\cite{yang2024hunyuan3d} & Feed-forward + Reconstruction & \checkmark & High & Fast (seconds) & Limited & \makecell[l]{Multi-view diffusion \\+ Sparse-view Reconstructor}\\
\hline
CraftsMan3D~\cite{li2024craftsman3d} & Native 3D Generation & -- & High & Fast (seconds) & \checkmark& \makecell[l]{3D Latent Diffusion} \\
\hline
Direct3D~\cite{wu2024direct3d} & Native 3D Generation & -- & High & Fast (seconds) & \checkmark& \makecell[l]{D3D-VAE + D3D-DiT} \\
\hline
MeshAnything~\cite{chen2024meshanything} & Native 3D Generation & -- & High & Fast (seconds) & \checkmark & \makecell[l]{VQ-VAE Mesh Codebook \\+ Transformer} \\
\hline
LLaMA-Mesh~\cite{wang2024llama} & Native 3D Generation & -- & High & Fast (seconds) & \checkmark& \makecell[l]{LLM Autoregressive Mesh Tokens} \\
\hline
\end{tabular}
}
\end{table*}

\hi{Protein-domain Task Enhancement}
Unlike 3D-Language Fusion which focuses on aligning spatial features with textual semantics, 3D-derived task enhancement emphasizes leveraging structural representations to improve downstream tasks that inherently require geometric reasoning  (e.g., spatial question answering) as shown in Figure~\ref{fig:3d-overview}. The central challenge lies in $(i)$ achieving genuine structural understanding rather than relying on superficial correlations, and $(ii)$ ensuring scalability, since incorporating high-dimensional 3D signals into large models can quickly lead to prohibitive memory and computation costs.

To evaluate LLMs' understanding of geometric structures , GeomRel~\cite{wang2025large} identifies failure modes,including inaccurate identification of geometric relations, lack of structured reasoning  in large language models (LLMs) by constructing a benchmark that isolates geometric relationship identification as a core subtask. To address this challenge, it encodes problems with structured relation pools (e.g., lines, angles, congruences) and labels target relations directly to facilitate accurate identification. For reasoning abilities, GeomRel introduces the Geometry Chain-of-Thought (GeoCoT), a two-stage prompting pipeline. The first stage forces enumeration and verification of candidate relations with reverse-checks for consistency, and the second stage performs rigorous reasoning over these relations. Ablations show that explicit supervision of relation identification helps LLMs learn spatial-structural semantics rather than producing numerically correct but conceptually hollow answers. 

For molecular representation learning, GPT-MolBERTa~\cite{wang20253dsmiles} replaces compact SMILES~\cite{weininger1988smiles} strings, which linearize molecular graphs into text sequences, with rich textual molecule descriptions. The model is pretrained on 326k molecules with natural language captions to learn embeddings that capture geometric cues and the biological co-evolution to predict the property of target protein.

At system levels, ProtChat~\cite{huang2024protchat} integrates specialized Protein Large Language Models (PLLMs) into practical workflows with a multi-agent orchestration framework. A high-capacity planner LLM (GPT-4) decomposes user instructions, dispatches domain-specific PLLMs (e.g., MASSA~\cite{hu2023multimodal}), and aggregates their outputs for tasks such as protein property prediction and protein-drug interaction profiling. The system standardizes prompt-tool contracts, result formatting, uncertainty reporting, and error handling, enabling non-expert users to leverage 3D biological models through a single natural-language interface without writing complex scripts. 


\hi{Cross-Modal Capability Refinement.}
As illustrated in Figure~\ref{fig:3d-overview}, cross-modal alignment seeks to establish a shared representational space that enables seamless interaction among heterogeneous modalities, such as 2D images, 3D assets, and natural language, by leveraging knowledge transferred from other modalities. However, several challenges arise in this process, including the semantic gap between low-level geometric features and high-level linguistic concepts, as well as differences in data formats (e.g., continuous 3D point clouds versus discrete textual tokens).

To handle  the limitation of domain adaptation technique in learning discriminative features for target samples due to the lack of label information in target domain, Song et al.~\cite{song2023self} improve image-mesh retrieval by combining adversarial domain alignment\ with self-supervised contrastive learning on unlabeled 3D data. The objective integrates $(i)$ supervised discriminative learning on labeled 2D images, $(ii)$ contrastive losses on multi-view 3D renderings that pull views of the same object together and push different objects apart, and $(iii)$ adversarial terms to minimize domain gaps between 2D and 3D embeddings. An entropy-aware memory bank maintains representative negative samples and reduces false negatives, producing highly discriminative 3D embeddings and boosting retrieval accuracy. 

To equip multimodal LLMs with 3D perception from single images, LLMI3D~\cite{yang2024llmi3d} augments a pretrained MLLM using parameter-efficient fine-tuning (PEFT) and three spatial reasoning modules including $(i)$ Spatial-Enhanced Local Feature Mining sharpens object boundary and micro-geometry features via multi-scale attention with positional cues, $(ii)$ 3D Query Token-Derived Info Decoding introduces explicit geometric-query tokens with regression heads for precise numeric output, and $(iii)$ Geometry Projection-based Reasoning corrects for focal-length and perspective distortion by projecting features through differentiable camera models. The result shows that carefully designed spatial modules with PEFT can make general MLLMs competitive for 3D geometric reasoning without full-model retraining.

\hi{3D Data for LLM.}
Large language models have shown remarkable capabilities in reasoning over text and 2D imagery, but extending them to 3D domains introduces unique challenges. Unlike text or images, 3D data inherently encodes geometry, spatial relations, and multi-view consistency, which makes it both richer in information and more difficult to model. Recent research has therefore explored different paradigms for leveraging or directly generating 3D representations conditioned on language, aiming to unlock applications such as text-to-3D asset creation, embodied AI, and interactive design. Table~\ref{tab:3d-compare} Existing methods can be broadly grouped into three categories: optimization-based approaches that adapt powerful 2D diffusion priors for 3D generation, feed-forward pipelines that prioritize speed and determinism, and emerging native 3D generative models that attempt to learn 3D distributions directly.

\noindent$\bullet$ \textit{\underline{Optimization-Based.}}
This paradigm leverages pre-trained 2D diffusion models as supervisory signals to directly optimize a 3D representation (e.g., NeRF or mesh). A key technique here is Score Distillation Sampling (SDS), which transfers the generative prior of 2D diffusion models into the 3D domain. In details , SDS adds noise to images rendered from the current 3D representation,and then uses the diffusion model to predict the denoising direction conditioned on the text prompt,  then distills this signal into a gradient that updates the 3D parameters. Over many iterations, the 3D asset is shaped so that its 2D projections from diverse viewpoints are judged as likely given the prompt. 

The primary advantage of this approach is its high quality ceiling:it can synthesize novel, intricate textures and surfaces by tapping into the rich visual priors of 2D diffusion models . However, it suffers from two major drawbacks: slow per-instance optimization (often taking tens of minutes) and multi-view inconsistency. The latter arises because the 2D prior lacks an inherent understanding of 3D consistency, often providing conflicting guidance from different views and leading to artifacts like the ``Janus head'' problem.

To address such challenges, Fantasia3D~\cite{chen2023fantasia3d}  tackled the issue of blurred geometry in volumetric representations by explicitly disentangling geometry from appearance. It represents geometry with an explicit DMTet mesh and models appearance via a spatially-varying BRDF material model. During training, it separately optimizes geometry using normal map renderings and appearance using shaded renders via an SDS loss, resulting in high-fidelity, relightable assets. To combat the multi-view inconsistency problem, SweetDreamer~\cite {li2023sweetdreamer} proposed innovating on the {prior itself} rather than just the optimizer. It fine-tunes a 2D diffusion model on Objaverse renderings to predict view-conditioned canonical coordinate maps, termed Aligned Geometric Priors. This embeds multi-view consistency directly into the diffusion model, leading to more stable SDS gradients and reduced failures. Building on this, RichDreamer~\cite{qiu2024richdreamer} pushed prior specialization even further. Instead of relying on a single RGB diffusion model, it trains geometry-specific  priors: one for normal maps, one for depth, and another for albedo (to factor out lighting). Applying SDS with these geometry-specific priors provides richer, higher-frequency supervisory signals, dramatically improving multi-view detail and surface fidelity compared to using a standard RGB-based SDS loss.

\noindent$\bullet$ \textit{\underline{Feed-Forward Generation with Reconstruction.}} This approach completely bypasses the slow, iterative SDS optimization process. Instead, it breaks the problem into two distinct, feed-forward stages: first, a multi-view synthesis model generates a set of coherent novel views from a single input image, and second, a fast 3D reconstruction network consumes these generated views to produce a 3D mesh in a single forward pass.

The key advantage is speed: generation time is reduced from minutes to seconds, offering deterministic behavior that is crucial for practical production pipelines. The trade-off is a loss of flexibility. The final output quality is bounded by the capabilities of the reconstruction network, and errors from the synthesis stage can propagate and accumulate in the reconstruction stage, potentially limiting the diversity and optimality of results.

This line of work focuses on improving the quality and coherence of the synthesized views to enable robust reconstruction. Zero-1-to-3~\cite {liu2023zero} laid a foundational stone by training a viewpoint-conditioned 2D diffusion model to generate novel views from a single image. These synthesized views, while not perfectly consistent, are sufficiently coherent to be fed into an off-the-shelf reconstruction network like NeuS\cite{wang2021neus} to produce a 3D mesh, all within seconds. Hunyuan3D-1.0~\cite{yang2024hunyuan3d} employs a two-stage ``synthesis-then-reconstruct'' pipeline. It first uses a multi-view diffusion model to generate coherent RGB views from a single image in about four seconds, then applies a learned sparse-view reconstructor to predict high-quality geometry in a single forward pass—emphasizing speed and determinism for rapid asset creation.


\noindent$\bullet$ \textit{\underline{Native 3D Generation.}}
This emerging paradigm seeks to move beyond reliance on 2D diffusion priors altogether. The goal is to train generative models directly on large datasets of 3D assets, allowing them to learn the underlying distribution of 3D geometry and appearance natively. This requires designing efficient neural representations for 3D data, such as compact latent spaces or discrete tokens.

The primary advantage is inherent 3D consistency, eliminating multi-view issues from the root. It also promises fast generation and native support for 3D editing tasks. The main disadvantage is its heavy dependence on large-scale, high-quality 3D datasets for training, which are scarce compared to 2D image data. The technical complexity and computational cost of building these architectures are also currently very high.

Research is branching into different sub-architectures for native generation. One branch focuses on 3D diffusion. CraftsMan3D~\cite{li2024craftsman3d} performs diffusion directly in a latent space that represents both mesh and volumetric properties, enabling the generation of high-fidelity and editable meshes.To capture intricate geometry distributions, Direct3D~\cite{wu2024direct3d} introduces an architecture comprising a D3D-VAE that encodes a 3D asset into a compact latent representation, and a D3D-DiT (Diffusion Transformer) that learns to generate these latents, completely bypassing 2D priors and SDS. Another branch explores 3D tokenization, reframing 3D generation as a language modeling task.To optimize mesh extraction methods to no longer rely on dense faces and focus on geometric features , MeshAnything~\cite{chen2024meshanything} learns a discrete codebook of mesh components using a VQ-VAE, allowing a transformer to then autoregressively decode a sequence of tokens into a high-quality mesh. LLaMA-Mesh~\cite{wang2024llama} takes this concept to its logical extreme by fine-tuning a large language model (LLM) to directly output the vertices and faces of a mesh as a sequence of text tokens, effectively making the mesh a first-class citizen in language-conditioned generation and enabling fascinating applications like conversational mesh editing.

\begin{sloppypar}
\vspace{-0.2cm}
\begin{tcolorbox}[colback=white,colframe=black!40,boxrule=0.5pt,arc=4pt,left=6pt,right=6pt,top=4pt,bottom=4pt]
\textbf{Takeaway (3D-Model Analysis):} 

\textbf{(O1):} {Interpret point clouds and meshes using 3D encoders and projection adapters~\cite{hong20233d, xiong20253ur} to fuse 3D spatial features with natural language semantics.}

\textbf{(O2):} {LLMs function as high-capacity planners~\cite{huang2024protchat} to decompose tasks and interact with natural language.}

\textbf{(O3)}: {LLMs perform geometric relation identification and structured reasoning to translate complex 3D structures into textual descriptions~\cite{wang20253dsmiles}.}

\textbf{(O4)}: Agentic solutions for 3D model analysis needs further research.

\end{tcolorbox}
\vspace{-0.2cm}
\end{sloppypar}

%% file: sections/heterogeneous-data.tex
\section{Heterogeneous Data Analysis}
\label{sec:heterogeneous-data}

Heterogeneous data refers to the combination of diverse data types (e.g., relational data, semi-structured tables, document images) that exist in different formats. 
Effective analysis of such data requires the ability to capture both structural and semantic information, as well as the interrelationships among different modalities. 
Traditional approaches to heterogeneous data analysis focus on the construction of data lake systems for storage and retrieval. In these systems, analysts are responsible for writing retrieval queries and employing human-in-the-loop methods to conduct analyses, which are often labor-intensive and prone to error. 
Recent advances in LLMs and MLLMs offer new opportunities to analyze heterogeneous data through natural language interactions, thereby reducing human intervention. 
Current research primarily focuses on modality alignment across diverse data types and the development of heterogeneous data analysis agents.

\hi{LLM for Modality Alignment.}
Modality alignment refers to the process of mapping heterogeneous data types into a shared representation space that preserves both structural and semantic information.
The main challenge lies in bridging the semantic gap among modalities with inherently different formats and levels of abstraction, while minimizing information loss.

Unicorn~\cite{fan2024unicorn} addresses the fundamental challenge of modality alignment, that determines whether two data objects convey the same semantic meaning. 
It employs DeBERTa~\cite{he2020deberta} to encode serialized pairs (e.g., text, table columns, tuples, or knowledge graph entities), incorporates a Mixture-of-Experts (MoE) architecture for feature alignment, and leverages an MLP-based binary classifier to assess whether the two objects match.
Symphony~\cite{chen2023symphony} introduces a novel approach from a natural language perspective. The framework offers question–answering capability tailored for heterogeneous data lakes encompassing text, databases, tables, and knowledge graph entities. It transforms data from diverse modalities into natural language summaries, which are subsequently encoded into vector representations. Questions are encoded within the same vector space, and cosine similarity is employed to measure their alignment with the data representations, thereby enabling the retrieval of relevant information for answer generation.

\hi{LLM for Heterogeneous Data Retrieval.} Data retrieval focuses on extracting task-relevant or related information from underlying data sources. The core challenge in heterogeneous data settings lies in the identifying and understanding of relationships both within and across different modalities.

Early efforts primarily targeted text-based modalities. For example, LOTUS~\cite{patel2024lotus} introduces a declarative programming interface that supports operations such as top-k search over data columns and the construction of similarity indexes for individual data items. Similarly, Wang et al.~\cite{wang2025towards} further abstract the retrieval process as a sequence of high-level APIs under a SQL-like paradigm, enabling operations such as selection, projection, join, and aggregation. While effective, these approaches remain confined to text data. Extending beyond this limitation, CAESURA~\cite{urban2023caesura} incorporates visual modalities by embedding visual question answering models into retrieval operations, thereby enabling multimodal queries such as counting the number of images relevant to a given query.

\hi{Heterogeneous Data Analysis Agents.}
A heterogeneous data analysis agent is an intelligent system that leverages LLMs or VLMs to automatically interpret, retrieve, and reason over heterogeneous data through natural language interaction.
The key challenge is to ensure accurate, context-aware analysis while reducing reliance on human intervention, particularly in scenarios involving complex cross-modal reasoning.

XMODE~\cite{nooralahzadeh2024explainable} assumes that heterogeneous metadata is organized within a database and leverages LLMs to decompose natural language queries into subtasks such as text-to-SQL generation and image analysis.
For instance, given the query ``Plot the number of paintings that depict war in each century'', XMODE generates an operation sequence comprising: $(i)$ retrieving painting metadata from the database via SQL, $(ii)$ applying VLMs to determine whether each image depicts war, and $(iii)$ producing visualization code.
In practice, however, heterogeneous data is often disorganized. To address this challenge, Wang et al.~\cite{wang2024interactive} encode data objects using modality-specific models (e.g., BERT for text, CLIP for images) to construct an index. They further introduce a vector weight learning model that adaptively adjusts modality weights to better capture their relative importance in similarity measurement~\cite{wang2024must}. The user query is used to retrieve relevant data through semantic similarity for final answer generation.
AgenticData~\cite{sun2025agenticdata} is a multi-agent framework designed to facilitate cross-domain data analysis through natural language interactions. It begins by generating a logical plan in a single step, utilizing a plan validator, a plan optimizer, and a plan executor to refine and carry out the plan. The framework then applies tailored strategies depending on the data type. Additionally, it supports a variety of relational operators (e.g., filter and project) as well as semantic operators (e.g., scan and extract).

\begin{tcolorbox}[colback=white,colframe=black!40,boxrule=0.5pt,arc=4pt,left=6pt,right=6pt,top=4pt,bottom=4pt]
\textbf{Takeaway (Heterogeneous Data):}

\textbf{(O1):} Capture both structural and semantic information across diverse modalities.

\textbf{(O2):} Facilitate cross-modal alignment, retrieval, and reasoning through natural language interaction.

\textbf{(O3):} Combine LLM-driven semantic operations with cross-modal processing capabilities.

\textbf{(O4):} Enable self-designing and adaptive evolution of analysis pipelines in response to changes in data sources.
\end{tcolorbox}

%% file: sections/industry.tex
\section{Industry Practice}
\label{sec:industry_practice}

The rapid evolution of LLM/Agent technologies has driven their integration into commercial BI and data science platforms as ``Copilot''-style assistants~\cite{chen2021evaluating}. This trend, fueled by the need to democratize data access and unlock insights from unstructured ``dark data''~\cite{bib:darkdata}, has created a vibrant ecosystem of specialized products for diverse data modalities.
Industrial applications, exemplified in Table~\ref{tab:industry_products}, differ notably across data modalities.

\begin{table}[!t]
\centering
\caption{Representative Industry Products for LLM/Agent-driven Data Analysis.}
\label{tab:industry_products}
\renewcommand{\arraystretch}{1.3}
\footnotesize
\begin{tabularx}{\columnwidth}{
  >{\raggedright\arraybackslash}m{0.2\columnwidth}
  >{\raggedright\arraybackslash}m{0.25\columnwidth}
  >{\raggedright\arraybackslash}X}
\toprule
\textbf{Data Type} & \textbf{Product} & \textbf{Core Functionality} \\
\midrule

\multirow{2}{*}{\makecell[l]{\textbf{Structured} \\ \textbf{Data}}}
& Dataiku~\cite{georgiev2022dataiku} & Unified AI platform; NL-to-Code; Automated ML pipelines. \\
\cmidrule(l){2-3}
& Cognite~\cite{bib:cognite_idc} & Industrial knowledge graph; NL query for assets \& time-series. \\
\midrule

\multirow{2}{*}{\makecell[l]{\textbf{Semi-} \\ \textbf{Structured} \\ \textbf{Data}}}
& Tableau~\cite{kvam2023chat2vis} & Conversational BI; NL-driven interactive visualization. \\
\cmidrule(l){2-3}
& Parseu~\cite{bib:parseur} & Agentic data extraction; Automated parsing for emails \& docs. \\
\midrule

\multirow{2}{*}{\makecell[l]{\textbf{Un-} \\ \textbf{Structured} \\ \textbf{Data}}}
& docAnalyzer.ai~\cite{bib:docanalyzer} & Document intelligence; Layout-aware RAG models. \\
\cmidrule(l){2-3}
& viAct~\cite{zhang2024towards} & Video analytics; Temporal reasoning for safety monitoring. \\
\midrule

\multirow{2}{*}{\makecell[l]{\textbf{Hetero-} \\ \textbf{geneous} \\ \textbf{Data}}}
& Dremio~\cite{bib:dremio_text2sql} & Lakehouse query engine; Agentic cross-source query planning. \\
\cmidrule(l){2-3}
& GE Predix~\cite{valle2023ge} & Industrial IoT analytics; Cross-modal alignment of sensor data. \\
\bottomrule
\end{tabularx}
\end{table}

For structured data, the NL-based interface is dominant. Platforms like Dataiku~\cite{georgiev2022dataiku} embed LLMs into autonomous pipelines, where natural language orchestrates entire workflows from data prep to model deployment. In vertical domains, Cognite~\cite{bib:cognite_idc} leverages LLMs for semantic analysis over industrial knowledge graphs, enabling complex queries on assets and time-series data. The key ``last-mile problem'' remains ensuring 100\% accuracy for critical queries.

In the semi-structured domain, agents tackle conversational analysis and automation. BI platforms such as Tableau~\cite{kvam2023chat2vis} facilitate conversational interaction for iterative visualization and insight discovery. Concurrently, extraction agents like Parseur~\cite{bib:parseur} automate semi-structured to structured transformations~\cite{abiteboul1997querying}, replacing brittle parsers with robust AI models. Their primary bottleneck is robustness against schematic or layout drifts.

For unstructured data, Retrieval-Augmented Generation~\cite{lewis2020retrieval} is the cornerstone. Advanced tools like docAnalyzer.ai~\cite{bib:docanalyzer} enhance RAG with multimodal layout-aware models to synergistically interpret text, tables, and images. In video analytics, platforms like viAct~\cite{zhang2024towards} apply temporal reasoning atop vision models, allowing agents to analyze event sequences for safety compliance. The industrial hurdle is scaling RAG while managing retrieval quality and permissions.

Finally, analyzing heterogeneous data remains a frontier. Lakehouse platforms like Dremio~\cite{bib:dremio_text2sql} employ LLMs as agentic query planners to orchestrate execution across disparate sources. Industrial IoT platforms such as GE Predix~\cite{valle2023ge} focus on deep cross-modal alignment, linking sensor data with textual logs for predictive analytics. The core challenge is enabling true cross-modal reasoning, a capability still in its infancy.

Placing these industry practices within the technical landscape of Figure~\ref{fig:content-overview} reveals the current state of maturity. While applications now span all data modalities and increasingly leverage semantic design, most commercial tools still operate within the ``tool-assisted'' paradigm.  \textit{Development autonomy}, where an agent can independently orchestrate novel analytical pipelines, remains a frontier. This gap highlights the significant hurdles in reliability and planning, setting the stage for future research.

%% file: sections/future-work.tex

\section{Challenges and Future Directions}
\label{sec:future-work}

\subsection{Structured Data}

\hi{Complex Analysis.} 
Structured data encapsulates both semantic information (e.g., cell contents in tables) and structural information (e.g., key–value relationships in tables or edge connections in graphs). Effective analysis requires integrating these two dimensions. For simple tasks, LLMs can often derive answers directly from explicit semantic cues or straightforward structural patterns. However, complex tasks demand deeper comprehension and multi-hop reasoning across semantics and structure, which remain challenging for current LLMs.
When LLMs alone fall short in handling such tasks end-to-end, multi-agent collaboration emerges as a promising direction. By decomposing a complex task into simpler subtasks, each agent can specialize in a narrower scope. Through prompt engineering or fine-tuning, LLMs can be tailored for specific subtasks, while intermediate outputs enhance both performance and interpretability.

\hi{Open World Task Adaptation.} 
Structured data spans diverse domains with varying specifications and task requirements. LLMs trained for general-purpose analysis may fail to meet domain-specific needs. For instance, analyzing tables or forms differs substantially from rigorous analyses in business or climate applications. Moreover, heterogeneity in system specifications, such as SQL dialects, poses additional adaptation challenges.
To address these issues, domain adaptation through task-specific training or fine-tuning is essential. As manual annotation is often insufficient to support large-scale training, efficient automatic data generation methods warrant further exploration. In addition, techniques such as automatic query conversion and intermediate representations are crucial for bridging modality gaps, particularly between natural language and diverse SQL dialects.

\subsection{Semi-Structured Data}

\hi{Diverse Markup Language Formats}.
Markup languages encompass a wide spectrum, including HTML, XML, JSON, Markdown, LaTeX, and domain-specific standards such as Journal Article Tag Suite (JATS). 
However, most existing research remains concentrated on a few widely used formats, primarily including HTML, XML, and JSON, which often derived from web-scraped data. 
This leaves domain-specific or less-studied markup languages, as well as proprietary formats in industry, underexplored.
Addressing this gap requires the creation of specialized benchmarks for languages such as JATS, LaTeX, and Markdown to evaluate tasks like structure-aware question answering and format conversion. 
Moreover, model architectures must evolve towards structure-aware tokenization and training paradigms that explicitly leverage the hierarchical organization of markup, rather than treating markup as plain text sequences.


\hi{Understanding Table with Complex Structures.}
Semi-structured tables often exhibit irregular relationships and intricate nested structures, which remain difficult for current LLMs to fully capture. 
While recent approaches attempt to encode table structures and leverage the semantic reasoning capacity of LLMs, these methods are usually tailored to specific table types (e.g., financial tables). 
It is quite challenging to handle diverse semi-structured tables with complex structures.
To address this, researchers have explored several directions. One line of work focuses on {Structure-Aware Encodings}, where the table’s hierarchical or graph-like organization (e.g., DOM trees for HTML tables or JSON schemas) is explicitly modeled and injected into LLMs. Another approach emphasizes {Intermediate Representations}, such as transforming semi-structured tables into canonical forms or domain-specific languages (DSLs) that better align with LLMs’ sequential reasoning ability. In addition, {Retrieval-Augmented Methods} leverage external knowledge bases or schema repositories to supplement missing semantics, enabling LLMs to ground their reasoning in richer contexts.

\hi{Large-Scale Table Comprehension.}
For both PLM- and LLM-based methods, efficiency and accuracy are constrained by token limitations and context window sizes. When processing large-scale tables, effectively compressing both structure and content while minimizing information loss remains an urgent challenge. 
Advances in hierarchical encoding, table summarization, and scalable retrieval-augmented mechanisms may provide promising directions for addressing this limitation.




\subsection{Unstructured Data}

\hi{High-Level Chart Understanding.}
While significant progress has been made in basic chart recognition and data extraction, critical higher-order tasks such as chart-based fact checking, detection of visual or statistical misrepresentations (e.g., truncated axes, inconsistent scales, or fabricated trends), and cross-modal chart data verification remain largely unaddressed. These tasks demand not only accurate perception of visual elements but also reasoning about data semantics, statistical validity, and alignment with external knowledge or textual assertions. The absence of standardized benchmarks and evaluation protocols for such tasks further hinders systematic progress.
We advocate for the development of comprehensive frameworks that support critical chart literacy—enabling systems to not just “read” charts, but to audit them. This includes creating datasets with annotated errors, factual inconsistencies, and deceptive visual practices, as well as designing models capable of joint visual-semantic reasoning to validate the integrity and truthfulness of the graph.

\hi{Inadequate Multimodal Reasoning for Diverse Chart Types.} Charts are inherently multimodal artifacts that fuse structured visual encodings (e.g., glyphs, spatial arrangements, color mappings) with natural language (titles, axis labels, captions). Current models often treat chart understanding as a vision-only or simplified vision-language problem, failing to capture the nuanced interplay between graphical conventions, domain-specific semantics, and contextual information. This limitation becomes especially pronounced when handling complex or non-standard chart types (e.g., small multiples, Sankey diagrams, or interactive dashboards).
A holistic, end-to-end chart understanding pipeline that unifies low-level perception (e.g., element detection), mid-level parsing (e.g., data-value mapping), and high-level inference (e.g., trend interpretation or anomaly detection) is needed. Such a pipeline should be modular, chart-type-agnostic, and grounded in both visualization theory and real-world usage patterns, enabling robust performance across diverse downstream applications.



\hi{Computational Efficiency of Video Analysis.} 
Video is long-form sequential data; fixed sampling or full-frame processing consumes large compute and wastes capacity. Compressing frames/tokens often sacrifices fine-grained temporal cues, hurting localization and detail recognition.
To address this, we can develop representations that preserve fine-grained visual detail while offering controllable context length (e.g., temporal separator tokens, dynamic sampling, learnable token merging, temporal positional encodings). Integrate temporal modeling and token-aggregation objectives into training to achieve a better trade-off between efficiency and fidelity.

\hi{Video Temporal Localization with Semantic Coherence.} 
Precise temporal localization requires detecting sparse, transient cues (e.g., action start/end, sudden events) and maintaining semantic consistency across long sequences. Current methods still struggle with event-boundary detection, timestamp reasoning, and preventing erroneous cross-time associations.
To address this,we can improve fine-grained event detection and semantic continuity by incorporating temporally aware tokens and supervision objectives directly into LLM-friendly representations.

\hi{Video Multimodal Fusion and Scalability.} 
Effectively time-aligning and fusing visual, audio, and textual streams is both modeling and engineering challenging; high-quality, fine-grained human annotation is costly and hard to scale. Domain-specific tasks (emotion, gesture, anomaly detection, etc.) often rely on specialized labels and bespoke pipelines, which limits generalization.
To address this,we can combine modular architectures (lightweight temporal adapters, instruction-aware selectors), multi-agent data curation/reviewer pipelines, and large-scale synthetic video–text generation to improve annotation efficiency, domain adaptation, and instruction-conditioned reasoning. Encourage task-centered cooperative agents that decompose complex video-understanding problems into simpler subtasks.


\hi{Generalizing Multimodal Understanding for Documents.}
A key challenge beyond OCR is achieving deep multimodal understanding—integrating text, layout, and visuals—across diverse domains and languages. General-purpose models often fail on specialized document structures, such as dense scientific layouts or non-linear historical manuscripts, and struggle with language-specific conventions like right-to-left scripts. Addressing this requires document-native architectures (e.g., DocLLM) that emphasize layout over costly visual encoding, combined with layout-aware instruction tuning to teach multi-granularity structural reasoning, supported by large-scale, multilingual datasets for robust generalization.


\hi{Efficiency in Large-Scale Document Processing.}
Processing large volumes of long-form documents, such as legal archives, presents a trade-off between computational efficiency and analytical depth. The limited context windows of most LLMs force chunking strategies that can sever long-range dependencies, while preserving fine-grained spatial details across thousands of pages remains a major technical hurdle.
Future work should focus on adaptive context management and hierarchical document representations. Combining these with progressive reasoning strategies, where a model first builds a coarse understanding before focusing on details, can create a better balance between computational cost and high-fidelity analytical accuracy.

\hi{High Quality 3D Modality Representation.} 
3D data, including point clouds, meshes, voxels, implicit fields, and NeRFs, vary in sampling, topology, and attributes. Converting them into LLM-ready inputs requires trade-offs, where aggressive simplification loses geometric detail, while token explosion exceeds context limits, leading to inconsistent cues and poor generalization. This can be mitigated with compact, trainable encoders that produce a small set of informative tokens, such as vector-quantized mesh tokens, learned point-feature aggregators, and geometry-aware positional encodings. Jointly training these encoders with lightweight adapters using multiview consistency, reprojection, and geometric-accuracy losses preserves topology and metric cues while staying within LLM context budgets.

\hi{Accurate 3D Geometric Grounding.} 
LLMs lack native metric reasoning and struggle with coordinate frames, units, and precise spatial numerics. The scarcity of richly labeled 3D datasets limits supervised learning for distances, poses, and geometric constraints, so systems often produce plausible qualitative descriptions but fail on quantitative queries, affecting navigation, manipulation, simulation, and safety-critical tasks. This can be addressed using domain adapters that encode coordinate conventions, units, and regression heads for distances/poses, trained with metric supervision from simulators/renderers or differentiable-rendering losses, and evaluated on numerically grounded benchmarks to ensure quantitative fidelity.


\subsection{Heterogeneous Data.}

\hi{Pluggable Modality Extensibility.}
A key challenge for heterogeneous data agents is integrating new data types without redesigning or retraining the system. Traditional architectures are rigid, making it difficult to handle emerging modalities like audio, video, or sensor data. Future work should focus on modular and extensible frameworks with independent encoders or adapters, standardized integration interfaces, adaptive weighting for multiple modalities, and self-supervised alignment to ensure compatibility with existing representations. Such approaches will enhance scalability, maintainability, and robustness in increasingly complex and evolving data.


\hi{High-Level Heterogeneous Data Analysis.}
A key challenge for heterogeneous data agents is performing high-level analysis, encompassing reasoning, summarization, and cross-modal inference beyond simple retrieval. This requires understanding interrelationships among modalities, capturing context, and generating actionable insights. Existing systems, such as XMODE~\cite{nooralahzadeh2024explainable} and~\cite{wang2024interactive}, offer partial solutions via step-wise query decomposition or adaptive modality weighting but still struggle with large-scale or disorganized datasets, limiting fully automated, context-aware analysis.
Future research should develop agents capable of robust, end-to-end reasoning across heterogeneous modalities. Promising directions include using multimodal embeddings for joint inference, integrating symbolic and neural reasoning for interpretability, and employing adaptive task planning to decompose complex queries and synthesize results across modalities for comprehensive real-world data analysis.



%% file: sections/conclusion.tex
\section{Conclusion}
\label{sec:conclusion}

In this paper, we summarize the recent techniques on LLM/Agent for data analysis, organized from the perspective of diverse data types, including structured data, semi-structured data, unstructured data, and heterogeneous data.
Five key design goals for intelligent data analysis agents are distilled from the technical evolution.
We also outline several remaining research challenges and propose some insights and practical directions for advancing LLM/Agents-powered data analysis.

%% file: main.bbl
\begin{thebibliography}{100}

\bibitem{blazegraph}
https://blazegraph.com/.

\bibitem{neo4j}
https://github.com/neo4j/neo4j.

\bibitem{graphdb}
https://graphdb.ontotext.com/.

\bibitem{pymol}
https://pymol.org/.

\bibitem{torchptpth}
https://pytorch.org/.

\bibitem{mongodb}
https://www.mongodb.com/.

\bibitem{iceberg}
Iceberg.

\bibitem{abiteboul1997querying}
S.~Abiteboul.
\newblock Querying semi-structured data.
\newblock In {\em Database Theory—ICDT'97: 6th International Conference Delphi, Greece, January 8--10, 1997 Proceedings 6}, pages 1--18. Springer, 1997.

\bibitem{abramovich2024visfocus}
O.~Abramovich, N.~Nayman, A.~Levi, R.~Magar, and J.~Goldberger.
\newblock {VisFocus: Prompt-Guided Vision Encoders for OCR-Free Dense Document Understanding}.
\newblock {\em arXiv preprint arXiv:2407.12594}, 2024.

\bibitem{adepu2020interviewee}
Y.~Adepu, V.~R. Boga, et~al.
\newblock Interviewee performance analyzer using facial emotion recognition and speech fluency recognition.
\newblock In {\em 2020 IEEE International Conference for Innovation in Technology (INOCON)}, pages 1--5. IEEE, 2020.

\bibitem{al2024chartsurvey}
M.~Al-Shetairy, H.~Hindy, D.~Khattab, and M.~M. Aref.
\newblock Transformers utilization in chart understanding: A review of recent advances \& future trends.
\newblock {\em arXiv preprint arXiv:2410.13883}, 2024.

\bibitem{temporal2022}
M.~M. Alam, L.~Torgo, and A.~Bifet.
\newblock A survey on spatio-temporal data analytics systems.
\newblock {\em ACM Comput. Surv.}, 54(10s), Nov. 2022.

\bibitem{alnegheimish2024can}
S.~Alnegheimish, L.~Nguyen, L.~Berti-Equille, and K.~Veeramachaneni.
\newblock Can large language models be anomaly detectors for time series?
\newblock In {\em 2024 IEEE 11th International Conference on Data Science and Advanced Analytics (DSAA)}, pages 1--10. IEEE, 2024.

\bibitem{defgraphdata}
R.~Angles.
\newblock A comparison of current graph database models.
\newblock In {\em 2012 IEEE 28th International Conference on Data Engineering Workshops}, pages 171--177, 2012.

\bibitem{graphdata}
R.~Angles and C.~Gutierrez.
\newblock Survey of graph database models.
\newblock {\em ACM Comput. Surv.}, 40(1), Feb. 2008.

\bibitem{appalaraju2021docformer}
S.~Appalaraju, B.~Jasani, B.~U. Kota, Y.~Xie, and R.~Manmatha.
\newblock Docformer: End-to-end transformer for document understanding.
\newblock In {\em Proceedings of the IEEE/CVF International Conference on Computer Vision (ICCV)}, pages 993--1003, October 2021.

\bibitem{arora2023language}
S.~Arora, B.~Yang, S.~Eyuboglu, A.~Narayan, A.~Hojel, I.~Trummer, and C.~R{\'e}.
\newblock Language models enable simple systems for generating structured views of heterogeneous data lakes.
\newblock {\em arXiv preprint arXiv:2304.09433}, 2023.

\bibitem{bai2025collaboration}
X.~Bai, S.~Huang, C.~Wei, and R.~Wang.
\newblock Collaboration between intelligent agents and large language models: A novel approach for enhancing code generation capability.
\newblock {\em Expert Systems with Applications}, 269:126357, 2025.

\bibitem{bakkali2023vlcdoc}
S.~Bakkali, Z.~Ming, M.~Coustaty, M.~Rusinol, A.~Fornes, and J.~Llados.
\newblock Vlcdoc: Vision-language contrastive pre-training model for cross-modal document classification.
\newblock {\em Pattern Recognition}, 137:109296, 2023.

\bibitem{bib:dremio_text2sql}
V.~Balamurali.
\newblock {Making Conversational BI on the Lakehouse a Reality with Dremio Text-to-SQL}.
\newblock Dremio Blog, August 2023.
\newblock Accessed on: 2024-10-25.

\bibitem{barboule2025docsurvey}
C.~Barboule, B.~Piwowarski, and Y.~Chabot.
\newblock Survey on question answering over visually rich documents: Methods, challenges, and trends.
\newblock {\em arXiv preprint arXiv:2501.02235}, 2025.

\bibitem{batko2022use}
K.~Batko and A.~{\'S}l{\k{e}}zak.
\newblock The use of big data analytics in healthcare.
\newblock {\em Journal of big Data}, 9(1):3, 2022.

\bibitem{baviskar2021efficient}
D.~Baviskar, S.~Ahirrao, V.~Potdar, and K.~Kotecha.
\newblock Efficient automated processing of the unstructured documents using artificial intelligence: A systematic literature review and future directions.
\newblock {\em Ieee Access}, 9:72894--72936, 2021.

\bibitem{blain2019complete}
J.~M. Blain.
\newblock {\em The complete guide to Blender graphics: computer modeling \& animation}.
\newblock AK Peters/CRC Press, 2019.

\bibitem{blattmann2023align}
A.~Blattmann, R.~Rombach, H.~Ling, T.~Dockhorn, S.~W. Kim, S.~Fidler, and K.~Kreis.
\newblock Align your latents: High-resolution video synthesis with latent diffusion models.
\newblock In {\em Proceedings of the IEEE/CVF conference on computer vision and pattern recognition}, pages 22563--22575, 2023.

\bibitem{cao2022ai}
L.~Cao.
\newblock Ai in finance: challenges, techniques, and opportunities.
\newblock {\em ACM Computing Surveys (CSUR)}, 55(3):1--38, 2022.

\bibitem{chang2025llm4ts}
C.~Chang, W.-Y. Wang, W.-C. Peng, and T.-F. Chen.
\newblock Llm4ts: Aligning pre-trained llms as data-efficient time-series forecasters.
\newblock {\em ACM Transactions on Intelligent Systems and Technology}, 16(3):1--20, 2025.

\bibitem{SV-RAG}
J.~Chen, R.~Zhang, Y.~Zhou, J.~Gu, J.~Kuen, C.~Chen, and T.~Sun.
\newblock Sv-rag: Lora-contextualizing adaptation of mllms for long document understanding.
\newblock {\em arXiv preprint arXiv:2411.01106}, 2024.

\bibitem{chen2025breakingsftplateaumultimodal}
L.~Chen, X.~Zhao, Z.~Zeng, J.~Huang, L.~Zheng, Y.~Zhong, and L.~Ma.
\newblock Breaking the sft plateau: Multimodal structured reinforcement learning for chart-to-code generation, 2025.

\bibitem{chen2021evaluating}
M.~Chen, J.~Tworek, H.~Jun, Q.~Yuan, H.~P. de~Oliveira~Pinto, J.~Kaplan, H.~Edwards, Y.~Burda, N.~Joseph, G.~Brockman, A.~Ray, R.~Puri, G.~Krueger, M.~Petrov, H.~Khlaaf, G.~Sastry, P.~Mishkin, B.~Chan, S.~Gray, N.~Ryder, M.~Pavlov, A.~Power, L.~Kaiser, M.~Bavarian, C.~Winter, P.~Tillet, F.~P. Such, D.~Cummings, M.~Plappert, F.~Chantzis, E.~Barnes, A.~Herbert-Voss, W.~H. Guss, A.~Nichol, A.~Paino, N.~Tezak, J.~Tang, I.~Babuschkin, S.~Balaji, S.~Jain, W.~Saunders, C.~Hesse, A.~N. Carr, J.~Leike, J.~Achiam, V.~Misra, E.~Hubinger, L.~Weng, V.~Kosaraju, M.~Murati, D.~Amodei, D.~Hendrikx, J.~Schulman, I.~Sutskever, and W.~Zaremba.
\newblock Evaluating large language models trained on code, 2021.

\bibitem{chen2023fantasia3d}
R.~Chen, Y.~Chen, N.~Jiao, and K.~Jia.
\newblock Fantasia3d: Disentangling geometry and appearance for high-quality text-to-3d content creation.
\newblock In {\em Proceedings of the IEEE/CVF international conference on computer vision}, pages 22246--22256, 2023.

\bibitem{chen2024timemarker}
S.~Chen, X.~Lan, Y.~Yuan, Z.~Jie, and L.~Ma.
\newblock Timemarker: A versatile video-llm for long and short video understanding with superior temporal localization ability.
\newblock {\em arXiv preprint arXiv:2411.18211}, 2024.

\bibitem{chen2024meshanything}
Y.~Chen, T.~He, D.~Huang, W.~Ye, S.~Chen, J.~Tang, X.~Chen, Z.~Cai, L.~Yang, G.~Yu, et~al.
\newblock Meshanything: Artist-created mesh generation with autoregressive transformers.
\newblock {\em arXiv preprint arXiv:2406.10163}, 2024.

\bibitem{chen2025large}
Y.~Chen, Z.~Li, C.~Yang, X.~Wang, and G.~Xu.
\newblock Large language models are few-shot multivariate time series classifiers.
\newblock {\em arXiv preprint arXiv:2502.00059}, 2025.

\bibitem{chen1992object}
Y.~Chen and G.~Medioni.
\newblock Object modelling by registration of multiple range images.
\newblock {\em Image and vision computing}, 10(3):145--155, 1992.

\bibitem{chen2023symphony}
Z.~Chen, Z.~Gu, L.~Cao, J.~Fan, S.~Madden, and N.~Tang.
\newblock Symphony: Towards natural language query answering over multi-modal data lakes.
\newblock In {\em CIDR}, pages 1--7, 2023.

\bibitem{cheng2024necessary}
S.~Cheng, Z.~Zhuang, Y.~Xu, F.~Yang, C.~Zhang, X.~Qin, X.~Huang, L.~Chen, Q.~Lin, D.~Zhang, S.~Rajmohan, and Q.~Zhang.
\newblock Call me when necessary: Llms can efficiently and faithfully reason over structured environments, 2024.

\bibitem{M3DOCRAG}
J.~Cho, D.~Mahata, O.~İrsoy, Y.~He, and M.~Bansal.
\newblock M3docrag: Multi-modal retrieval is what you need for multi-page multi-document understanding.
\newblock {\em arXiv preprint arXiv:2411.04952}, 2024.

\bibitem{chowdhery2022palm}
A.~Chowdhery, S.~Narang, J.~Devlin, M.~Bosma, G.~Mishra, A.~Roberts, P.~Barham, H.~W. Chung, C.~Sutton, S.~Gehrmann, P.~Schuh, K.~Shi, S.~Tsvyashchenko, J.~Maynez, A.~Rao, P.~Barnes, Y.~Tay, N.~Shazeer, V.~Prabhakaran, E.~Reif, N.~Du, B.~Hutchinson, R.~Pope, J.~Bradbury, J.~Austin, M.~Isard, G.~Gur-Ari, P.~Yin, T.~Duke, A.~Levskaya, S.~Ghemawat, S.~Dev, H.~Michalewski, X.~Garcia, V.~Misra, K.~Robinson, L.~Fedus, D.~Zhou, D.~Ippolito, D.~Luan, H.~Lim, B.~Zoph, A.~Spiridonov, R.~Sepassi, D.~Dohan, S.~Agrawal, M.~Omernick, A.~M. Dai, T.~S. Pillai, M.~Pellat, A.~Lewkowycz, E.~Moreira, R.~Child, O.~Polozov, K.~Lee, Z.~Zhou, X.~Wang, B.~Saeta, M.~Diaz, O.~Firat, M.~Catasta, J.~Wei, K.~Meier-Hellstern, D.~Eck, J.~Dean, S.~Petrov, and N.~Fiedel.
\newblock Palm: Scaling language modeling with pathways, 2022.

\bibitem{relationaldata}
E.~F. Codd.
\newblock A relational model of data for large shared data banks.
\newblock {\em Commun. ACM}, 13(6):377–387, June 1970.

\bibitem{crescenzi2001automatic}
V.~Crescenzi, G.~Mecca, and P.~Merialdo.
\newblock Automatic web information extraction in the roadrunner system.
\newblock In {\em International Conference on Conceptual Modeling}, pages 264--277. Springer, 2001.

\bibitem{dai2024uqequery}
H.~Dai, B.~Y. Wang, X.~Wan, B.~Dai, S.~Yang, A.~Nova, P.~Yin, P.~M. Phothilimthana, C.~Sutton, and D.~Schuurmans.
\newblock Uqe: A query engine for unstructured databases, 2024.

\bibitem{10943225}
S.~Darabi, P.~Bigaj, D.~Majchrowski, A.~Kasymov, P.~Morkisz, and A.~Fit-Florea.
\newblock A framework for large-scale synthetic graph dataset generation.
\newblock {\em IEEE Transactions on Neural Networks and Learning Systems}, pages 1--11, 2025.

\bibitem{das2025chartsofthoughtenhancingllmvisualization}
A.~K. Das, M.~Tarun, and K.~Mueller.
\newblock Charts-of-thought: Enhancing llm visualization literacy through structured data extraction, 2025.

\bibitem{bigdataanalysis}
T.~K. Das and P.~M. Kumar.
\newblock Big data analytics: A framework for unstructured data analysis.
\newblock {\em International Journal of Engineering Science \& Technology}, 5(1):153, 2013.

\bibitem{de2024video}
U.~De~Silva, L.~Fernando, K.~Bandara, and R.~Nawaratne.
\newblock Video summarisation with incident and context information using generative ai.
\newblock In {\em IECON 2024-50th Annual Conference of the IEEE Industrial Electronics Society}, pages 1--6. IEEE, 2024.

\bibitem{deng2025seq2time}
A.~Deng, Z.~Gao, A.~Choudhuri, B.~Planche, M.~Zheng, B.~Wang, T.~Chen, C.~Chen, and Z.~Wu.
\newblock Seq2time: Sequential knowledge transfer for video llm temporal grounding.
\newblock In {\em Proceedings of the Computer Vision and Pattern Recognition Conference}, pages 13766--13775, 2025.

\bibitem{deng2022dom}
X.~Deng, P.~Shiralkar, C.~Lockard, B.~Huang, and H.~Sun.
\newblock Dom-lm: Learning generalizable representations for html documents.
\newblock {\em arXiv preprint arXiv:2201.10608}, 2022.

\bibitem{dernbach2024glam}
S.~Dernbach, K.~Agarwal, A.~Zuniga, M.~Henry, and S.~Choudhury.
\newblock Glam: Fine-tuning large language models for domain knowledge graph alignment via neighborhood partitioning and generative subgraph encoding.
\newblock In {\em Proceedings of the AAAI Symposium Series}, volume~3, pages 82--89, 2024.

\bibitem{devlin2019bert}
J.~Devlin, M.-W. Chang, K.~Lee, and K.~Toutanova.
\newblock Bert: Pre-training of deep bidirectional transformers for language understanding.
\newblock In {\em Proceedings of the 2019 conference of the North American chapter of the association for computational linguistics: human language technologies, volume 1 (long and short papers)}, pages 4171--4186, 2019.

\bibitem{ding2024docsurvey}
Y.~Ding, S.~C. Han, J.~Lee, and E.~Hovy.
\newblock Deep learning based visually rich document content understanding: A survey.
\newblock {\em arXiv preprint arXiv:2408.01287}, 2024.

\bibitem{bib:docanalyzer}
{docAnalyzer.ai}.
\newblock {docAnalyzer.ai: Chat with your documents using AI}.
\newblock \url{https://docanalyzer.ai/}, 2024.
\newblock Accessed on: 2024-10-25.

\bibitem{10.1145/3726302.3730071}
H.~Dong, Y.~Hu, and Y.~Cao.
\newblock Reasoning and retrieval for complex semi-structured tables via reinforced relational data transformation.
\newblock In {\em Proceedings of the 48th International ACM SIGIR Conference on Research and Development in Information Retrieval}, SIGIR '25, page 1382–1391, New York, NY, USA, 2025. Association for Computing Machinery.

\bibitem{dong2024spreadsheetllm}
H.~Dong, J.~Zhao, Y.~Tian, J.~Xiong, S.~Xia, M.~Zhou, Y.~Lin, J.~Cambronero, Y.~He, S.~Han, et~al.
\newblock Spreadsheetllm: encoding spreadsheets for large language models.
\newblock {\em arXiv preprint arXiv:2407.09025}, 2024.

\bibitem{fan2024combining}
J.~Fan, Z.~Gu, S.~Zhang, Y.~Zhang, Z.~Chen, L.~Cao, G.~Li, S.~Madden, X.~Du, and N.~Tang.
\newblock Combining small language models and large language models for zero-shot nl2sql.
\newblock {\em Proceedings of the VLDB Endowment}, 17(11):2750--2763, 2024.

\bibitem{fan2024unicorn}
J.~Fan, J.~Tu, G.~Li, P.~Wang, X.~Du, X.~Jia, S.~Gao, and N.~Tang.
\newblock Unicorn: a unified multi-tasking matching model.
\newblock {\em ACM SIGMOD Record}, 53(1):44--53, 2024.

\bibitem{farahani2023chartsurvey}
A.~M. Farahani, P.~Adibi, M.~S. Ehsani, H.-P. Hutter, and A.~Darvishy.
\newblock Automatic chart understanding: a review.
\newblock {\em IEEE Access}, 11:76202--76221, 2023.

\bibitem{faysse2025colpali}
M.~Faysse, H.~Sibille, T.~Wu, B.~Omrani, G.~Viaud, C.~Hudelot, and P.~Colombo.
\newblock Colpali: Efficient document retrieval with vision language models, 2025.

\bibitem{garland1997surface}
M.~Garland and P.~S. Heckbert.
\newblock Surface simplification using quadric error metrics.
\newblock In {\em Proceedings of the 24th annual conference on Computer graphics and interactive techniques}, pages 209--216, 1997.

\bibitem{bib:darkdata}
{Gartner}.
\newblock {Gartner Glossary: Dark Data}.
\newblock \url{https://www.gartner.com/en/information-technology/glossary/dark-data}, 2024.
\newblock The industry-standard definition. Accessed on: 2024-10-25.

\bibitem{georgiev2022dataiku}
G.~Georgiev, K.~Maman, and N.~L’Hévéder.
\newblock Dataiku: The all-in-one platform for enterprise ai.
\newblock {\em Proceedings of the 28th ACM SIGKDD Conference on Knowledge Discovery and Data Mining}, pages 4845--4846, 2022.

\bibitem{guo2023proteinchat}
H.~Guo, M.~Huo, R.~Zhang, and P.~Xie.
\newblock Proteinchat: Towards achieving chatgpt-like functionalities on protein 3d structures.
\newblock {\em Authorea Preprints}, 2023.

\bibitem{guo2024xs}
J.~Guo, Z.~Xu, L.~Wu, F.~Gao, W.~Liu, and X.~Wang.
\newblock Xs-vid: An extremely small video object detection dataset.
\newblock {\em arXiv preprint arXiv:2407.18137}, 2024.

\bibitem{guo2022trust}
Z.~Guo, Y.~Yu, P.~Lv, C.~Zhang, H.~Li, Z.~Wang, K.~Yao, J.~Liu, and J.~Wang.
\newblock Trust: An accurate and end-to-end table structure recognizer using splitting-based transformers.
\newblock {\em arXiv preprint arXiv:2208.14687}, 2022.

\bibitem{gupta2023temptabqa}
V.~Gupta, P.~Kandoi, M.~B. Vora, S.~Zhang, Y.~He, R.~Reinanda, and V.~Srikumar.
\newblock Temptabqa: Temporal question answering for semi-structured tables, 2023.

\bibitem{gupta2020infotabs}
V.~Gupta, M.~Mehta, P.~Nokhiz, and V.~Srikumar.
\newblock Infotabs: Inference on tables as semi-structured data.
\newblock {\em arXiv preprint arXiv:2005.06117}, 2020.

\bibitem{ha2022information}
H.~T. Ha and A.~Hor{\'a}k.
\newblock Information extraction from scanned invoice images using text analysis and layout features.
\newblock {\em Signal Processing: Image Communication}, 102:116601, 2022.

\bibitem{10.5555/3294771.3294869}
W.~L. Hamilton, R.~Ying, and J.~Leskovec.
\newblock Inductive representation learning on large graphs.
\newblock In {\em Proceedings of the 31st International Conference on Neural Information Processing Systems}, NIPS'17, page 1025–1035, Red Hook, NY, USA, 2017. Curran Associates Inc.

\bibitem{han2023chartllama}
Y.~Han, C.~Zhang, X.~Chen, X.~Yang, Z.~Wang, G.~Yu, B.~Fu, and H.~Zhang.
\newblock Chartllama: A multimodal llm for chart understanding and generation.
\newblock {\em arXiv preprint arXiv:2311.16483}, 2023.

\bibitem{hardy2004handbook}
M.~A. Hardy and A.~Bryman.
\newblock Handbook of data analysis.
\newblock 2004.

\bibitem{he2020deberta}
P.~He, X.~Liu, J.~Gao, and W.~Chen.
\newblock Deberta: Decoding-enhanced bert with disentangled attention.
\newblock {\em arXiv preprint arXiv:2006.03654}, 2020.

\bibitem{he2024g}
X.~He, Y.~Tian, Y.~Sun, N.~Chawla, T.~Laurent, Y.~LeCun, X.~Bresson, and B.~Hooi.
\newblock G-retriever: Retrieval-augmented generation for textual graph understanding and question answering.
\newblock {\em Advances in Neural Information Processing Systems}, 37:132876--132907, 2024.

\bibitem{herzig2020tapas}
J.~Herzig, P.~K. Nowak, T.~M{\"u}ller, F.~Piccinno, and J.~M. Eisenschlos.
\newblock Tapas: Weakly supervised table parsing via pre-training.
\newblock {\em arXiv preprint arXiv:2004.02349}, 2020.

\bibitem{ho2022imagen}
J.~Ho, W.~Chan, C.~Saharia, J.~Whang, R.~Gao, A.~Gritsenko, D.~P. Kingma, B.~Poole, M.~Norouzi, D.~J. Fleet, et~al.
\newblock Imagen video: High definition video generation with diffusion models.
\newblock {\em arXiv preprint arXiv:2210.02303}, 2022.

\bibitem{hoff2015multilinear}
P.~D. Hoff.
\newblock Multilinear tensor regression for longitudinal relational data.
\newblock {\em The annals of applied statistics}, 9(3):1169, 2015.

\bibitem{hong2024datainterpreter}
S.~Hong, Y.~Lin, B.~Liu, B.~Liu, B.~Wu, C.~Zhang, C.~Wei, D.~Li, J.~Chen, J.~Zhang, J.~Wang, L.~Zhang, L.~Zhang, M.~Yang, M.~Zhuge, T.~Guo, T.~Zhou, W.~Tao, X.~Tang, X.~Lu, X.~Zheng, X.~Liang, Y.~Fei, Y.~Cheng, Z.~Gou, Z.~Xu, and C.~Wu.
\newblock Data interpreter: An llm agent for data science, 2024.

\bibitem{hong20233d}
Y.~Hong, H.~Zhen, P.~Chen, S.~Zheng, Y.~Du, Z.~Chen, and C.~Gan.
\newblock 3d-llm: Injecting the 3d world into large language models.
\newblock {\em Advances in Neural Information Processing Systems}, 36:20482--20494, 2023.

\bibitem{hoppe1997view}
H.~Hoppe.
\newblock View-dependent refinement of progressive meshes.
\newblock In {\em Proceedings of the 24th annual conference on Computer graphics and interactive techniques}, pages 189--198, 1997.

\bibitem{hsieh2024rationale}
C.-Y. Hsieh, C.~Zhang, Y.~Jiang, Y.-S. Lee, W.-L. Cheng, B.~Li, J.~Wu, A.~Kembhavi, and A.~Farhadi.
\newblock {Rationale Distillation for Efficient VLM Training}.
\newblock {\em arXiv preprint arXiv:2406.03182}, 2024.

\bibitem{hu2024mplugdocowl15}
A.~Hu, H.~Xu, J.~Ye, M.~Yan, L.~Zhang, B.~Zhang, C.~Li, J.~Zhang, Q.~Jin, F.~Huang, and J.~Zhou.
\newblock mplug-docowl 1.5: Unified structure learning for ocr-free document understanding, 2024.

\bibitem{hu2024mplugdocowl2}
A.~Hu, H.~Xu, L.~Zhang, J.~Ye, M.~Yan, J.~Zhang, Q.~Jin, F.~Huang, and J.~Zhou.
\newblock mplug-docowl2: High-resolution compressing for ocr-free multi-page document understanding, 2024.

\bibitem{hu2023multimodal}
F.~Hu, Y.~Hu, W.~Zhang, H.~Huang, Y.~Pan, and P.~Yin.
\newblock A multimodal protein representation framework for quantifying transferability across biochemical downstream tasks.
\newblock {\em Advanced Science}, 10(22):2301223, 2023.

\bibitem{huang2024protchat}
H.~Huang, X.~Shi, H.~Lei, F.~Hu, and Y.~Cai.
\newblock Protchat: An ai multi-agent for automated protein analysis leveraging gpt-4 and protein language model.
\newblock {\em Journal of Chemical Information and Modeling}, 65(1):62--70, 2024.

\bibitem{huang2024datacoder}
J.~Huang, D.~Guo, C.~Wang, J.~Gu, S.~Lu, J.~P. Inala, C.~Yan, J.~Gao, N.~Duan, and M.~R. Lyu.
\newblock Contextualized data-wrangling code generation in computational notebooks.
\newblock In {\em Proceedings of the 39th IEEE/ACM International Conference on Automated Software Engineering}, ASE ’24, page 1282–1294. ACM, Oct. 2024.

\bibitem{huang2025evochart}
M.~Huang, H.~Lai, X.~Zhang, W.~Wu, J.~Ma, L.~Zhang, and J.~Liu.
\newblock Evochart: A benchmark and a self-training approach towards real-world chart understanding.
\newblock In {\em Proceedings of the AAAI Conference on Artificial Intelligence}, volume~39, pages 3680--3688, 2025.

\bibitem{huang2024image}
S.~Huang, H.~Zhang, L.~Zhong, H.~Chen, Y.~Gao, Y.~Hu, and Z.~Qin.
\newblock From image to video, what do we need in multimodal llms?
\newblock {\em arXiv preprint arXiv:2404.11865}, 2024.

\bibitem{huang2024mmp}
T.-Y. Huang, S.-Y. Chuang, L.-H. Sun, and M.-H. Chen.
\newblock Mmp: Towards robust multi-modal learning with masked modality projection.
\newblock {\em arXiv preprint arXiv:2410.03010}, 2024.

\bibitem{huang2022layoutlmv3}
Y.~Huang, T.~Lv, L.~Cui, Y.~Lu, and F.~Wei.
\newblock Layoutlmv3: Pre-training for document ai with unified text and image masking, 2022.

\bibitem{hudovernik2024relational}
V.~Hudovernik.
\newblock Relational data generation with graph neural networks and latent diffusion models.
\newblock In {\em NeurIPS 2024 Third Table Representation Learning Workshop}, 2024.

\bibitem{hui2023ldgm}
M.~Hui, Z.~Zhang, X.~Zhang, W.~Xie, Y.~Wang, and Y.~Lu.
\newblock Unifying layout generation with a decoupled diffusion model.
\newblock {\em arXiv preprint arXiv:2303.05049}, 2023.

\bibitem{bib:cognite_idc}
{IDC}.
\newblock {Cognite and Generative AI: The Next Wave of Industrial Data Management}.
\newblock IDC Analyst Connection, sponsored by Cognite, September 2023.
\newblock Accessed on: 2024-10-25.

\bibitem{inoue2023layoutdm}
N.~Inoue, K.~Kikuchi, M.~Otani, E.~Simo-Serra, and K.~Yamaguchi.
\newblock Layoutdm: Discrete diffusion model for controllable layout generation.
\newblock {\em arXiv preprint arXiv:2303.08137}, 2023.

\bibitem{jadhav2023ai}
A.~Jadhav, R.~Ghodake, K.~Muralidharan, and G.~T. Varma.
\newblock Ai based multimodal emotion and behavior analysis of interviewee.
\newblock 2023.

\bibitem{jiang2023structgpt}
J.~Jiang, K.~Zhou, Z.~Dong, K.~Ye, W.~X. Zhao, and J.-R. Wen.
\newblock Structgpt: A general framework for large language model to reason over structured data, 2023.

\bibitem{jiang2023unikgqa}
J.~Jiang, K.~Zhou, W.~X. Zhao, and J.-R. Wen.
\newblock Unikgqa: Unified retrieval and reasoning for solving multi-hop question answering over knowledge graph, 2023.

\bibitem{jiang2025explainable}
Y.~Jiang, W.~Yu, G.~Lee, D.~Song, K.~Shin, W.~Cheng, Y.~Liu, and H.~Chen.
\newblock Explainable multi-modal time series prediction with llm-in-the-loop.
\newblock {\em arXiv preprint arXiv:2503.01013}, 2025.

\bibitem{LLMQO}
R.~L. J. X. Y. C. P. H. C. H. M. D.~Z. Jie~Tan, Kangfei~Zhao and Y.~Rong.
\newblock Can large language models be query optimizer for relational databases?
\newblock {\em CoRR}, abs/2502.05562, 2025.

\bibitem{DBLP:conf/iclr/0005WMCZSCLLPW24}
M.~Jin, S.~Wang, L.~Ma, Z.~Chu, J.~Y. Zhang, X.~Shi, P.~Chen, Y.~Liang, Y.~Li, S.~Pan, and Q.~Wen.
\newblock Time-llm: Time series forecasting by reprogramming large language models.
\newblock In {\em The Twelfth International Conference on Learning Representations, {ICLR} 2024, Vienna, Austria, May 7-11, 2024}. OpenReview.net, 2024.

\bibitem{jin2022tablesurvey}
N.~Jin, J.~Siebert, D.~Li, and Q.~Chen.
\newblock A survey on table question answering: recent advances.
\newblock In {\em China Conference on Knowledge Graph and Semantic Computing}, pages 174--186. Springer, 2022.

\bibitem{kafle2018dvqaunderstandingdatavisualizations}
K.~Kafle, B.~Price, S.~Cohen, and C.~Kanan.
\newblock Dvqa: Understanding data visualizations via question answering, 2018.

\bibitem{khachatryan2023text2video}
L.~Khachatryan, A.~Movsisyan, V.~Tadevosyan, R.~Henschel, Z.~Wang, S.~Navasardyan, and H.~Shi.
\newblock Text2video-zero: Text-to-image diffusion models are zero-shot video generators.
\newblock In {\em Proceedings of the IEEE/CVF International Conference on Computer Vision}, pages 15954--15964, 2023.

\bibitem{kipf2017gcn}
T.~N. Kipf and M.~Welling.
\newblock Semi-supervised classification with graph convolutional networks.
\newblock In {\em Proceedings of the 5th International Conference on Learning Representations (ICLR)}, 2017.
\newblock Published as a conference paper at ICLR 2017.

\bibitem{klopries2024itf}
H.~Klopries and A.~Schwung.
\newblock Itf-gan: Synthetic time series dataset generation and manipulation by interpretable features.
\newblock {\em Knowledge-Based Systems}, 283:111131, 2024.

\bibitem{kvam2023chat2vis}
P.~Kvam, Z.~Lu, N.~De~Sverchkov, S.~L'Yi, D.~Wigdor, and J.~Zhao.
\newblock Chat2vis: Generating data visualizations via natural language using chatgpt, codex and gpt-3 large language models.
\newblock In {\em 2023 IEEE Visualization and Visual Analytics (VIS)}, pages 101--105. IEEE, 2023.

\bibitem{lee2024can}
C.~T. Lee.
\newblock {\em Can an LLM find its way around a Spreadsheet?}
\newblock PhD thesis, Virginia Tech, 2024.

\bibitem{lee2025timecap}
G.~Lee, W.~Yu, K.~Shin, W.~Cheng, and H.~Chen.
\newblock Timecap: Learning to contextualize, augment, and predict time series events with large language model agents.
\newblock In {\em Proceedings of the AAAI Conference on Artificial Intelligence}, volume~39, pages 18082--18090, 2025.

\bibitem{lee2024video}
S.-H. Lee, J.~Wang, Z.~Zhang, D.~Fan, and X.~Li.
\newblock Video token merging for long-form video understanding.
\newblock {\em arXiv preprint arXiv:2410.23782}, 2024.

\bibitem{lei2023s3hqa}
F.~Lei, X.~Li, Y.~Wei, S.~He, Y.~Huang, J.~Zhao, and K.~Liu.
\newblock {S}3{HQA}: A three-stage approach for multi-hop text-table hybrid question answering.
\newblock In A.~Rogers, J.~Boyd-Graber, and N.~Okazaki, editors, {\em Proceedings of the 61st Annual Meeting of the Association for Computational Linguistics (Volume 2: Short Papers)}, pages 1731--1740, Toronto, Canada, July 2023. Association for Computational Linguistics.

\bibitem{lei2024visdommultimodal}
M.~Lei, K.~Karthikeyan, W.~Liu, X.~Chen, B.~Li, Z.~Xu, Y.~K.~L. Wang, and J.~Gao.
\newblock {VisDoM: Multi-Document QA with Visually Rich Elements Using Multimodal Retrieval-Augmented Generation}.
\newblock {\em arXiv preprint arXiv:2412.10704}, 2024.

\bibitem{DBLP:conf/acl/LewisLGGMLSZ20}
M.~Lewis, Y.~Liu, N.~Goyal, M.~Ghazvininejad, A.~Mohamed, O.~Levy, V.~Stoyanov, and L.~Zettlemoyer.
\newblock {BART:} denoising sequence-to-sequence pre-training for natural language generation, translation, and comprehension.
\newblock In D.~Jurafsky, J.~Chai, N.~Schluter, and J.~R. Tetreault, editors, {\em Proceedings of the 58th Annual Meeting of the Association for Computational Linguistics, {ACL} 2020, Online, July 5-10, 2020}, pages 7871--7880. Association for Computational Linguistics, 2020.

\bibitem{lewis2020retrieval}
P.~Lewis, E.~Perez, A.~Piktus, F.~Petroni, V.~Karpukhin, N.~Goyal, H.~K{\"u}ttler, M.~Lewis, W.-t. Yih, T.~Rockt{\"a}schel, et~al.
\newblock Retrieval-augmented generation for knowledge-intensive nlp tasks.
\newblock {\em Advances in Neural Information Processing Systems}, 33:9459--9474, 2020.

\bibitem{SQL360}
B.~Li, Y.~Luo, C.~Chai, G.~Li, and N.~Tang.
\newblock The dawn of natural language to sql: Are we fully ready?
\newblock {\em Proceedings of the VLDB Endowment}, 17(11):3318–3331, July 2024.

\bibitem{li2025seed}
F.~Li, Y.~Wang, Y.~Liu, M.~Huang, D.~Hong, and J.~Ma.
\newblock Seed: A structural encoder for embedding-driven decoding in time series prediction with llms.
\newblock {\em arXiv preprint arXiv:2506.20167}, 2025.

\bibitem{LLM4DM}
G.~Li, X.~Zhou, and X.~Zhao.
\newblock Llm for data management.
\newblock {\em Proc. VLDB Endow.}, 17(12):4213–4216, Aug. 2024.

\bibitem{li2024codes}
H.~Li, J.~Zhang, H.~Liu, J.~Fan, X.~Zhang, J.~Zhu, R.~Wei, H.~Pan, C.~Li, and H.~Chen.
\newblock Codes: Towards building open-source language models for text-to-sql, 2024.

\bibitem{birdsql}
J.~Li, B.~Hui, G.~Qu, and et~al.
\newblock Can llm already serve as a database interface? a big bench for large-scale database grounded text-to-sqls.
\newblock {\em Advances in Neural Information Processing Systems}, 36, 2024.

\bibitem{li2021markuplm}
J.~Li, Y.~Xu, L.~Cui, and F.~Wei.
\newblock Markuplm: Pre-training of text and markup language for visually-rich document understanding.
\newblock {\em arXiv preprint arXiv:2110.08518}, 2021.

\bibitem{li2023autotablessynthesizingmultisteptransformations}
P.~Li, Y.~He, C.~Yan, Y.~Wang, and S.~Chaudhuri.
\newblock Auto-tables: Synthesizing multi-step transformations to relationalize tables without using examples, 2023.

\bibitem{li2023tablegpt}
P.~Li, Y.~He, D.~Yashar, W.~Cui, S.~Ge, H.~Zhang, D.~R. Fainman, D.~Zhang, and S.~Chaudhuri.
\newblock Table-gpt: Table-tuned gpt for diverse table tasks, 2023.

\bibitem{li2024towards}
S.~Li, Z.~Liu, Y.~Luo, X.~Wang, X.~He, K.~Kawaguchi, T.-S. Chua, and Q.~Tian.
\newblock Towards 3d molecule-text interpretation in language models.
\newblock {\em arXiv preprint arXiv:2401.13923}, 2024.

\bibitem{li2023sweetdreamer}
W.~Li, R.~Chen, X.~Chen, and P.~Tan.
\newblock Sweetdreamer: Aligning geometric priors in 2d diffusion for consistent text-to-3d.
\newblock {\em arXiv preprint arXiv:2310.02596}, 2023.

\bibitem{li2024craftsman3d}
W.~Li, J.~Liu, H.~Yan, R.~Chen, Y.~Liang, X.~Chen, P.~Tan, and X.~Long.
\newblock Craftsman3d: High-fidelity mesh generation with 3d native generation and interactive geometry refiner.
\newblock {\em arXiv preprint arXiv:2405.14979}, 2024.

\bibitem{li2024graph}
X.~Li, Z.~Wu, J.~Wu, H.~Cui, J.~Jia, R.-H. Li, and G.~Wang.
\newblock Graph learning in the era of llms: A survey from the perspective of data, models, and tasks, 2024.

\bibitem{li2024llms}
Y.~Li, X.~Chen, B.~Hu, and M.~Zhang.
\newblock Llms meet long video: Advancing long video comprehension with an interactive visual adapter in llms.
\newblock {\em arXiv preprint arXiv:2402.13546}, 3(7), 2024.

\bibitem{li2024xpath}
Y.~Li, B.~Wang, and X.~Luan.
\newblock Xpath agent: An efficient xpath programming agent based on llm for web crawler.
\newblock {\em arXiv preprint arXiv:2502.15688}, 2024.

\bibitem{li2024mimotable}
Z.~Li, Y.~Du, M.~Zheng, and M.~Song.
\newblock Mimotable: A multi-scale spreadsheet benchmark with meta operations for table reasoning, 2024.

\bibitem{li2024flexkbqa}
Z.~Li, S.~Fan, Y.~Gu, X.~Li, Z.~Duan, B.~Dong, N.~Liu, and J.~Wang.
\newblock Flexkbqa: A flexible llm-powered framework for few-shot knowledge base question answering, 2024.

\bibitem{petsql}
Z.~Li, X.~Wang, J.~Zhao, S.~Yang, G.~Du, X.~Hu, B.~Zhang, Y.~Ye, Z.~Li, R.~Zhao, and H.~Mao.
\newblock Pet-sql: A prompt-enhanced two-round refinement of text-to-sql with cross-consistency, June 2024.

\bibitem{liang2024aligning}
Y.~Liang, K.~Tan, T.~Xie, W.~Tao, S.~Wang, Y.~Lan, and W.~Qian.
\newblock Aligning large language models to a domain-specific graph database for nl2gql.
\newblock In {\em Proceedings of the 33rd ACM International Conference on Information and Knowledge Management}, pages 1367--1377, 2024.

\bibitem{liang2024natnl2gql}
Y.~Liang, T.~Xie, G.~Peng, Z.~Huang, Y.~Lan, and W.~Qian.
\newblock Nat-nl2gql: A novel multi-agent framework for translating natural language to graph query language, 2024.

\bibitem{lin2024zendb}
Y.~Lin, M.~Hulsebos, R.~Ma, S.~Shankar, S.~Zeigham, A.~G. Parameswaran, and E.~Wu.
\newblock Towards accurate and efficient document analytics with large language models, 2024.

\bibitem{liu2025timecma}
C.~Liu, Q.~Xu, H.~Miao, S.~Yang, L.~Zhang, C.~Long, Z.~Li, and R.~Zhao.
\newblock Timecma: Towards llm-empowered multivariate time series forecasting via cross-modality alignment.
\newblock In {\em Proceedings of the AAAI Conference on Artificial Intelligence}, volume~39, pages 18780--18788, 2025.

\bibitem{liu2025towards}
C.~Liu, S.~Zhou, Q.~Xu, H.~Miao, C.~Long, Z.~Li, and R.~Zhao.
\newblock Towards cross-modality modeling for time series analytics: A survey in the llm era.
\newblock {\em arXiv preprint arXiv:2505.02583}, 2025.

\bibitem{liu2024llava}
H.~Liu, C.~Li, Y.~Li, and Y.~J. Lee.
\newblock Improved baselines with visual instruction tuning, 2024.

\bibitem{liu2024chartthinker}
M.~Liu, D.~Chen, Y.~Li, G.~Fang, and Y.~Shen.
\newblock Chartthinker: A contextual chain-of-thought approach to optimized chart summarization.
\newblock {\em arXiv preprint arXiv:2403.11236}, 2024.

\bibitem{liu2025calf}
P.~Liu, H.~Guo, T.~Dai, N.~Li, J.~Bao, X.~Ren, Y.~Jiang, and S.-T. Xia.
\newblock Calf: Aligning llms for time series forecasting via cross-modal fine-tuning.
\newblock In {\em Proceedings of the AAAI Conference on Artificial Intelligence}, volume~39, pages 18915--18923, 2025.

\bibitem{liu2023zero}
R.~Liu, R.~Wu, B.~Van~Hoorick, P.~Tokmakov, S.~Zakharov, and C.~Vondrick.
\newblock Zero-1-to-3: Zero-shot one image to 3d object.
\newblock In {\em Proceedings of the IEEE/CVF international conference on computer vision}, pages 9298--9309, 2023.

\bibitem{Liu2024ASO}
X.~Liu, S.~Shen, B.~Li, P.~Ma, R.~Jiang, Y.~xin Zhang, J.~Fan, G.~Li, N.~Tang, and Y.~Luo.
\newblock A survey of text-to-sql in the era of llms: Where are we, and where are we going?
\newblock {\em IEEE Transactions on Knowledge and Data Engineering}, 37:5735--5754, 2024.

\bibitem{sqlsurvey}
X.~Liu, S.~Shen, B.~Li, P.~Ma, R.~Jiang, Y.~Zhang, J.~Fan, G.~Li, N.~Tang, and Y.~Luo.
\newblock A survey of nl2sql with large language models: Where are we, and where are we going?, 2025.

\bibitem{RoBERTa}
Y.~Liu, M.~Ott, N.~Goyal, J.~Du, M.~Joshi, D.~Chen, O.~Levy, M.~Lewis, L.~Zettlemoyer, and V.~Stoyanov.
\newblock Roberta: A robustly optimized bert pretraining approach, 2019.

\bibitem{liu2019roberta}
Y.~Liu, M.~Ott, N.~Goyal, J.~Du, M.~Joshi, D.~Chen, O.~Levy, M.~Lewis, L.~Zettlemoyer, and V.~Stoyanov.
\newblock Roberta: A robustly optimized bert pretraining approach.
\newblock {\em arXiv preprint arXiv:1907.11692}, 2019.

\bibitem{long2025bridging}
L.~Long, X.~Gu, X.~Sun, W.~Ye, H.~Wang, S.~Wu, G.~Chen, and J.~Zhao.
\newblock Bridging the semantic gap between text and table: A case study on nl2sql.
\newblock In {\em The Thirteenth International Conference on Learning Representations}, 2025.

\bibitem{lu2025bridging}
J.~Lu, Y.~Song, Z.~Qin, H.~Zhang, C.~Zhang, and R.~C.-W. Wong.
\newblock Bridging the gap: Enabling natural language queries for nosql databases through text-to-nosql translation.
\newblock {\em arXiv preprint arXiv:2502.11201}, 2025.

\bibitem{luo2025kbqa}
H.~Luo, Y.~Guo, Q.~Lin, X.~Wu, X.~Mu, W.~Liu, M.~Song, Y.~Zhu, L.~A. Tuan, et~al.
\newblock Kbqa-o1: Agentic knowledge base question answering with monte carlo tree search.
\newblock {\em arXiv preprint arXiv:2501.18922}, 2025.

\bibitem{luo2025natural}
Y.~Luo, G.~Li, J.~Fan, C.~Chai, and N.~Tang.
\newblock Natural language to sql: State of the art and open problems.
\newblock {\em Proceedings of the VLDB Endowment}, 18(12):5466--5471, 2025.

\bibitem{ma2023missmodal}
M.~Ma, J.~Ren, L.~Zhao, S.~Tulyakov, C.~Wu, and X.~Peng.
\newblock Missmodal: Increasing robustness to missing modality in multimodal sentiment analysis.
\newblock {\em Transactions of the Association for Computational Linguistics}, 11:1606--1620, 2023.

\bibitem{ma2023dreamtalk}
Y.~Ma, S.~Zhang, J.~Wang, X.~Wang, Y.~Zhang, and Z.~Deng.
\newblock Dreamtalk: When emotional talking head generation meets diffusion probabilistic models.
\newblock {\em arXiv preprint arXiv:2312.09767}, 2023.

\bibitem{ma2024spreadsheet}
Z.~Ma, B.~Zhang, J.~Zhang, J.~Yu, X.~Zhang, X.~Zhang, S.~Luo, X.~Wang, and J.~Tang.
\newblock Spreadsheetbench: Towards challenging real world spreadsheet manipulation, 2024.

\bibitem{maervoet2012outlier}
J.~Maervoet, C.~Vens, G.~Vanden~Berghe, H.~Blockeel, and P.~De~Causmaecker.
\newblock Outlier detection in relational data: A case study in geographical information systems.
\newblock {\em Expert Systems with Applications}, 39(5):4718--4728, 2012.

\bibitem{masry2023unichart}
A.~Masry, P.~Kavehzadeh, X.~L. Do, E.~Hoque, and S.~Joty.
\newblock Unichart: A universal vision-language pretrained model for chart comprehension and reasoning.
\newblock {\em arXiv preprint arXiv:2305.14761}, 2023.

\bibitem{masry2022chartqabenchmarkquestionanswering}
A.~Masry, D.~X. Long, J.~Q. Tan, S.~Joty, and E.~Hoque.
\newblock Chartqa: A benchmark for question answering about charts with visual and logical reasoning, 2022.

\bibitem{masry2024chartgemmavisualinstructiontuningchart}
A.~Masry, M.~Thakkar, A.~Bajaj, A.~Kartha, E.~Hoque, and S.~Joty.
\newblock Chartgemma: Visual instruction-tuning for chart reasoning in the wild, 2024.

\bibitem{mittal-etal-1998-describing}
V.~O. Mittal, J.~D. Moore, G.~Carenini, and S.~Roth.
\newblock Describing complex charts in natural language: A caption generation system.
\newblock {\em Computational Linguistics}, 24(3):431--467, 1998.

\bibitem{mu2024snagscalableaccuratevideo}
F.~Mu, S.~Mo, and Y.~Li.
\newblock Snag: Scalable and accurate video grounding, 2024.

\bibitem{muller2024predicting}
M.~M{\"u}ller, A.~Dupuis, T.~Zeulner, I.~Vazquez, J.~Hagerer, and P.~A. Gloor.
\newblock Predicting team well-being through face video analysis with ai.
\newblock {\em Applied Sciences}, 14(3):1284, 2024.

\bibitem{nan2022fetaqa}
L.~Nan, C.~Hsieh, Z.~Mao, X.~V. Lin, N.~Verma, R.~Zhang, W.~Kry{\'s}ci{\'n}ski, H.~Schoelkopf, R.~Kong, X.~Tang, et~al.
\newblock Fetaqa: Free-form table question answering.
\newblock {\em Transactions of the Association for Computational Linguistics}, 10:35--49, 2022.

\bibitem{naqvi2017time}
S.~N.~Z. Naqvi, S.~Yfantidou, and E.~Zim{\'a}nyi.
\newblock Time series databases and influxdb.
\newblock {\em Studienarbeit, Universit{\'e} Libre de Bruxelles}, 12:1--44, 2017.

\bibitem{nooralahzadeh2024explainable}
F.~Nooralahzadeh, Y.~Zhang, J.~Furst, and K.~Stockinger.
\newblock Explainable multi-modal data exploration in natural language via llm agent.
\newblock {\em arXiv preprint arXiv:2412.18428}, 2024.

\bibitem{obeid-hoque-2020-chart}
J.~Obeid and E.~Hoque.
\newblock Chart-to-text: Generating natural language descriptions for charts by adapting the transformer model.
\newblock In B.~Davis, Y.~Graham, J.~Kelleher, and Y.~Sripada, editors, {\em Proceedings of the 13th International Conference on Natural Language Generation}, pages 138--147, Dublin, Ireland, Dec. 2020. Association for Computational Linguistics.

\bibitem{okamoto2024crepe}
Y.~Okamoto, Y.~Baek, G.~Kim, R.~Nakao, D.~Kim, M.~B. Yim, S.~Park, and B.~Lee.
\newblock Crepe: Coordinate-aware end-to-end document parser.
\newblock {\em arXiv preprint arXiv:2406.04093}, 2024.

\bibitem{otani2024ltsim}
M.~Otani, N.~Inoue, K.~Kikuchi, and R.~Togashi.
\newblock Ltsim: Layout transportation-based similarity measure for evaluating layout generation.
\newblock {\em arXiv preprint arXiv:2407.12356}, 2024.

\bibitem{pan2024sipllm}
Z.~Pan, Y.~Jiang, S.~Garg, A.~Schneider, Y.~Nevmyvaka, and D.~Song.
\newblock \$s{\textasciicircum}2\${IP}-{LLM}: Semantic space informed prompt learning with {LLM} for time series forecasting.
\newblock In {\em Forty-first International Conference on Machine Learning}, 2024.

\bibitem{bib:parseur}
{Parseur}.
\newblock {Parseur: Powerful \& simple email, PDF and document parsing}.
\newblock \url{https://parseur.com/}, 2024.
\newblock Accessed on: 2024-10-25.

\bibitem{pasupat2015compositional}
P.~Pasupat and P.~Liang.
\newblock Compositional semantic parsing on semi-structured tables.
\newblock {\em arXiv preprint arXiv:1508.00305}, 2015.

\bibitem{patel2024lotus}
L.~Patel, S.~Jha, C.~Guestrin, and M.~Zaharia.
\newblock Lotus: Enabling semantic queries with llms over tables of unstructured and structured data.
\newblock {\em arXiv e-prints}, pages arXiv--2407, 2024.

\bibitem{patnaik2024cabinet}
S.~Patnaik, H.~Changwal, M.~Aggarwal, S.~Bhatia, Y.~Kumar, and B.~Krishnamurthy.
\newblock Cabinet: Content relevance based noise reduction for table question answering, 2024.

\bibitem{patnaik2025aesthetiq}
S.~Patnaik, R.~Jain, B.~Krishnamurthy, and M.~Sarkar.
\newblock Aesthetiq: Enhancing graphic layout design via aesthetic-aware preference alignment of multi-modal large language models.
\newblock {\em arXiv preprint arXiv:2503.00591}, 2025.

\bibitem{pei2024bio}
Q.~Pei, L.~Wu, K.~Gao, J.~Zhu, Y.~Wang, Z.~Wang, T.~Qin, and R.~Yan.
\newblock Leveraging biomolecule and natural language through multi-modal learning: A survey, 2024.

\bibitem{pisaneschi2023automatic}
M.~Pisaneschi, S.~Appalaraju, and R.~Manmatha.
\newblock Automatic generation of scientific papers for data augmentation in document layout analysis.
\newblock {\em Pattern Recognition Letters}, 167:87--94, 2023.

\bibitem{DINSQL}
M.~Pourreza and D.~Rafiei.
\newblock Din-sql: Decomposed in-context learning of text-to-sql with self-correction, 2023.

\bibitem{pp2024hysem}
N.~PP and A.~P.~N. Iyer.
\newblock Hysem: A context length optimized llm pipeline for unstructured tabular extraction.
\newblock {\em arXiv preprint arXiv:2408.09434}, 2024.

\bibitem{qiu2024richdreamer}
L.~Qiu, G.~Chen, X.~Gu, Q.~Zuo, M.~Xu, Y.~Wu, W.~Yuan, Z.~Dong, L.~Bo, and X.~Han.
\newblock Richdreamer: A generalizable normal-depth diffusion model for detail richness in text-to-3d.
\newblock In {\em Proceedings of the IEEE/CVF conference on computer vision and pattern recognition}, pages 9914--9925, 2024.

\bibitem{qwen2025qwen25}
Qwen, :, A.~Yang, B.~Yang, B.~Zhang, B.~Hui, B.~Zheng, B.~Yu, C.~Li, D.~Liu, F.~Huang, H.~Wei, H.~Lin, J.~Yang, J.~Tu, J.~Zhang, J.~Yang, J.~Yang, J.~Zhou, J.~Lin, K.~Dang, K.~Lu, K.~Bao, K.~Yang, L.~Yu, M.~Li, M.~Xue, P.~Zhang, Q.~Zhu, R.~Men, R.~Lin, T.~Li, T.~Tang, T.~Xia, X.~Ren, X.~Ren, Y.~Fan, Y.~Su, Y.~Zhang, Y.~Wan, Y.~Liu, Z.~Cui, Z.~Zhang, and Z.~Qiu.
\newblock Qwen2.5 technical report, 2025.

\bibitem{rafailov2024dpo}
R.~Rafailov, A.~Sharma, E.~Mitchell, S.~Ermon, C.~D. Manning, and C.~Finn.
\newblock Direct preference optimization: Your language model is secretly a reward model, 2024.

\bibitem{rahmani2022association}
F.~Rahmani and M.~H. Fattahi.
\newblock Association between forecasting models’ precision and nonlinear patterns of daily river flow time series.
\newblock {\em Modeling Earth Systems and Environment}, 8(3):4267--4276, 2022.

\bibitem{reiter-2007-architecture}
E.~Reiter.
\newblock An architecture for data-to-text systems.
\newblock In S.~Busemann, editor, {\em Proceedings of the Eleventh {E}uropean Workshop on Natural Language Generation ({ENLG} 07)}, pages 97--104, Saarbr{\"u}cken, Germany, June 2007. DFKI GmbH.

\bibitem{ren2024graphsurvey}
X.~Ren, J.~Tang, D.~Yin, N.~Chawla, and C.~Huang.
\newblock A survey of large language models for graphs.
\newblock In {\em Proceedings of the 30th ACM SIGKDD Conference on Knowledge Discovery and Data Mining}, pages 6616--6626, 2024.

\bibitem{sengonul2025abnormal}
E.~Sengonul, R.~Samet, Q.~Abu Al-Haija, A.~Alqahtani, R.~A. Alsemmeari, B.~Alghamdi, B.~Alturki, and A.~A. Alsulami.
\newblock Abnormal event detection in surveillance videos through lstm auto-encoding and local minima assistance.
\newblock {\em Discover Internet of Things}, 5(1):32, 2025.

\bibitem{senin2008dynamic}
P.~Senin.
\newblock Dynamic time warping algorithm review.
\newblock {\em Information and Computer Science Department University of Hawaii at Manoa Honolulu, USA}, 855(1-23):40, 2008.

\bibitem{seol2024posterllama}
J.~Seol, S.~Kim, and J.~Yoo.
\newblock Posterllama: Bridging design ability of langauge model to content-aware layout generation.
\newblock {\em arXiv preprint arXiv:2404.00995}, 2024.

\bibitem{shang2024graphsurvey}
W.~Shang and X.~Huang.
\newblock A survey of large language models on generative graph analytics: Query, learning, and applications.
\newblock {\em arXiv preprint arXiv:2404.14809}, 2024.

\bibitem{shankar2025docetl}
S.~Shankar, T.~Chambers, T.~Shah, A.~G. Parameswaran, and E.~Wu.
\newblock Docetl: Agentic query rewriting and evaluation for complex document processing, 2025.

\bibitem{shen2024tempme}
L.~Shen, T.~Hao, T.~He, S.~Zhao, Y.~Zhang, P.~Liu, Y.~Bao, and G.~Ding.
\newblock Tempme: Video temporal token merging for efficient text-video retrieval.
\newblock {\em arXiv preprint arXiv:2409.01156}, 2024.

\bibitem{shi2025layoutcot}
H.~Shi, J.~Su, R.~Xu, and J.~Gao.
\newblock Layoutcot: Unleashing the deep reasoning potential of large language models for layout generation.
\newblock {\em arXiv preprint arXiv:2504.10829}, 2025.

\bibitem{shi2024sqlsurvey}
L.~Shi, Z.~Tang, N.~Zhang, X.~Zhang, and Z.~Yang.
\newblock A survey on employing large language models for text-to-sql tasks.
\newblock {\em ACM Computing Surveys}, 2024.

\bibitem{shi2024survey}
L.~Shi, Z.~Tang, N.~Zhang, X.~Zhang, and Z.~Yang.
\newblock A survey on employing large language models for text-to-sql tasks.
\newblock {\em ACM Computing Surveys}, 2024.

\bibitem{9005997}
S.~Siami-Namini, N.~Tavakoli, and A.~S. Namin.
\newblock The performance of lstm and bilstm in forecasting time series.
\newblock In {\em 2019 IEEE International Conference on Big Data (Big Data)}, pages 3285--3292, 2019.

\bibitem{8614252}
S.~Siami-Namini, N.~Tavakoli, and A.~Siami~Namin.
\newblock A comparison of arima and lstm in forecasting time series.
\newblock In {\em 2018 17th IEEE International Conference on Machine Learning and Applications (ICMLA)}, pages 1394--1401, 2018.

\bibitem{singer2022make}
U.~Singer, A.~Polyak, T.~Hayes, X.~Yin, J.~An, S.~Zhang, Q.~Hu, H.~Yang, O.~Ashual, O.~Gafni, et~al.
\newblock Make-a-video: Text-to-video generation without text-video data.
\newblock {\em arXiv preprint arXiv:2209.14792}, 2022.

\bibitem{singh2025figcapshffiguretocaptiongenerativeframework}
A.~Singh, A.~Singh, P.~Agarwal, Z.~Huang, A.~Singh, T.~Yu, S.~Kim, V.~Bursztyn, N.~K. Ahmed, P.~Mathur, E.~Learned-Miller, F.~Dernoncourt, and R.~A. Rossi.
\newblock Figcaps-hf: A figure-to-caption generative framework and benchmark with human feedback, 2025.

\bibitem{solatorio2023realtabformer}
A.~V. Solatorio and O.~Dupriez.
\newblock Realtabformer: Generating realistic relational and tabular data using transformers.
\newblock {\em arXiv preprint arXiv:2302.02041}, 2023.

\bibitem{song2023self}
D.~Song, C.-M. Zhang, X.-Q. Zhao, T.~Wang, W.-Z. Nie, X.-Y. Li, and A.-A. Liu.
\newblock Self-supervised image-based 3d model retrieval.
\newblock {\em ACM Transactions on Multimedia Computing, Communications and Applications}, 19(2):1--18, 2023.

\bibitem{su2024tablegpt2}
A.~Su, A.~Wang, C.~Ye, C.~Zhou, G.~Zhang, G.~Chen, G.~Zhu, H.~Wang, H.~Xu, H.~Chen, H.~Li, H.~Lan, J.~Tian, J.~Yuan, J.~Zhao, J.~Zhou, K.~Shou, L.~Zha, L.~Long, L.~Li, P.~Wu, Q.~Zhang, Q.~Huang, S.~Yang, T.~Zhang, W.~Ye, W.~Zhu, X.~Hu, X.~Gu, X.~Sun, X.~Li, Y.~Yang, and Z.~Xiao.
\newblock Tablegpt2: A large multimodal model with tabular data integration, 2024.

\bibitem{suhara2022annotating}
Y.~Suhara, J.~Li, Y.~Li, D.~Zhang, {\c{C}}.~Demiralp, C.~Chen, and W.-C. Tan.
\newblock Annotating columns with pre-trained language models.
\newblock In {\em Proceedings of the 2022 International Conference on Management of Data}, pages 1493--1503, 2022.

\bibitem{sun2025agenticdata}
J.~Sun, G.~Li, P.~Zhou, Y.~Ma, J.~Xu, and Y.~Li.
\newblock Agenticdata: An agentic data analytics system for heterogeneous data, 2025.

\bibitem{sun2025quest}
Z.~Sun, Q.~Deng, C.~Chai, K.~Jin, X.~Guo, H.~Han, Y.~Yuan, G.~Wang, and L.~Cao.
\newblock Quest: Query optimization in unstructured document analysis, 2025.

\bibitem{talaei2024chess}
S.~Talaei, M.~Pourreza, Y.-C. Chang, A.~Mirhoseini, and A.~Saberi.
\newblock Chess: Contextual harnessing for efficient sql synthesis, 2024.

\bibitem{tanaka2024scipostlayout}
S.~Tanaka, H.~Wang, and Y.~Ushiku.
\newblock Scipostlayout: A dataset for layout analysis and layout generation of scientific posters.
\newblock {\em arXiv preprint arXiv:2407.19787}, 2024.

\bibitem{tang2024graphgpt}
J.~Tang, Y.~Yang, W.~Wei, L.~Shi, L.~Su, S.~Cheng, D.~Yin, and C.~Huang.
\newblock Graphgpt: Graph instruction tuning for large language models, 2024.

\bibitem{tang2025videosurvey}
Y.~Tang, J.~Bi, S.~Xu, L.~Song, S.~Liang, T.~Wang, D.~Zhang, J.~An, J.~Lin, R.~Zhu, et~al.
\newblock Video understanding with large language models: A survey.
\newblock {\em IEEE Transactions on Circuits and Systems for Video Technology}, 2025.

\bibitem{tang2025st}
Z.~Tang, B.~Niu, X.~Zhou, B.~Li, W.~Zhou, J.~Wang, G.~Li, X.~Zhang, and F.~Wu.
\newblock St-raptor: Llm-powered semi-structured table question answering.
\newblock {\em arXiv preprint arXiv:2508.18190}, 2025.

\bibitem{taskar2001probabilistic}
B.~Taskar, E.~Segal, and D.~Koller.
\newblock Probabilistic classification and clustering in relational data.
\newblock In {\em International joint conference on artificial intelligence}, volume~17, pages 870--878. Lawrence Erlbaum Associates LTD, 2001.

\bibitem{tickoo2018autodesk}
S.~Tickoo.
\newblock {\em Autodesk Maya 2019: A Comprehensive Guide}.
\newblock Cadcim Technologies, 2018.

\bibitem{Unicorn}
J.~Tu, J.~Fan, N.~Tang, P.~Wang, G.~Li, X.~Du, X.~Jia, and S.~Gao.
\newblock Unicorn: {A} unified multi-tasking model for supporting matching tasks in data integration.
\newblock {\em Proc. {ACM} Manag. Data}, 1(1):84:1--84:26, 2023.

\bibitem{urban2023caesura}
M.~Urban and C.~Binnig.
\newblock Caesura: Language models as multi-modal query planners, 2023.

\bibitem{valle2023ge}
C.~Valle.
\newblock Ge predix and the digital twin.
\newblock In {\em AIAA SCITECH 2023 Forum}, page 2626, 2023.

\bibitem{wan2024omniparser}
J.~Wan, S.~Song, W.~Yu, Y.~Liu, W.~Cheng, F.~Huang, X.~Bai, C.~Yao, and Z.~Yang.
\newblock Omniparser: A unified framework for text spotting, key information extraction and table recognition.
\newblock In {\em 2024 IEEE/CVF Conference on Computer Vision and Pattern Recognition (CVPR)}, pages 15641--15653, 2024.

\bibitem{wang2024protchatgpt}
C.~Wang, H.~Fan, R.~Quan, and Y.~Yang.
\newblock Protchatgpt: Towards understanding proteins with large language models.
\newblock {\em arXiv preprint arXiv:2402.09649}, 2024.

\bibitem{wang2024docllm}
D.~Wang, Z.~Ma, A.~Nourbakhsh, K.~Mangalampalli, and S.~Shah.
\newblock Docllm: A layout-aware generative language model for multimodal document understanding.
\newblock {\em arXiv preprint arXiv:2401.00908}, 2024.

\bibitem{wang2024grounded}
H.~Wang, Z.~Xu, Y.~Cheng, S.~Diao, Y.~Zhou, Y.~Cao, Q.~Wang, W.~Ge, and L.~Huang.
\newblock Grounded-videollm: Sharpening fine-grained temporal grounding in video large language models.
\newblock {\em arXiv preprint arXiv:2410.03290}, 2024.

\bibitem{wang2025towards}
J.~Wang, Y.~Feng, C.~Shen, S.~Rahman, and E.~Kandogan.
\newblock Towards operationalizing heterogeneous data discovery.
\newblock {\em arXiv preprint arXiv:2504.02059}, 2025.

\bibitem{wang2024dlaformer}
J.~Wang, K.~Hu, and Q.~Huo.
\newblock Dlaformer: An end-to-end transformer for document layout analysis.
\newblock {\em arXiv preprint arXiv:2405.11757}, 2024.

\bibitem{wang2025aop}
J.~Wang and G.~Li.
\newblock Aop: Automated and interactive llm pipeline orchestration for answering complex queries.
\newblock CIDR, 2025.

\bibitem{9713986}
J.~Wang, R.~Li, R.~Li, B.~Fu, and D.~Z. Chen.
\newblock Hmckrautoencoder: An interpretable deep learning framework for time series analysis.
\newblock {\em IEEE Transactions on Emerging Topics in Computing}, 10(1):99--111, 2022.

\bibitem{wang20253dsmiles}
J.~Wang, H.~Luo, R.~Qin, M.~Wang, X.~Wan, M.~Fang, O.~Zhang, Q.~Gou, Q.~Su, C.~Shen, et~al.
\newblock 3dsmiles-gpt: 3d molecular pocket-based generation with token-only large language model.
\newblock {\em Chemical Science}, 16(2):637--648, 2025.

\bibitem{wang2024instructgraph}
J.~Wang, J.~Wu, Y.~Hou, Y.~Liu, M.~Gao, and J.~McAuley.
\newblock Instructgraph: Boosting large language models via graph-centric instruction tuning and preference alignment, 2024.

\bibitem{wang2017heterogeneous}
L.~Wang.
\newblock Heterogeneous data and big data analytics.
\newblock In {\em ACIS}, volume~3, pages 8--15, 2017.

\bibitem{wang2024must}
M.~Wang, X.~Ke, X.~Xu, L.~Chen, Y.~Gao, P.~Huang, and R.~Zhu.
\newblock Must: An effective and scalable framework for multimodal search of target modality.
\newblock In {\em 2024 IEEE 40th International Conference on Data Engineering (ICDE)}. IEEE, 2024.

\bibitem{wang2024interactive}
M.~Wang, H.~Wu, X.~Ke, Y.~Gao, X.~Xu, and L.~Chen.
\newblock An interactive multi-modal query answering system with retrieval-augmented large language models, 2024.

\bibitem{wang2021neus}
P.~Wang, L.~Liu, Y.~Liu, C.~Theobalt, T.~Komura, and W.~Wang.
\newblock Neus: Learning neural implicit surfaces by volume rendering for multi-view reconstruction.
\newblock {\em arXiv preprint arXiv:2106.10689}, 2021.

\bibitem{wang2022webformer}
Q.~Wang, Y.~Fang, A.~Ravula, F.~Feng, X.~Quan, and D.~Liu.
\newblock Webformer: The web-page transformer for structure information extraction.
\newblock In {\em Proceedings of the ACM Web Conference 2022}, pages 3124--3133, 2022.

\bibitem{wang2025videoitg}
S.~Wang, G.~Chen, D.-a. Huang, Z.~Li, M.~Li, G.~Li, J.~M. Alvarez, L.~Zhang, and Z.~Yu.
\newblock Videoitg: Multimodal video understanding with instructed temporal grounding.
\newblock {\em arXiv preprint arXiv:2507.13353}, 2025.

\bibitem{wang2024disco}
T.~Wang, L.~Li, K.~Lin, Y.~Zhai, C.-C. Lin, Z.~Yang, H.~Zhang, Z.~Liu, and L.~Wang.
\newblock Disco: Disentangled control for realistic human dance generation.
\newblock In {\em Proceedings of the IEEE/CVF Conference on Computer Vision and Pattern Recognition}, pages 9326--9336, 2024.

\bibitem{wang2025plugging}
X.~Wang, M.~Costa, J.~Kovaceva, S.~Wang, and F.~C. Pereira.
\newblock Plugging schema graph into multi-table qa: A human-guided framework for reducing llm reliance.
\newblock {\em arXiv preprint arXiv:2506.04427}, 2025.

\bibitem{wang2024news}
X.~Wang, M.~Feng, J.~Qiu, J.~Gu, and J.~Zhao.
\newblock From news to forecast: Integrating event analysis in llm-based time series forecasting with reflection.
\newblock {\em Advances in Neural Information Processing Systems}, 37:58118--58153, 2024.

\bibitem{wang2025large}
X.~Wang, Y.~Wang, W.~Zhu, and R.~Wang.
\newblock Do large language models truly understand geometric structures?
\newblock {\em arXiv preprint arXiv:2501.13773}, 2025.

\bibitem{wang2002detecting}
Y.~Wang and J.~Hu.
\newblock Detecting tables in html documents.
\newblock In {\em International Workshop on Document Analysis Systems}, pages 249--260. Springer, 2002.

\bibitem{DBLP:journals/corr/abs-2411-13724}
Y.~Wang and H.~A. Karimi.
\newblock Exploring large language models for climate forecasting.
\newblock {\em CoRR}, abs/2411.13724, 2024.

\bibitem{wang2021tuta}
Z.~Wang, H.~Dong, R.~Jia, J.~Li, Z.~Fu, S.~Han, and D.~Zhang.
\newblock Tuta: Tree-based transformers for generally structured table pre-training.
\newblock In {\em Proceedings of the 27th ACM SIGKDD Conference on Knowledge Discovery \& Data Mining}, pages 1780--1790, 2021.

\bibitem{wang2024llama}
Z.~Wang, J.~Lorraine, Y.~Wang, H.~Su, J.~Zhu, S.~Fidler, and X.~Zeng.
\newblock Llama-mesh: Unifying 3d mesh generation with language models.
\newblock {\em arXiv preprint arXiv:2411.09595}, 2024.

\bibitem{wang2024chainoftable}
Z.~Wang, H.~Zhang, C.-L. Li, J.~M. Eisenschlos, V.~Perot, Z.~Wang, L.~Miculicich, Y.~Fujii, J.~Shang, C.-Y. Lee, and T.~Pfister.
\newblock Chain-of-table: Evolving tables in the reasoning chain for table understanding, 2024.

\bibitem{weininger1988smiles}
D.~Weininger.
\newblock Smiles, a chemical language and information system. 1. introduction to methodology and encoding rules.
\newblock {\em Journal of chemical information and computer sciences}, 28(1):31--36, 1988.

\bibitem{weng2025artificial}
Z.~Weng, L.~Bravo-S{\'a}nchez, Z.~Wang, C.~Howard, M.~Xenochristou, N.~Meister, A.~Kanazawa, A.~Milstein, E.~Bergelson, K.~L. Humphreys, et~al.
\newblock Artificial intelligence--powered 3d analysis of video-based caregiver-child interactions.
\newblock {\em Science Advances}, 11(8):eadp4422, 2025.

\bibitem{whitehead2025utilizing}
R.~Whitehead, A.~Nguyen, and S.~J{\"a}rvel{\"a}.
\newblock Utilizing multimodal large language models for video analysis of posture in studying collaborative learning: A case study.
\newblock {\em Journal of Learning Analytics}, 12(1):186--200, 2025.

\bibitem{wikipedia_3dmodel}
{Wikipedia contributors}.
\newblock 3d model, 2025.
\newblock Accessed: 2025-09-08.

\bibitem{wu2022nuwa}
C.~Wu, J.~Liang, X.~Hu, Z.~Gan, J.~Wang, L.~Wang, Z.~Liu, Y.~Fang, and N.~Duan.
\newblock Nuwa-infinity: Autoregressive over autoregressive generation for infinite visual synthesis.
\newblock {\em arXiv preprint arXiv:2207.09814}, 2022.

\bibitem{wu2024direct3d}
S.~Wu, Y.~Lin, F.~Zhang, Y.~Zeng, J.~Xu, P.~Torr, X.~Cao, and Y.~Yao.
\newblock Direct3d: Scalable image-to-3d generation via 3d latent diffusion transformer.
\newblock {\em Advances in Neural Information Processing Systems}, 37:121859--121881, 2024.

\bibitem{wu2025tablesurvey}
X.~Wu, A.~Ritter, and W.~Xu.
\newblock Tabular data understanding with llms: A survey of recent advances and challenges.
\newblock {\em arXiv preprint arXiv:2508.00217}, 2025.

\bibitem{wu2024chartinsights}
Y.~Wu, L.~Yan, L.~Shen, Y.~Wang, N.~Tang, and Y.~Luo.
\newblock Chartinsights: Evaluating multimodal large language models for low-level chart question answering.
\newblock {\em arXiv preprint arXiv:2405.07001}, 2024.

\bibitem{xiao2024cellagent}
Y.~Xiao, J.~Liu, Y.~Zheng, X.~Xie, J.~Hao, M.~Li, R.~Wang, F.~Ni, Y.~Li, J.~Luo, S.~Jiao, and J.~Peng.
\newblock Cellagent: An llm-driven multi-agent framework for automated single-cell data analysis, 2024.

\bibitem{DBLP:journals/pacmmod/XieXZG25}
X.~Xie, G.~Xu, L.~Zhao, and R.~Guo.
\newblock Opensearch-sql: Enhancing text-to-sql with dynamic few-shot and consistency alignment.
\newblock {\em Proc. {ACM} Manag. Data}, 3(3):194:1--194:24, 2025.

\bibitem{xie2024chatts}
Z.~Xie, Z.~Li, X.~He, L.~Xu, X.~Wen, T.~Zhang, J.~Chen, R.~Shi, and D.~Pei.
\newblock Chatts: Aligning time series with llms via synthetic data for enhanced understanding and reasoning.
\newblock {\em arXiv preprint arXiv:2412.03104}, 2024.

\bibitem{xing2025mmtu}
J.~Xing, Y.~He, M.~Zhou, H.~Dong, S.~Han, L.~Chen, D.~Zhang, S.~Chaudhuri, and H.~Jagadish.
\newblock Mmtu: A massive multi-task table understanding and reasoning benchmark.
\newblock {\em arXiv preprint arXiv:2506.05587}, 2025.

\bibitem{xiong2024interactivekbqa}
G.~Xiong, J.~Bao, and W.~Zhao.
\newblock Interactive-kbqa: Multi-turn interactions for knowledge base question answering with large language models, 2024.

\bibitem{xiong2025mcts}
G.~Xiong, H.~Li, and W.~Zhao.
\newblock Mcts-kbqa: Monte carlo tree search for knowledge base question answering.
\newblock {\em arXiv preprint arXiv:2502.13428}, 2025.

\bibitem{xiong20253ur}
H.~Xiong, Y.~Zhuge, J.~Zhu, L.~Zhang, and H.~Lu.
\newblock 3ur-llm: An end-to-end multimodal large language model for 3d scene understanding.
\newblock {\em IEEE Transactions on Multimedia}, 2025.

\bibitem{xu2024hierarchical}
H.~Xu, L.~Chen, Z.~Zhao, D.~Ma, R.~Cao, Z.~Zhu, and K.~Yu.
\newblock Hierarchical multimodal pre-training for visually rich webpage understanding.
\newblock In {\em Proceedings of the 17th ACM International Conference on Web Search and Data Mining}, pages 864--872, 2024.

\bibitem{xu2022synthetic}
K.~Xu, G.~Ganev, E.~Joubert, R.~Davison, O.~Van~Acker, and L.~Robinson.
\newblock Synthetic data generation of many-to-many datasets via random graph generation.
\newblock In {\em The Eleventh International Conference on Learning Representations}, 2022.

\bibitem{xu2020layoutlm}
Y.~Xu, M.~Li, L.~Cui, S.~Huang, F.~Wei, and M.~Zhou.
\newblock Layoutlm: Pre-training of text and layout for document image understanding.
\newblock {\em arXiv preprint arXiv:1912.13318}, 2020.

\bibitem{xu2021layoutlmv2}
Y.~Xu, Y.~Xu, T.~Lv, L.~Cui, F.~Wei, G.~Wang, Y.~Lu, D.~Florencio, C.~Zhang, W.~Che, M.~Zhang, and L.~Zhou.
\newblock Layoutlmv2: Multi-modal pre-training for visually-rich document understanding.
\newblock pages 2579--2591, 2021.

\bibitem{xu2024chartbenchbenchmarkcomplexvisual}
Z.~Xu, S.~Du, Y.~Qi, C.~Xu, C.~Yuan, and J.~Guo.
\newblock Chartbench: A benchmark for complex visual reasoning in charts, 2024.

\bibitem{xu2025chartmoemixturediverselyaligned}
Z.~Xu, B.~Qu, Y.~Qi, S.~Du, C.~Xu, C.~Yuan, and J.~Guo.
\newblock Chartmoe: Mixture of diversely aligned expert connector for chart understanding, 2025.

\bibitem{yan2024corrective}
S.-Q. Yan, J.-C. Gu, Y.~Zhu, and Z.-H. Ling.
\newblock Corrective retrieval augmented generation.
\newblock {\em arXiv preprint arXiv:2401.15884}, 2024.

\bibitem{yang2025chartmimicevaluatinglmmscrossmodal}
C.~Yang, C.~Shi, Y.~Liu, B.~Shui, J.~Wang, M.~Jing, L.~Xu, X.~Zhu, S.~Li, Y.~Zhang, G.~Liu, X.~Nie, D.~Cai, and Y.~Yang.
\newblock Chartmimic: Evaluating lmm's cross-modal reasoning capability via chart-to-code generation, 2025.

\bibitem{yang2024llmi3d}
F.~Yang, S.~Zhao, Y.~Zhang, H.~Chen, H.~Chen, W.~Tang, H.~Lu, P.~Xu, Z.~Yang, J.~Han, et~al.
\newblock Llmi3d: Empowering llm with 3d perception from a single 2d image.
\newblock {\em arXiv preprint arXiv:2408.07422}, 2024.

\bibitem{yang2022tableformer}
J.~Yang, A.~Gupta, S.~Upadhyay, L.~He, R.~Goel, and S.~Paul.
\newblock Tableformer: Robust transformer modeling for table-text encoding.
\newblock {\em arXiv preprint arXiv:2203.00274}, 2022.

\bibitem{yang2025timerag}
S.~Yang, D.~Wang, H.~Zheng, and R.~Jin.
\newblock Timerag: Boosting llm time series forecasting via retrieval-augmented generation.
\newblock In {\em ICASSP 2025-2025 IEEE International Conference on Acoustics, Speech and Signal Processing (ICASSP)}, pages 1--5. IEEE, 2025.

\bibitem{yang2024hunyuan3d}
X.~Yang, H.~Shi, B.~Zhang, F.~Yang, J.~Wang, H.~Zhao, X.~Liu, X.~Wang, Q.~Lin, J.~Yu, et~al.
\newblock Hunyuan3d 1.0: A unified framework for text-to-3d and image-to-3d generation.
\newblock {\em arXiv preprint arXiv:2411.02293}, 2024.

\bibitem{ye2024mplugowlmodularizationempowerslarge}
Q.~Ye, H.~Xu, G.~Xu, J.~Ye, M.~Yan, Y.~Zhou, J.~Wang, A.~Hu, P.~Shi, Y.~Shi, C.~Li, Y.~Xu, H.~Chen, J.~Tian, Q.~Qian, J.~Zhang, F.~Huang, and J.~Zhou.
\newblock mplug-owl: Modularization empowers large language models with multimodality, 2024.

\bibitem{ye2024instructglm}
R.~Ye, C.~Zhang, R.~Wang, S.~Xu, and Y.~Zhang.
\newblock Language is all a graph needs, 2024.

\bibitem{yin2022code}
P.~Yin, W.-D. Li, K.~Xiao, A.~Rao, Y.~Wen, K.~Shi, J.~Howland, P.~Bailey, M.~Catasta, H.~Michalewski, A.~Polozov, and C.~Sutton.
\newblock Natural language to code generation in interactive data science notebooks, 2022.

\bibitem{DBLP:conf/emnlp/YuYYZWLR18}
T.~Yu, M.~Yasunaga, K.~Yang, R.~Zhang, D.~Wang, Z.~Li, and D.~R. Radev.
\newblock Syntaxsqlnet: Syntax tree networks for complex and cross-domain text-to-sql task.
\newblock In E.~Riloff, D.~Chiang, J.~Hockenmaier, and J.~Tsujii, editors, {\em Proceedings of the 2018 Conference on Empirical Methods in Natural Language Processing, Brussels, Belgium, October 31 - November 4, 2018}, pages 1653--1663. Association for Computational Linguistics, 2018.

\bibitem{yu2018spider}
T.~Yu, R.~Zhang, K.~Yang, M.~Yasunaga, D.~Wang, Z.~Li, J.~Ma, I.~Li, Q.~Yao, S.~Roman, et~al.
\newblock Spider: A large-scale human-labeled dataset for complex and cross-domain semantic parsing and text-to-sql task.
\newblock {\em arXiv preprint arXiv:1809.08887}, 2018.

\bibitem{yu2023temporal}
X.~Yu, Z.~Chen, Y.~Ling, S.~Dong, Z.~Liu, and Y.~Lu.
\newblock Temporal data meets llm--explainable financial time series forecasting.
\newblock {\em arXiv preprint arXiv:2306.11025}, 2023.

\bibitem{yuan2025videorefer}
Y.~Yuan, H.~Zhang, W.~Li, Z.~Cheng, B.~Zhang, L.~Li, X.~Li, D.~Zhao, W.~Zhang, Y.~Zhuang, et~al.
\newblock Videorefer suite: Advancing spatial-temporal object understanding with video llm.
\newblock In {\em Proceedings of the Computer Vision and Pattern Recognition Conference}, pages 18970--18980, 2025.

\bibitem{zadeh2025text2chart31instructiontuningchart}
F.~P. Zadeh, J.~Kim, J.-H. Kim, and G.~Kim.
\newblock Text2chart31: Instruction tuning for chart generation with automatic feedback, 2025.

\bibitem{zhang2024finsql}
C.~Zhang, Y.~Mao, Y.~Fan, Y.~Mi, Y.~Gao, L.~Chen, D.~Lou, and J.~Lin.
\newblock Finsql: Model-agnostic llms-based text-to-sql framework for financial analysis, 2024.

\bibitem{zhang2024towards}
H.~Zhang.
\newblock Towards autonomous construction with viact.ai.
\newblock {\em White Paper, viAct}, 2024.
\newblock Accessed on: 2024-10-25.

\bibitem{zhang2024vascar}
J.~Zhang, R.~Yoshihashi, R.~Kawakami, and Y.~Nakashima.
\newblock Vascar: Content-aware layout generation via visual-aware self-correction.
\newblock {\em arXiv preprint arXiv:2412.04237}, 2024.

\bibitem{zhang2022subgraph}
J.~Zhang, X.~Zhang, J.~Yu, J.~Tang, J.~Tang, C.~Li, and H.~Chen.
\newblock Subgraph retrieval enhanced model for multi-hop knowledge base question answering.
\newblock In S.~Muresan, P.~Nakov, and A.~Villavicencio, editors, {\em Proceedings of the 60th Annual Meeting of the Association for Computational Linguistics (Volume 1: Long Papers)}, pages 5773--5784, Dublin, Ireland, May 2022. Association for Computational Linguistics.

\bibitem{dococr}
Q.~Zhang, B.~Wang, V.~S.-J. Huang, J.~Zhang, Z.~Wang, H.~Liang, C.~He, and W.~Zhang.
\newblock Document parsing unveiled: Techniques, challenges, and prospects for structured information extraction.
\newblock {\em arXiv preprint arXiv:2410.21169}, 2024.

\bibitem{zhang2024raft}
T.~Zhang, Z.~Chen, W.~Li, and D.~Chen.
\newblock {RAFT}: Retrieval augmented fine-tuning for language models.
\newblock {\em arXiv preprint arXiv:2403.10131}, 2024.

\bibitem{zhang2023sadtalker}
W.~Zhang, X.~Cun, X.~Wang, Y.~Zhang, X.~Shen, Y.~Guo, Y.~Shan, and F.~Wang.
\newblock Sadtalker: Learning realistic 3d motion coefficients for stylized audio-driven single image talking face animation.
\newblock In {\em Proceedings of the IEEE/CVF conference on computer vision and pattern recognition}, pages 8652--8661, 2023.

\bibitem{zhang2023reactable}
Y.~Zhang, J.~Henkel, A.~Floratou, J.~Cahoon, S.~Deep, and J.~M. Patel.
\newblock Reactable: Enhancing react for table question answering, 2023.

\bibitem{zhao2024tabpedia}
W.~Zhao, H.~Feng, Q.~Liu, J.~Tang, S.~Wei, B.~Wu, L.~Liao, Y.~Ye, H.~Liu, W.~Zhou, H.~Li, and C.~Huang.
\newblock Tabpedia: Towards comprehensive visual table understanding with concept synergy, 2024.

\bibitem{chat2data2024}
X.~Zhao, X.~Zhou, and G.~Li.
\newblock Chat2data: An interactive data analysis system with rag, vector databases and llms.
\newblock {\em Proc. VLDB Endow.}, 17(12):4481–4484, Aug. 2024.

\bibitem{zhao2024tapera}
Y.~Zhao, L.~Chen, A.~Cohan, and C.~Zhao.
\newblock {T}a{PERA}: Enhancing faithfulness and interpretability in long-form table {QA} by content planning and execution-based reasoning.
\newblock In L.-W. Ku, A.~Martins, and V.~Srikumar, editors, {\em Proceedings of the 62nd Annual Meeting of the Association for Computational Linguistics (Volume 1: Long Papers)}, pages 12824--12840, Bangkok, Thailand, Aug. 2024. Association for Computational Linguistics.

\bibitem{zheng2023judging}
L.~Zheng, W.-L. Chiang, Y.~Sheng, S.~Zhuang, Z.~Wu, Y.~Zhuang, Z.~Lin, Z.~Li, D.~Li, E.~Xing, et~al.
\newblock Judging llm-as-a-judge with mt-bench and chatbot arena.
\newblock {\em Advances in Neural Information Processing Systems}, 36:46595--46623, 2023.

\bibitem{zheng2024multimodal}
M.~Zheng, X.~Feng, Q.~Si, Q.~She, Z.~Lin, W.~Jiang, and W.~Wang.
\newblock Multimodal table understanding, 2024.

\bibitem{zhou2025cracksql}
W.~Zhou, Y.~Gao, X.~Zhou, and G.~Li.
\newblock {Cracking SQL Barriers:} {An} llm-based dialect transaltion system.
\newblock {\em Proc. {ACM} Manag. Data}, 3(3 (SIGMOD)), 2025.

\bibitem{zhou2025cracksqldemo}
W.~Zhou, Y.~Gao, X.~Zhou, and G.~Li.
\newblock Cracksql: A hybrid sql dialect translation system powered by large language models.
\newblock {\em arXiv Preprint}, 2025.

\bibitem{zhou2024r3}
Y.~Zhou, Y.~He, S.~Tian, Y.~Ni, Z.~Yin, X.~Liu, C.~Ji, S.~Liu, X.~Qiu, G.~Ye, and H.~Chai.
\newblock $r^3$-{NL}2{GQL}: A model coordination and knowledge graph alignment approach for {NL}2{GQL}.
\newblock In Y.~Al-Onaizan, M.~Bansal, and Y.-N. Chen, editors, {\em Findings of the Association for Computational Linguistics: EMNLP 2024}, pages 13679--13692, Miami, Florida, USA, Nov. 2024. Association for Computational Linguistics.

\bibitem{tatllm}
F.~Zhu, Z.~Liu, F.~Feng, C.~Wang, M.~Li, and T.~S. Chua.
\newblock Tat-llm: A specialized language model for discrete reasoning over financial tabular and textual data.
\newblock In {\em Proceedings of the 5th ACM International Conference on AI in Finance}, ICAIF '24, page 310–318, New York, NY, USA, 2024. Association for Computing Machinery.

\bibitem{TKGGen}
J.~Zhu, Y.~Fu, J.~Zhou, and D.~Chen.
\newblock A temporal knowledge graph generation dataset supervised distantly by large language models.
\newblock {\em Scientific Data}, 12, 05 2025.

\bibitem{zhu2024sqlsurvey}
X.~Zhu, Q.~Li, L.~Cui, and Y.~Liu.
\newblock Large language model enhanced text-to-sql generation: A survey.
\newblock {\em arXiv preprint arXiv:2410.06011}, 2024.

\end{thebibliography}
